\documentclass[twoside,11pt]{article}

\usepackage[preprint]{jmlr2e}

\usepackage{natbib,hyperref} 
\let\cite\citep

\usepackage{amsmath}
\usepackage{dsfont,amssymb}
\newtheorem{comment}{Comment}
\usepackage[utf8]{inputenc}

\newtheorem{lemmarep}{Lemma}
\usepackage{graphicx}
\usepackage{multirow}
\usepackage{color}
\usepackage{comment}
\usepackage{microtype}
\usepackage{wrapfig}
\usepackage[
font=footnotesize 
]{subfig}
\usepackage{booktabs}
\usepackage{tikz}
\usetikzlibrary{arrows, positioning, automata}
\usepackage{caption}
\usepackage{enumitem,kantlipsum}

\newcommand{\BN}{\ensuremath{\mathsf{BN}}}
\newcommand{\red}[1]{\textcolor{red}{#1}}
\newcommand{\blue}[1]{\textcolor{blue}{#1}}

\newcommand{\R}{\ensuremath{\raisebox{\depth}{\rotatebox{180}{R}}}}
\newcommand{\framework}{\ensuremath{\mathsf{FVGM}}}

\newcommand{\parent}{\ensuremath{\mathrm{Pa}}}
\newcommand{\nonsensitive}{\ensuremath{\mathbf{X}}}
\newcommand{\sensitive}{\ensuremath{\mathbf{A}}}
\newcommand{\mediator}{\ensuremath{\mathbf{Z}}}
\newcommand{\stochastic}{\ensuremath{\mathsf{S3P}}}
\newcommand{\bool}{\ensuremath{B}}
\newcommand{\alg}{\mathcal{M}}
\newcommand{\graph}{\ensuremath{G}}
\newcommand{\factors}{\ensuremath{\theta}}

\definecolor{terminate}{gray}{0.8}
\definecolor{collision}{rgb}{0.2,1,0.3}
\definecolor{existential}{rgb}{0.9,0.9, 0.1}

\DeclareMathOperator*{\argmax}{arg\,max}
\DeclareMathOperator*{\argmin}{arg\,min}

\title{Algorithmic Fairness Verification with Graphical Models\thanks{Source code: https://github.com/meelgroup/justicia}}

\author{\name Bishwamittra Ghosh \\
	\addr School of Computing\\
	National University of Singapore\\
	Singapore
	\AND
	\name Debabrota Basu \\
	\addr \'Equipe Scool, Univ. Lille, Inria, UMR 9189 - CRIStAL, CNRS\\
	Centrale Lille, France 
	\AND
	\name Kuldeep S. Meel  \\
	\addr School of Computing\\
	National University of Singapore\\
	Singapore}

\begin{document}

\maketitle

\graphicspath{{../}}
\begin{abstract}
In recent years, machine learning (ML) algorithms have been deployed in safety-critical and high-stake decision-making, where the \textit{fairness} of algorithms is of paramount importance. Fairness in ML  centers on detecting bias towards certain demographic populations induced by an ML classifier and proposes algorithmic solutions to mitigate the bias with respect to different fairness definitions.  To this end, several \textit{fairness verifiers} have been proposed that compute the bias in the prediction of an ML classifier\textemdash essentially beyond a finite dataset\textemdash given the probability distribution of input features. In the context of verifying \textit{linear classifiers}, existing fairness verifiers are limited by \emph{accuracy} due to imprecise modeling of correlations among features and \emph{scalability} due to restrictive formulations of the classifiers as SSAT/SMT formulas or by sampling. 

In this paper, we propose an efficient fairness verifier, called \framework, that encodes the correlations among features as a Bayesian network. In contrast to existing verifiers, \framework~proposes a \textit{stochastic subset-sum} based approach for verifying linear classifiers. Experimentally, we show that  \framework~leads to an \textit{accurate} and \textit{scalable} assessment for more diverse families of fairness-enhancing algorithms, fairness attacks, and group/causal fairness metrics than the state-of-the-art fairness verifiers. We also demonstrate that {\framework} facilitates the computation of fairness influence functions as a stepping stone to detect the source of bias induced by subsets of features.
\end{abstract}


\section{Introduction}
\begin{figure}[h!]
	\begin{center}
		\includegraphics[scale = 0.5]{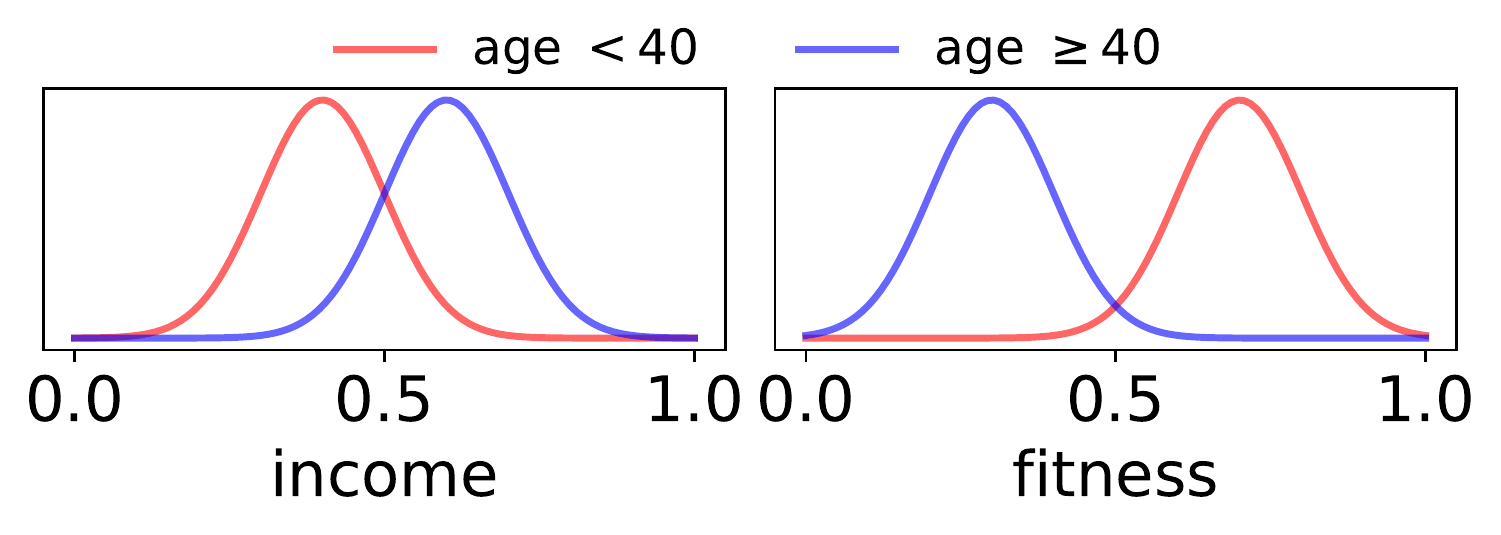}
	\end{center}
	\caption{Age-dependent distributions of income and fitness in Example~\ref{example:intro}.}\label{fig:example1}
\end{figure}

	The significant improvement of machine learning (ML) over the decades has led to a host of applications of ML in  high-stake decision-making such as college admission~\cite{martinez2021using}, hiring of employees~\cite{ajunwa2016hiring}, and recidivism prediction~\cite{tollenaar2013method,dressel2018accuracy}. ML algorithms often have an accuracy-centric learning objective, which may cause them to be biased towards certain part of the dataset belonging to a certain economically or socially sensitive groups~\cite{landy1978correlates,zliobaite2015relation,berk2019accuracy}.
	 The following example illustrates a plausible case of unfairness induced by ML algorithms. 
	\begin{example}\label{example:intro}
		Following~\citep[Example 1.]{ghosh2020justicia}, let us consider an ML problem where the classifier decides the eligibility of an individual for health insurance given their income and fitness (Figure~\ref{fig:example1}). Here, the sensitive feature `age' ($ A $) follows a Bernoulli distribution, and income ($ I $) and fitness ($ F $) follow Gaussian distributions. We generate 1000 samples from these distributions and use them to train a Support Vector Machine (SVM) classifier. The decision boundary of this classifier is $9.37I + 9.75F - 0.34A \ge 9.4$, where $A = 1$ denotes the sensitive group `age $ \ge $ 40'. This classifier selects an individual above and below $40$ years of age with probabilities $0.24$ and $0.86$, respectively. This illustrates a disparate treatment of individuals of two age groups by the SVM classifier. 
	\end{example}

	In order to identify and mitigate the bias of ML classifiers, different fairness definitions and fairness algorithms have been proposed~\cite{hardt2016equality,kusner2017counterfactual,mehrabi2019survey}.	In this paper, we focus on two families of fairness definitions: \textit{group} and \textit{causal} fairness. Group fairness metrics, such as disparate impact and equalized odds constrain the probability of the positive prediction of the classifier to be (almost) equal among different sensitive groups~\cite{dwork2012fairness,feldman2015certifying}.	On the other hand, causal fairness metrics assess the difference in positive predictions  if every feature in the causal relation  remains identical except the sensitive feature~\cite{nabi2018fair,zhang2018fairness}. The early works on \textit{fairness verification} focused on measuring fairness metrics of a classifier for a given dataset~\cite{aif360-oct-2018}. Naturally, such techniques were limited in enhancing confidence of users for wide deployment. Consequently, recent verifiers seek to achieve verification beyond  finite dataset and in turn focus on the  probability distribution of features~\cite{albarghouthi2017fairsquare, bastani2019probabilistic, ghosh2020justicia}.  More specifically, the input to the verifier is a classifier and  the probability distribution of features, and the output is an estimate of fairness metrics that the classifier obtains given the distribution. For Example~\ref{example:intro}, a fairness verifier takes the SVM classifier and the distribution of features $ I, F, A $ as an input and outputs the probability of positive prediction of the classifier for  different sensitive groups. 
	

	In order to solve the fairness-verification problem, existing works have proposed two principled approaches.	Firstly, \citet{ghosh2020justicia} and \citet{albarghouthi2017fairsquare} propose formal methods that reduce the problem into a solution of an SSAT or an SMT~ formula respectively.	Secondly, \citet{bastani2019probabilistic} propose a sampling approach that relies on extensively enumerating the conditional probabilities of prediction given different sensitive features and thus, incurs high computational cost. Additionally, existing works assume feature independence of non-sensitive features and consider correlated features within a limited scope, such as conditional probabilities of non-sensitive features w.r.t. sensitive features and ignore correlations among non-sensitive features. As a result, the \textit{scalability} and \textit{accuracy} of existing  verifiers remains a major challenge.
	
	 In this work, we seek to remedy the aforementioned situation. As a first step, we focus on  \textit{linear classifiers}, which has attracted significant attention from researchers in the context of fair algorithms~\cite{pleiss2017fairness,zafar2017fairness,dressel2018accuracy, john2020verifying}. At this point, it is worth highlighting that our empirical evaluation demonstrates that the existing techniques fail to scale beyond small examples or provide highly inaccurate estimates for comparatively {\em small} linear classifiers. 
	



%
%

	\textbf{Our Contributions.} In this paper, we propose a fairness verification framework, namely {\framework} (\textbf{F}airness \textbf{V}erification with \textbf{G}raphical \textbf{M}odels), for accurately and efficiently verifying linear classifiers. {\framework} proposes a novel \textit{stochastic subset-sum} encoding for linear classifiers with an efficient pseudo-polynomial solution using dynamic programming. To address feature-correlations, {\framework} considers a graphical model, particularly a Bayesian Network that represents conditional dependence (and independence) among features in the form of a Directed Acyclic Graph (DAG).  
	Experimentally,  {\framework} is more accurate and scalable than existing fairness verifiers; {\framework} can verify group and causal fairness metrics for multiple fairness algorithms. We also demonstrate two novel applications of {\framework} as a fairness verifier: (a) detecting fairness attacks, and (b) computing Fairness Influence Functions (FIF) of features as a mean of identifying (un)fairness contribution of a subset of features.

	\section{Background}
	In this section, we define different fairness metrics proposed for classification. Following that, we state basics of stochastic subset sum and Bayesian networks that are the main components of our methodology.
	
	\textbf{Fairness in ML.}
	We consider\footnote{{We represent sets/vectors by bold letters, and the corresponding distributions by calligraphic letters. We express random variables in uppercase, and an assignment of a random variable in lowercase.}} a dataset $ \mathbf{D} $ as a collection of triples $ (\nonsensitive, \sensitive, Y) $ generated from an underlying distribution $\mathcal{D}$. $ \nonsensitive \triangleq \{X_1, \dots, X_{m_1}\} $ are non-sensitive features whereas $ \sensitive \triangleq \{A_1, \dots, A_{m_2}\} $ are categorical sensitive features.  $Y \in \{0,1\}$ is the binary label (or class) of $(\nonsensitive,\sensitive)$. Each non-sensitive feature $ X_i$ is sampled from a continuous probability distribution {$ \mathcal{X}_i $}, and each sensitive feature $ A_j \in \{0, \dots, N_j\}  $ is sampled from a discrete probability distribution {$ \mathcal{A}_j $}. 
 	We use $ (\mathbf{x}, \mathbf{a}) $ to denote the feature-values of  $ (\nonsensitive, \sensitive) $.  For sensitive features, a valuation vector $ \mathbf{a} = [a_1, .., a_{m_2}] $ is called a \textit{compound sensitive group}. For example, consider $ \sensitive = $ \{race, sex\} where race $ \in $ \{Asian, Color, White\} and sex $ \in $ \{female, male\}. Thus $ \mathbf{a} = $ [Asian, female]  is a compound sensitive group. 
 	We represent a binary classifier trained on the dataset $\mathbf{D}$ as $\alg: (\nonsensitive, \sensitive) \rightarrow \hat{Y} $. Here, $\hat{Y} \in \{0,1\}$ is the predicted class of $ (\nonsensitive, \sensitive) $.
 	
 	We now discuss different fairness metrics in the literature based on the prediction of a classifier~\citep{feldman2015certifying,hardt2016equality,nabi2018fair}.  	
 	In this paper, {\framework} verifies two families of fairness metrics: group fairness (first three in the following) and path-specific causal fairness.

 	\begin{enumerate}[leftmargin=*]
 		\itemsep0em 
 		\item \textit{Disparate Impact} (DI): A classifier  satisfies $(1 - \epsilon)$-disparate impact if for $\epsilon \in [0,1] $,
 		$
 		\min_{\mathbf{a}} \Pr[\hat{Y} =1 | \mathbf{A} =  \mathbf{a}]  \ge (1 - \epsilon) \max_{\mathbf{a}} \Pr[\hat{Y} =1 | \mathbf{A} =  \mathbf{a}].
 		$
 		\item \textit{Statistical Parity} (SP): A classifier satisfies $\epsilon$-statistical parity if for $\epsilon \in [0,1] $, 
 		$
 		\max_{\mathbf{a}} \Pr[\hat{Y} =1 | \mathbf{A} =  \mathbf{a}] - \min_{\mathbf{a}} \Pr[\hat{Y} =1 | \mathbf{A} =  \mathbf{a}] \le \epsilon.
 		$
 		\item \textit{Equalized Odds} (EO): 	A classifier satisfies $\epsilon$-equalized odds if for $\epsilon \in [0,1] $,
 		$ \max_{\mathbf{a}}\Pr[\hat{Y} =1 |\mathbf{A}= \mathbf{a}, Y= 0  ] - \min_{\mathbf{a}}\Pr [\hat{Y} = 1|\mathbf{A}= \mathbf{a}, Y = 0] \le \epsilon, $ and $
 		\max_{\mathbf{a}}\Pr[\hat{Y} =1 |\mathbf{A}= \mathbf{a}, Y= 1  ] - \min_{\mathbf{a}}\Pr [\hat{Y} = 1|\mathbf{A}= \mathbf{a}, Y = 1] \le \epsilon.
 		$
 		\item \textit{Path-specific Causal Fairness} (PCF): 
 		Let $ \mathbf{a}_{\max}  \triangleq \argmax_{ \mathbf{a}} \Pr[\hat{Y} =1 |\mathbf{A}=  \mathbf{a}] $. We consider mediator features $ \mediator \subseteq \nonsensitive $ sampled from the conditional distribution $ {\mathcal{Z}_{|\mathbf{A} = \mathbf{a}_{\max}}} $. This emulates the fact that mediator variables have the same sensitive features $ \mathbf{a}_{\max} $.  For $ \epsilon \in [0,1] $,  path-specific causal fairness is defined as 
 		$
 		\max_{\mathbf{a}}\Pr[\hat{Y} = 1 | \sensitive =  \mathbf{a}, \mediator] - \min_{\mathbf{a}}\Pr[\hat{Y} = 1 | \sensitive = \mathbf{a}, \mediator ] \le \epsilon
 		$.
 	\end{enumerate}

 	  For all of the above metrics, lower value of $\epsilon$ indicates higher fairness demonstrated by the classifier $\mathcal{M}$. Following the observation of~\cite{ghosh2020justicia},  computing all of the aforementioned fairness metrics is equivalent to computing the maximum and minimum of the \textit{probability of positive prediction} of the classifier, denoted as $\Pr[\hat{Y}=1|\mathbf{A} =\mathbf{a}]$, for all compound sensitive groups $\mathbf{a}$ from $ \sensitive $. Thus, in Section~\ref{sec:fvgm}, we focus on computing the maximum and minimum probability of positive prediction of the classifier and then extend it to assess corresponding fairness metrics. We call the group for which the probability of positive prediction is maximum (minimum) as the \textit{most (least) favored group} of the classifier.

	\textbf{Stochastic Subset Sum Problem ({\stochastic}).} 
	Let $ \mathbf{B} \triangleq \{B_i\}_{i=1}^{|\mathbf{B}|}$ be a set of Boolean variables and $ w_i \in \mathbb{Z} $ be the weight of $ B_i $. Given a constraint of the form  $\sum_{i = 1}^ {|\mathbf{B}|} w_i B_i = \tau $, for a constant threshold $ \tau \in \mathbb{Z} $, the subset-sum problem seeks to compute an assignment $\mathbf{b} \in \{0,1\}^{|\mathbf{B}|}$ such that the constraint evaluates to true when $\mathbf{B}$ is substituted with $\mathbf{b}$. Subset sum problem is known to be a $ \mathrm{NP} $-complete problem and well-studied in theoretical computer science~\cite{kleinberg2006algorithm}. The \textit{counting} version of the subset-sum problem counts all $ \mathbf{b} $'s for which the above constraint holds; and this problem belongs to the complexity class $ \mathrm{\#P} $. In this paper, we consider the counting problem for the constraint $\sum_{i = 1}^ {|\mathbf{B}|} w_i B_i \ge \tau $ where variables $ B_i $'s are stochastic. More precisely, we define a \textit{stochastic subset-sum} problem, namely {\stochastic}, that computes $ \Pr[\sum_{i = 1}^ {|\mathbf{B}|} w_iB_i \ge \tau] $.    Details of {\stochastic} are in Section~\ref{sec:stochastic_sum_set_sum}.

	\textbf{Bayesian Network.}
	In general, a Probabilistic Graphical Model~\citep{koller2009probabilistic}, and specifically a \textit{Bayesian network}~\cite{pearl1985bayesian,chavira2008probabilistic}, encodes the dependencies and conditional independence between a set of random variables. In this paper, we leverage an access to a Bayesian network on $ \nonsensitive \cup \sensitive $ that represents the joint distribution on them. 	A Bayesian network is denoted by a pair $ (\graph, \theta)$, where $ \graph \triangleq (\mathbf{V}, \mathbf{E}) $ is a DAG (Directed Acyclic Graph), and $\theta$ is a set of parameters encoding the conditional probabilities induced by the joint distribution under investigation. Each vertex $V_i \in \mathbf{V}$ corresponds to a random variable. Edges $ \mathbf{E} \in \mathbf{V} \times \mathbf{V} $ imply conditional dependencies among variables. For each variable $ V_i \in \mathbf{V} $, let $ \parent(V_i) \subseteq \mathbf{V} \setminus \{V_i\} $ denote the set of parents of $ V_i $. Given $\parent(V_i)$ and parameters $\theta$, $ V_i $ is independent of its other non-descendant variables in $\graph$. Thus, for the assignment $ v_i $ of $ V_i $ and $ \mathbf{u} $ of $ \parent(V_i) $, the aforementioned semantics of a Bayesian network encodes the joint distribution of $\mathbf{V}$ as:
	
	\begin{equation}
		\begin{split}
				\Pr[V_1=v_1, \dots, &V_{|\mathbf{V}|}=v_{|\mathbf{V}|}] = \prod_{i=1}^{|\mathbf{V}|} \Pr[V_i = v_i | \parent(V_i) = \mathbf{u}; \theta].
		\end{split}
	\label{eq:BN}
	\end{equation}

\section{{\framework}: Fairness Verification with Graphical Models}\label{sec:fvgm}

In this section, we present {\framework}, a fairness verification framework for linear classifiers that accounts for correlated features represented as a graphical model. The core idea of verifying fairness of a classifier is to compute the probability of positive prediction of the classifier with respect to all compound sensitive groups. To this end, {\framework} solves a stochastic subset sum problem, {\stochastic}, that is equivalent to computing the probability of positive prediction of the classifier for the most (and the least) favored sensitive group. In this section, we first define {\stochastic} and present an efficient dynamic programming solution for {\stochastic}. We then extend {\stochastic} to consider correlated features as input. Finally, we conclude by discussing fairness verification based on the solution of {\stochastic}.

\textbf{Problem Formulation.}	
Given a linear classifier $ \alg: (\nonsensitive, \sensitive) \rightarrow \hat{Y} $ and a probability distribution $ \mathcal{D} $ of $ \nonsensitive \cup \sensitive $, our objective is to compute $ \max_{ \mathbf{a}} \Pr[\hat{Y} =1 | \sensitive = \mathbf{a}] $ and $ \min_{ \mathbf{a}} \Pr[\hat{Y} =1 | \sensitive = \mathbf{a}] $ with respect to $ \mathcal{D} $. In this study, we express a linear classifier $\mathcal{M}$ as 

\[	\hat{Y} = \mathds{1}\Big[\sum_{i} w_{X_i}X_i + \sum_{j} w_{A_j}A_j \ge \tau\Big].\]

Here, $ w $ denotes the weight (or coefficients) of a feature, $ \tau $ denotes the bias or the offset parameter of the classifier, and $\mathds{1}$ is an indicator function. Hence, the prediction $ \hat{Y} =1 $ if and only if the inner inequality holds.
Thus, computing the maximum (resp.\ minimum) probability of positive prediction is equivalent to finding out the assignment of $A_j$'s for which the probability of satisfying the inner inequality is highest (resp.\ lowest). We reduce this computation into an instance of {\stochastic}. To perform this reduction, we assume  weights $ w $ and bias $ \tau $ as integers, and features $\nonsensitive \cup \sensitive $ as Boolean. In Sec.~\ref{sec:practical}, we relax these assumptions and extend to the practical settings. 

\subsection{$ \stochastic $: Stochastic Subset Sum Problem}\label{sec:stochastic_sum_set_sum}
Now, we formally describe the specification and semantics of {\stochastic}.
{\stochastic} operates on  a set of Boolean variables $ \mathbf{B} = \{B_i\}_{i=1}^{n} \in \{0,1\}^{n} $, where $ w_i \in \mathbb{Z} $ is the weight of $ B_i $, and $n \triangleq |\mathbf{B}|$. Given a constant threshold $ \tau \in \mathbb{Z} $, {\stochastic} computes the \textit{probability} of a subset of $ \mathbf{B} $ with sum of weights of non-zero variables to be at least $ \tau $. Formally,

\[S(\mathbf{B}, \tau) \triangleq \Pr\Big[\sum_{i} w_iB_i \ge \tau \Big] \in [0,1].\]

Aligning with terminologies in stochastic satisfiability (SSAT)~\cite{littman2001stochastic}, we categorize the variables $ \mathbf{B} $ into two types: (i) \textit{chance variables} that are stochastic and have an associated probability of being assigned to $ 1 $ and (ii) \textit{choice variables} that we  optimize while computing $ S(\mathbf{B}, \tau) $.  To specify the category of variables, we consider a \textit{quantifier} $ q_i \in \{\R^{p_i}, \exists, \forall\} $ for each $ B_i $. Elaborately, $ \R^{p} $ is a \textit{random quantifier} corresponding to a chance variable $ B \in \mathbf{B} $, where  $ p\triangleq \Pr[B = 1]$. In contrast, $ \exists $ is an \textit{existential quantifier} corresponding to a choice variable $ B $ such that a Boolean assignment of $ B $  \textit{maximizes}  $ S(\mathbf{B}, \tau) $. Finally, $ \forall $ is an \textit{universal quantifier} for a choice variable $ B $ that fixes an assignment to $ B $ that \textit{minimizes} $ S(\mathbf{B}, \tau) $. 
 
Now, we formally present the semantics of $ S(\mathbf{B}, \tau) $ provided that each variable $ B_i $ has weight $ w_i $ and quantifier $ q_i $. Let  $ \mathbf{B}[2:n] \triangleq \{B_j\}_{j=2}^{n} $ be the subset of $\mathbf{B}$ without the first variable $ B_1 $. Then $ S(\mathbf{B}, \tau) $ is recursively defined as:

\begin{align}\label{eq:semantics_recurse}
  S(\mathbf{B},\tau) =
 \begin{cases}
 \mathds{1}[ \tau \le 0 ], \; \text{if } \textbf{B} = \emptyset \\
 S(\mathbf{B}[2:n],\tau - \max\{w_1, 0\}), \; \text{if } q_1 = \exists\\
 S(\mathbf{B}[2:n],\tau - \min\{w_1, 0\}), \; \text{if } q_1 = \forall\\
 p_1\times S(\mathbf{B}[2:n],\tau - w_1) +\\ (1-p_1)\times S(\mathbf{B}[2:n],\tau),\; \text{if } q_1 = \R^{p_1}
 \end{cases}
\end{align}


Observe that when $ \textbf{B} $ is empty, $S$ is computed as $ 1 $ if $ \tau \le 0 $, and  $ S = 0 $ otherwise. For  existential and universal quantifiers, we compute $ S $ based on the weight. Specifically, if $ q_1 = \exists $, we decrement the threshold $ \tau $ by the maximum between $ w_1 $ and $ 0 $. For example, if $ w_1 > 0 $, $ B_1 $ is assigned $ 1 $, and assigned $ 0 $ otherwise. Therefore, by solving for an existential variable, we  maximize $ S $. In contrast, when if $ q_1 = \forall $, we fix an assignment of $ B_1 $ that minimizes $ S $ by choosing between the minimum of $ w_1 $ and $ 0 $. Finally, for random quantifiers, we decompose the computation of $S$ into two sub-problems: one sub-problem where $ B_1 = 1 $ and the updated threshold becomes $ \tau - w_1 $ and another sub-problem where $ B_1 = 0 $ and the updated threshold remains the same. Herein, we compute $ S $ as the expected output of both sub-problems.

\begin{remark}
	\label{lm:property_s3p}
	$ S(\mathbf{B},\tau) $ does not depend on the order of $ \mathbf{B} $.
\end{remark}

\textbf{Computing Minimum and Maximum probability of positive prediction of Linear Classifiers Using  {\stochastic}.} For computing $ \max_{ \mathbf{a}} \Pr[\hat{Y} =1 | \sensitive = \mathbf{a}] $ of a linear classifier, we set existential quantifiers $ \exists $ to sensitive features $ A_j $, randomized quantifiers $ \R $ to non-sensitive features $ X_i $ and construct a set $ \mathbf{B} = \sensitive \cup \nonsensitive $.  The coefficients $ w_{A_j} $ and $ w_{X_i} $ of the classifier become weights of $ \mathbf{B} $. Also, we get $n=m_1 +m_2$. For non-sensitive variables $ X_i $, which are chance variables, we derive their marginal probability $ p_i = \Pr[X_i = 1] $ from the distribution $ \mathcal{D} $.  According to semantic of {\stochastic}, setting $ \exists $ quantifiers on $ \sensitive $ computes the maximum value of $ S(\mathbf{B}, \tau) $ that equalizes the maximum probability of positive prediction of the classifier. In this case, the \textit{inferred} assignment of $ \sensitive $ implies the most favored group $ \mathbf{a}_{{\max}} =  \argmax_{ \mathbf{a}} \Pr[\hat{Y} =1 | \sensitive = \mathbf{a}] $. In contrast, to compute the minimum probability of positive prediction, we instead assign variable  $ A_j $ a universal  quantifier  while keeping random quantifiers over $ X_i $, and infer the least favored group $ \mathbf{a}_{{\min}} =  \argmin_{ \mathbf{a}} \Pr[\hat{Y} =1 | \sensitive = \mathbf{a}] $.

 \subsection{A Dynamic Programming Solution for {\stochastic}}
 \label{sec:dp_formulation}

We propose a dynamic programming approach~\cite{pisinger1999linear,woeginger1992equal} to solve {\stochastic} as the problem has overlapping sub-problem properties. 
For example, $S(\mathbf{B}, \tau)$ can be solved by solving $S(\mathbf{B}[2:n], \tau')$, where the updated threshold $\tau'$, called the \textit{residual threshold}, depends on the original threshold $\tau$ and the assignment of $B_1$ as shown in Eq.~\eqref{eq:semantics_recurse}.
Building on this observation, we propose the recursion and terminating condition leading to our dynamic programming algorithm. 

\textit{Recursion.} We consider a function $ \mathsf{dp}(i, \tau) $ that solves the sub-problem $ S(\mathbf{B}[i:n],\tau)  $, for $ i \in \{1,\ldots,n\}  $. The semantics of  $ S(\mathbf{B},\tau) $ in Eq.~\eqref{eq:semantics_recurse} induces the recursive definition of $ \mathsf{dp}(i,\tau) $ as: 

\begin{align}\label{eq:dp_recurse}
 \mathsf{dp}(i,\tau)=
 \begin{cases}
 \mathsf{dp}(i+1, \tau-\max\{w_i, 0\}), \; \text{if } q_i = \exists\\
 \mathsf{dp}(i+1, \tau-\min\{w_i, 0\}), \; \text{if } q_i = \forall\\
 p_i \times \mathsf{dp}(i+1, \tau-w_i) + \\ (1- p_i) \times \mathsf{dp}(i+1, \tau) ,\; \text{if } q_i = \R^{p_i}
 \end{cases}
\end{align} 

Eq.~\eqref{eq:dp_recurse} shows that $ S(\mathbf{B},\tau) $ can be solved by instantiating $ \mathbf{dp}(1, \tau) $, which includes all the variables in $ \mathbf{B} $. 

\textit{Terminating Condition.} 
Let $ w_{neg} $, $ w_{pos} $, and $ w_{all} $ be the sum of negative, positive, and all weights of $ \mathbf{B} $, respectively. We observe that $ w_{neg} \le w_{all} \le w_{pos}$. Thus, for any $ i $, if the {residual} threshold $ \tau \le w_{neg}$, there is always a subset of $ \mathbf{B}[i:n] $ with sum of weights at least $ \tau $. Conversely, when $ \tau > w_{pos}$, there is no subset of $ \mathbf{B}[i:n] $ with sum of weights at least $ \tau $.	We leverage this bound and tighten the terminating conditions of $ \mathsf{dp}(i, \tau) $ in Eq.~\eqref{eq:dp_terminus}. 

\begin{align}\label{eq:dp_terminus}
 \mathsf{dp}(i, \tau) =
 \begin{cases}
 1\text{ if }\tau \le w_{neg}\\
 0\text{ if } \tau > w_{pos}\\
 \mathds{1}[\tau \le 0] \text{ if } i=n + 1
 \end{cases} 
\end{align}

Eq.~\eqref{eq:dp_recurse} and~\eqref{eq:dp_terminus} together define our dynamic programming algorithm. While deploying the algorithm, we store $ \mathsf{dp}(i, \tau) $ in memory to avoid repetitive computations.  This allows us to achieve a pseudo-polynomial algorithm (Lemma~\ref{lemma:complexity_sss}) instead of a na\"ive exponential algorithm enumerating all possible assignments. In particular, the time complexity is pseudo-polynomial for chance (random) variables and linear for choice (existential and universal) variables.
 
\begin{lemma}\label{lemma:complexity_sss}
 	Let $ n' $ be the number of existential and universal variables in $ \mathbf{B} $. Let $ w_{\exists} = \sum_{B_i \in \mathbf{B} | q_i = \exists} \max\{w_i, 0\}$  and $ w_{\forall} = \sum_{B_i \in \mathbf{B} | q_i = \forall} \min\{w_i, 0\}$ be the considered sum of weights of existential and universal variables, respectively. We can exactly solve {\stochastic} using dynamic programming with time complexity $ \mathcal{O}((n - n')(\tau + |w_{neg}| - w_{\exists} - w_{\forall}) + n') $. The space complexity is  $ \mathcal{O}((n - n')(\tau + |w_{neg}| - w_{\exists} - w_{\forall})) $.
\end{lemma}

\textbf{A Heuristic for Faster Computation.} 
We propose two improvements for a faster computation of the dynamic programming solution. Firstly, we observe that in Eq.~\eqref{eq:dp_recurse}, existential/universal variables are deterministically  assigned based on their weights. Hence, we reorder $ \mathbf{B} $ such that existential/universal variables appear earlier in $ \mathbf{B} $ than random variables. This allows us to avoid unnecessary repeated exploration of existential/universal variables in $ \mathsf{dp} $. Moreover, according to the remark in Section~\ref{lm:property_s3p}, reordering $ \mathbf{B} $ still produces the same exact solution of {\stochastic}. Secondly, to reach the terminating condition of $ \mathsf{dp}(i, \tau) $ more frequently, we sort $ \mathbf{B} $ based on their weights\textemdash more specifically, within each cluster of random, existential, and universal variables. In particular, if $ \tau \le 0.5(w_{pos} - w_{neg}) $, $ \tau $ is closer to $ w_{pos} $ than $ w_{neg} $. Hence, we sort each cluster in descending order of weights. Otherwise,  we sort in ascending order. We illustrate our dynamic programming approach in Example~\ref{example:subset-sum}.
	
\begin{figure*}[t!]
	\centering

	\subfloat[Known marginal  probabilities.]{
		\scalebox{0.73}{
		\begin{tikzpicture}[>=stealth',shorten >=1pt, on grid,initial/.style={}]
			\node[state, align=center, fill=existential] (T1) {$ 1, 2 $\\$ 0.55 $};
			\node[state, align=center, label={[align=left,right]0:$ 0.85\Pr[Q] + 0.35\Pr[\neg Q] $\\$ =0.55 $}] (T2) [below = 1.6 cm of T1] {$2,1$\\$ 0.55 $};
			\node[state, align=center] (T3) [below left = 1.5 cm and 1.8 cm of T2] {$ 3,0 $\\$ 0.85$};
			\node[state, align=center, fill=terminate] (T4) [below left = 1.5 cm and 1 cm of T3] {$ 4, -1$\\$ 1 $};
			\node[state, align=center] (T5) [below right = 1.5 cm and 1 cm of T3] {$ 4, 0$\\$ 0.7 $};
			\node[state, align=center, fill=terminate] (T6) [below left = 1.5 cm and 1 cm of T5] {$ 5,1 $\\ $0 $};
			\node[state, align=center, fill=terminate] (T7) [below right = 1.5 cm and 1 cm of T5] {$ 5,0 $\\$ 1 $};
			\node[state, align=center] (T8) [below right = 1.5 cm and 1.6 cm of T2] {$ 3,1 $\\$ 0.35 $};
			\node[state, align=center, fill=collision] (T9) [below left = 1.5 cm and 1 cm of T8] {$ 4, 0$\\$ 0.7$};
			\node[state, align=center] (T10) [below right = 1.5 cm and 1 cm of T8] {$ 4, 1$\\$ 0 $};
			\node[state, align=center, fill=terminate] (T11) [below left = 1.5 cm and 1 cm of T10] {$ 5, 2$\\$ 0 $};
			\node[state, align=center, fill=terminate] (T12) [below right = 1.5 cm and 1 cm of T10] {$ 5,1 $\\$ 0 $};

			\tikzset{every node/.style={fill=white}}
			\path (T1) edge [right] node {$P$}  (T2);
			\path (T2) edge [left] node {$Q$}  (T3);
			\path (T3) edge [left] node {$R$}  (T4);
			\path (T3) edge [right] node {$\neg R$}  (T5);
			\path (T5) edge [left] node {$S$}  (T6);
			\path (T5) edge [right] node {$\neg S$}  (T7);
			\path (T2) edge [right] node {$\neg Q$}  (T8);
			\path (T8) edge [left] node {$R$}  (T9);
			\path (T8) edge [left] node {$\neg R$}  (T10);
			\path (T10) edge [left] node {$S$}  (T11);
			\path (T10) edge [left] node {$\neg S$}  (T12);
			\end{tikzpicture}
		}
\label{fig:example_dp}}
\subfloat[Probabilities computed with a Bayesian network.]
{  
		\scalebox{0.73}{
			\begin{tikzpicture}[>=stealth',shorten >=1pt, on grid,initial/.style={}]
			\node[state, align=center, fill=existential, accepting] (T1) {$ 1, 2 $\\$ 0.65 $};
			\node[state, align=center, accepting, label={[align=left,right]0:$ 0.85\Pr[Q|P] + $ \\ $   0.35\Pr[\neg Q|P] $\\$ =0.65 $}] (T2) [below left = 1.6 cm and 2 cm of T1] {$2,1$\\$ 0.65 $};
			\node[state, align=center] (T3) [below left = 1.5 cm and 1.8 cm of T2] {$ 3,0 $\\$ 0.85$};
			\node[state, align=center, fill=terminate] (T4) [below left = 1.5 cm and 1 cm of T3] {$ 4, -1$\\$ 1 $};
			\node[state, align=center] (T5) [below right = 1.5 cm and 1 cm of T3] {$ 4, 0$\\$ 0.7 $};
			\node[state, align=center, fill=terminate] (T6) [below left = 1.5 cm and 1 cm of T5] {$ 5,1 $\\ $0 $};
			\node[state, align=center, fill=terminate] (T7) [below right = 1.5 cm and 1 cm of T5] {$ 5,0 $\\$ 1 $};
			\node[state, align=center] (T8) [below right = 1.5 cm and 1.6 cm of T2] {$ 3,1 $\\$ 0.35 $};
			\node[state, align=center, fill=collision] (T9) [below left = 1.5 cm and 1 cm of T8] {$ 4, 0$\\$ 0.7$};
			\node[state, align=center] (T10) [below right = 1.5 cm and 1 cm of T8] {$ 4, 1$\\$ 0 $};
			\node[state, align=center, fill=terminate] (T11) [below left = 1.5 cm and 1 cm of T10] {$ 5, 2$\\$ 0 $};
			\node[state, align=center, fill=terminate] (T12) [below right = 1.5 cm and 1 cm of T10] {$ 5,1 $\\$ 0 $};

			\node[state, align=center, accepting] (T13) [below right = 1.6 and 2 cm of T1] {$2,2$\\$ 0.11 $};
			\node[state, align=center, fill=collision] (T14) [below left = 1.5 and 1 cm of T13] {$3,1$\\$ 0.35 $};
			\node[state, align=center] (T15) [below right = 1.5 and 1 cm of T13] {$3,2$\\$ 0 $};
			\node[state, align=center, fill=collision] (T16) [below left = 1.5 and 1 cm of T15] {$4,1$\\$ 0 $};
			\node[state, align=center] (T17) [below right = 1.5 and 1 cm of T15] {$4,2$\\$ 0 $};
			\node[state, align=center, fill=terminate] (T18) [below left = 1.5 and 1 cm of T17] {$5,3$\\$ 0 $};
			\node[state, align=center, fill=terminate] (T19) [below right = 1.5 and 1 cm of T17] {$5,2$\\$ 0 $};

			\node[state, align=center] (P) [right = 5 cm of T1] {$P$};
			\node[state, align=center] (Q) [right = 2 cm of T15] {$Q$};

			\tikzset{every node/.style={fill=white}}
			\path (T1) edge [right] node {$P$}  (T2);
			\path (T2) edge [left] node {$Q$}  (T3);
			\path (T3) edge [left] node {$R$}  (T4);
			\path (T3) edge [right] node {$\neg R$}  (T5);
			\path (T5) edge [left] node {$S$}  (T6);
			\path (T5) edge [right] node {$\neg S$}  (T7);
			\path (T2) edge [left] node {$\neg Q$}  (T8);
			\path (T8) edge [left] node {$R$}  (T9);
			\path (T8) edge [left] node {$\neg R$}  (T10);
			\path (T10) edge [left] node {$S$}  (T11);
			\path (T10) edge [left] node {$\neg S$}  (T12);
			\path (T1) edge [right] node {$\neg P$}  (T13);
			\path (T13) edge [right] node {$Q$}  (T14);
			\path (T13) edge [right] node {$\neg Q$}  (T15);
			\path (T15) edge [right] node {$R$}  (T16);
			\path (T15) edge [right] node {$\neg R$}  (T17);
			\path (T17) edge [right] node {$S$}  (T18);
			\path (T17) edge [right] node {$\neg S$}  (T19);

			\path (P) edge [->] node [left=0.1cm] {\parbox{01.5cm}{Bayesian\\network}}(Q);
		\end{tikzpicture}
	}
	\label{fig:example_BN}
}

\caption{Search tree representation of {\stochastic} for computing the maximum probability of positive prediction of the classifier on variables $ \mathbf{B} =  \{P,Q,R,S\} $ with weights $ \{1,1,1,-1\} $ and threshold $ \tau = 2 $ . Each node is labeled by $ (i,\tau') $, where $ i $ is the index of $ \mathbf{B} $ and $ \tau' $ is the residual threshold. The tree is explored using Depth-First Search (DFS) starting with left child. Within a node, the value in the bottom denotes $ \mathsf{dp}(i, \tau') $ that is solved recursively based on sub-problems $ \mathsf{dp}(i+1, \cdot) $ in child nodes. 	Yellow nodes denote \textit{existential} variables and all other nodes are  \textit{random} variables. Additionally, a green node denotes a collision, in which case a previously computed $ \mathsf{dp} $ solution is returned. Leaf nodes (gray) are computed based on terminating conditions in Eq.~\ref{eq:dp_terminus}. In Figure~\ref{fig:example_BN},  nodes with double circles, such as $ \{(1,2), (2,1), (2,2)\} $,  are enumerated exponentially to compute conditional probabilities from the Bayesian network.}
\label{fig:example_tree_exploration}
\end{figure*}
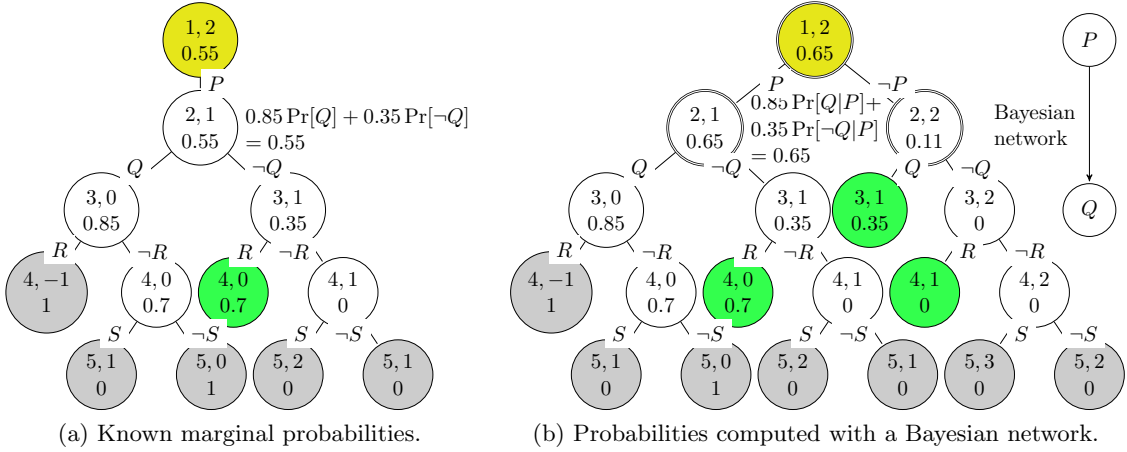

\begin{example}\label{example:subset-sum}
We consider a linear classifier $ P + Q + R - S \ge 2$. Herein, $ P $ is a Boolean sensitive feature, and $ Q, R, S $ are Boolean non-sensitive features with $ \Pr[Q] = 0.4,  \Pr[R] = 0.5 $, and $ \Pr[S] = 0.3 $. To compute the maximum probability of positive prediction of the classifier,  we impose an existential quantifier on $P$ and randomized quantifiers on others. This leads us to the observation that $ P = 1 $ is the optimal assignment as $ w_P = 1 > 0 $. We now require to compute $ \Pr[Q + R - S \ge 1] $, which by dynamic programming, is computed as $ 0.55 $. The solution is visualized as a search tree in Figure~\ref{fig:example_dp}, where we observe that storing the solution of sub-problems in the memory avoids repetitive computation, such as exploring the node $ (4,0) $. Similarly, the minimum probability of positive prediction  of the classifier is $ 0.14 $ (not shown in Figure~\ref{fig:example_dp}) where we impose a universal quantifier on $P$ to obtain $ P = 0 $ as the optimal assignment. 
\end{example}

\subsection{{\stochastic} with Correlated Variables} 
\label{sec:dp_with_BN}
In {\stochastic} presented in Section~\ref{sec:stochastic_sum_set_sum}, we consider all  Boolean variables to be probabilistically independent. This independence assumption often leads to an \textit{inaccurate estimate} of the probability of positive prediction of the classifier because both sensitive and non-sensitive features can be correlated in practical fairness problems. Therefore, we extend {\stochastic} to include correlations among variables.

We consider a Bayesian network $ \BN = (\graph, \theta) $ to represent correlated variables, where $ G \triangleq (\mathbf{V}, \mathbf{E}) $, $ \mathbf{V} \subseteq \mathbf{B} $, $ \mathbf{E} \subseteq \mathbf{V} \times \mathbf{V}  $, and $ \theta $ is the parameter of the network.  In  $ \BN $, we constrain that there is no conditional probability of choice (i.e., existential and universal) variables as we optimize their assignment in {\stochastic}. Choice variables, however, can impose conditions on chance (i.e., random) variables. In practice, we achieve this by allowing no incoming edge on choice variables while learning $ \BN $ (ref. Section~\ref{sec:experiments}).

For a chance variable $ B_i \in \mathbf{V} $, let $ \parent(B_i) $ denote its parents. According to Eq.~\eqref{eq:BN},  for an assignment $ \mathbf{u} $ of $ \parent(B_i) $, $ \BN $ ensures $ B_i $ to be independent of other non-descendant variables in $ \mathbf{V} $. Hence, in the recursion of Eq.~\eqref{eq:dp_recurse}, we substitute  $ p_i $  with  $ \Pr[B_i = 1| \parent(B_i) = \mathbf{u}] $. In order to explicate the dependence on $ \mathbf{u} $, we denote the expected solution of $ S(\mathbf{B}[i:n], \tau) $ as 
$ \mathsf{dp}(i, \tau, \mathbf{u}) $, which for $ B_i \in \mathbf{V} $ is modified as follows:

\begin{align*}
	\mathsf{dp}(i,  &\tau, \mathbf{u}) = \Pr[B_i = 0| \parent(B_i) = \mathbf{u}] \mathsf{dp}(i+1, \tau, \mathbf{u} \cup \{0\})  \\
	& + \Pr[B_i = 1| \parent(B_i) = \mathbf{u}]  \mathsf{dp}(i+1, \tau-w_i, \mathbf{u} \cup \{1\}).
\end{align*}

Since $ \mathsf{dp}(i, \tau, \mathbf{u}) $ involves  $ \mathbf{u} $, we initially perform a topological sort of $ \mathbf{V} $ to know the assignment of parents before computing $ \mathsf{dp} $ on the child. Moreover, there are $ 2^{|\parent(B_i)|} $ assignments of $ \parent(B_i) $, and we compute $ \mathsf{dp}(i, \tau, \mathbf{u}) $ for $ \mathbf{u} \in \{0,1\}^{|\parent(B_i)|} $ to incorporate all conditional probabilities into $ \stochastic $.  For this enumeration, we do not store $ \mathsf{dp}(i, \tau, \mathbf{u}) $ in memory. However, for $ B_i \not \in \mathbf{V} $ that does not appear in the network, we instead compute $ \mathsf{dp}(i, \tau) $ and store it in memory as in Section~\ref{sec:dp_formulation}, because $ B_i $ is not correlated with other variables.  Lemma~\ref{lm:complexity_dp_with_bn} presents the complexity of solving {\stochastic} with correlated variables, wherein unlike Lemma~\ref{lemma:complexity_sss}, the  complexity differentiates based on variables in $ \mathbf{V} $ (exponential) and $ \mathbf{B}\setminus \mathbf{V} $ (pseudo-polynomial).

\begin{lemma}
	\label{lm:complexity_dp_with_bn}
	Let $ \mathbf{V} \subseteq \mathbf{B} $ be the set of vertices in the Bayesian network and $ n'' $ be the number of existential and universal variables in $ \mathbf{B} \setminus \mathbf{V} $. Let $ w'_{\exists} = \sum_{B_i \in \mathbf{B} \setminus \mathbf{V} | q_i = \exists} \max\{w_i, 0\}$  and $ w'_{\forall} = \sum_{B_i \in \mathbf{B} \setminus \mathbf{V} | q_i = \forall} \min\{w_i, 0\}$ be the sum of considered weights of existential and universal variables, respectively that only appear in $ \mathbf{B} \setminus \mathbf{V} $. To exactly compute {\stochastic} with correlated variables in the dynamic programming approach,  time complexity is $ \mathcal{O}(2^{|\mathbf{V}|} + (n - n'' - |\mathbf{V}|)(\tau + |w_{neg}| - w'_{\exists} - w'_{\forall}) + n'') $ and space complexity is $ \mathcal{O}((n - n'' - |\mathbf{V}|)(\tau + |w_{neg}| - w'_{\exists} - w'_{\forall})) $.
\end{lemma}	

\textbf{A Heuristic for Faster Computation.} We observe that to encode conditional probabilities, we enumerate all assignments of variables in $ \mathbf{V} $. For computing the probability of positive prediction of a linear classifier with correlated features, we consider a heuristic to sort variables in $ \mathbf{B} = \sensitive \cup \nonsensitive $. Let $ \mathbf{V} \subseteq \mathbf{B} $ be the set of vertices in the network and $ \mathbf{V}^c = \mathbf{B} \setminus \mathbf{V} $. In this heuristic, we sort sensitive variables $ \sensitive $ by positioning $ \sensitive \cap \mathbf{V} $ in the beginning followed by $ \sensitive \cap \mathbf{V}^c $. Then we order the variables $ \mathbf{B} $ such that variables in $ \sensitive $ precedes those in $ \nonsensitive \cap \mathbf{V} $, and the variables in $ \nonsensitive \cap \mathbf{V}^c $ follows the ones in $ \nonsensitive \cap \mathbf{V} $. This sorting allows to avoid repetitive enumeration of variables in $ \mathbf{V} \subseteq \mathbf{B} $ as they are placed earlier in $ \mathbf{B} $.

	\begin{example}
		We extend Example~\ref{example:subset-sum} with a Bayesian Network $ (\graph, \theta) $ with $ \mathbf{V} = \{P, Q\} $ and $ \mathbf{E} = \{(P,Q)\} $. Parameters $ \theta $ imply conditional probabilities $ \Pr[Q|P] = 0.6 $ and $ \Pr[Q|\neg P] = 0.3 $. 	In Figure~\ref{fig:example_BN}, we enumerate all  assignment of $ P $ and $ Q $ to  incorporate all conditional probabilities of $ Q $ given $ P $. We, however, observe that the dynamic programming solution in Section~\ref{sec:dp_formulation} still prunes search space for variables that do not appear in $ \mathbf{V} $, such as $ \{R, S\} $. Hence following the calculation in Figure~\ref{fig:example_BN}, we obtain the maximum probability of positive prediction of the classifier as $ 0.65 $ for $ P = 1 $. The minimum probability of positive prediction (not shown)  is similarly calculated as $ 0.11 $ for $ P = 0 $. 
	\end{example}

	\subsection{Fairness Verification with Computed Probability of Positive Prediction} 
	Given a classifier $\mathcal{M}$, a  distribution $\mathcal{D}$, and a fairness metric $f$, verifying whether a classifier is $\epsilon$-fair ($\epsilon \in [0,1]$) is equivalent to computing $\mathds{1}[f(\mathcal{M}|\mathcal{D})\leq \epsilon]$. We now compute $f(\mathcal{M}|\mathcal{D})$ based on the maximum probability of positive prediction $ \max_{ \mathbf{a}} \Pr[\hat{Y} =1 | \sensitive = \mathbf{a}] $ and the  minimum probability of positive prediction $ \min_{ \mathbf{a}} \Pr[\hat{Y} =1 | \sensitive = \mathbf{a}] $ of a classifier.
	
	For measuring fairness metric SP, we compute the difference $ \max_{ \mathbf{a}} \Pr[\hat{Y} =1 | \sensitive = \mathbf{a}]  - \min_{ \mathbf{a}} \Pr[\hat{Y} =1 | \sensitive = \mathbf{a}] $. We, however, deploy {\framework} twice while measuring EO: one for the distribution $ \mathcal{D} $ conditioned on $ Y = 1  $ and another for $ Y = 0 $. 
	In each case, we compute $ \max_{ \mathbf{a}} \Pr[\hat{Y} =1 | \sensitive = \mathbf{a}, Y = y ]  - \min_{ \mathbf{a}} \Pr[\hat{Y} =1 | \sensitive = \mathbf{a}, Y = y ] $ for $ y \in \{0,1\} $ and take the  maximum difference as the value of EO.  
	For measuring causal metric PCF, we compute  $ \max_{ \mathbf{a}} \Pr[\hat{Y} =1 | \sensitive = \mathbf{a}, \mathbf{Z}] $ and  $ \min_{ \mathbf{a}} \Pr[\hat{Y} =1, \mathbf{Z}| \sensitive = \mathbf{a} , \mathbf{Z}] $ conditioned on mediator features $ \mathbf{Z} $ and take their difference. 
	To measure DI, we compute the ratio $ \max_{ \mathbf{a}} \Pr[\hat{Y} =1 | \sensitive = \mathbf{a}] / \min_{ \mathbf{a}} \Pr[\hat{Y} =1 | \sensitive = \mathbf{a}] $. In contrast to other fairness metrics, DI closer to 1 indicates higher fairness level. Thus, we verify whether a classifier achieves $(1 - \epsilon)$-DI by checking $ \mathds{1}[DI(\alg| \mathcal{D}) \ge 1 - \epsilon] $. 
	
	\subsection{Extension to Practical Settings}\label{sec:practical}
	For verifying linear classifiers with real-valued features and coefficients, we preprocess them so that {\framework} can be invoked. Let $ X_c \in \mathbb{R} $ be a continuous real-valued feature with coefficient $ w_c \in \mathbb{R} $ in the classifier. We discretize $ X_c $ to a set $ \mathbf{B}_{c} $ of $ k $ Boolean variables using binning-based discretization and assign a Boolean variable to each bin. Hence, $ B_i \in \mathbf{B}_{c} $ becomes $ 1 $, when $ X_c $ belongs to the $ i^\text{th} $ bin. Let $ \mu_i $ denote the mean of feature-values within $ i^\text{th} $ bin. We then set the coefficient of $ B_i $ as $ w_c\mu_i $. By law of large numbers, $ X_c \approx \sum_i \mu_iB_i $ for infinitely many bins~\cite{grimmett2020probability}. Finally, we multiply the coefficients of discretized variables by $ l \in \mathbb{N} \setminus \{0\} $ and round to an integer. Accuracy of the preprocessing step relies on the number of bins $ k $ and the multiplier $ l $. Therefore, we fine-tune both $ k $ and $ l $ by comparing  the processed classifier with the initial classifier on a validation dataset.
\begin{comment}
	\begin{figure}
		\centering
		\subfloat[Exploration tree of stochastic sub-set sum problem.]{	\includegraphics[scale=0.3]{figures/exploration_tree_DP}\label{fig:example_dp}}
		\subfloat[Bayesian network]{\includegraphics[scale=.4]{figures/example_BN}\label{fig:example_BN}}
		\subfloat[Exploration tree of stochastic sub-set sum problem with Bayesian network]{\includegraphics[scale=.4]{figures/exploration_tree_DP_BN}	\label{fig:example_dp_BN}
		}
		
	\end{figure}
\end{comment}

\section{Empirical Performance Analysis}
\label{sec:experiments}
In this section, we empirically evaluate the performance of {\framework}. We first present the experimental setup and the objective of our experiments, followed by experimental results.

\begin{figure}[t!]
	\begin{center}		
		\subfloat{\includegraphics[scale=0.3]{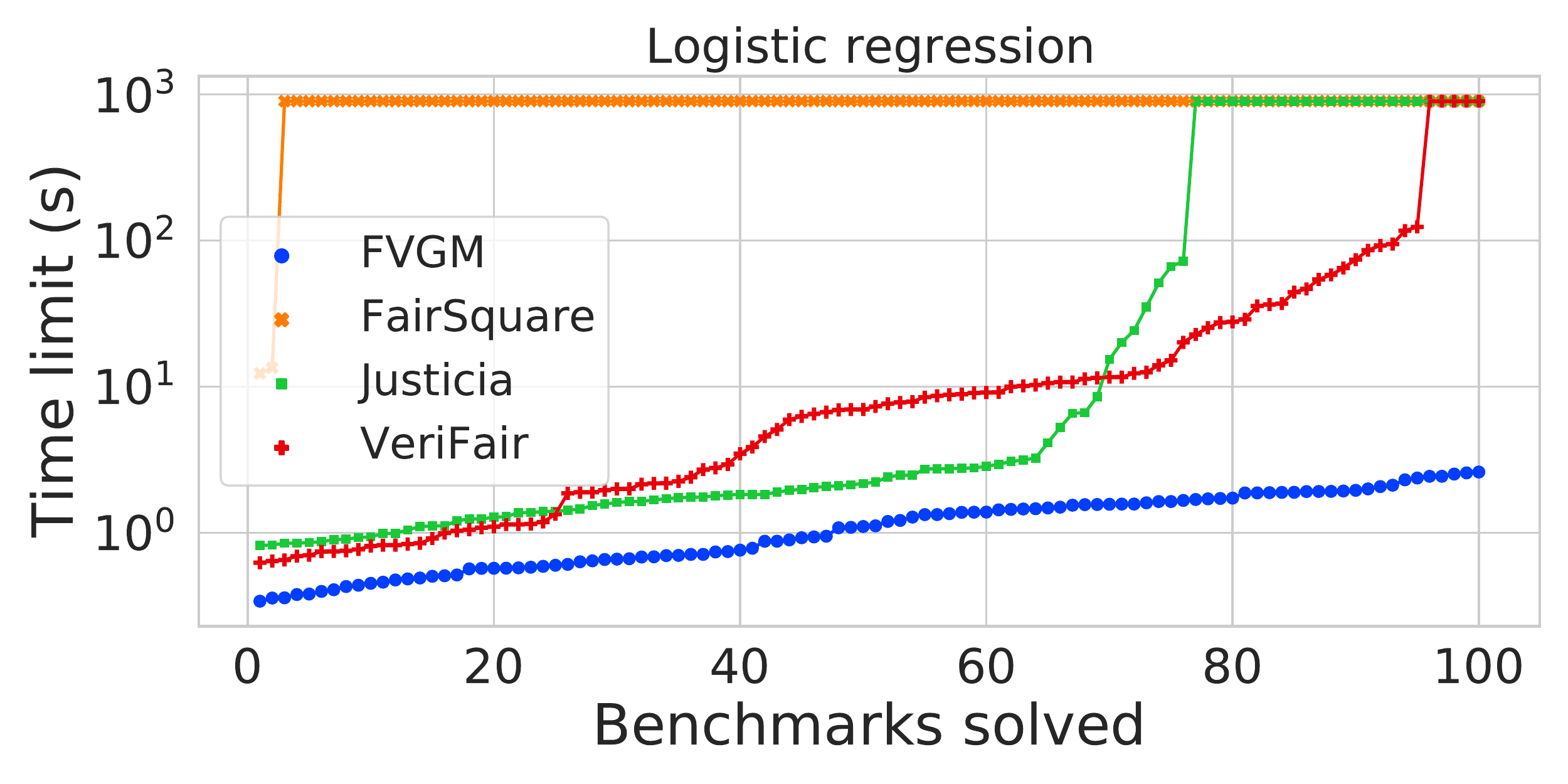}}
		\subfloat{\includegraphics[scale=0.3]{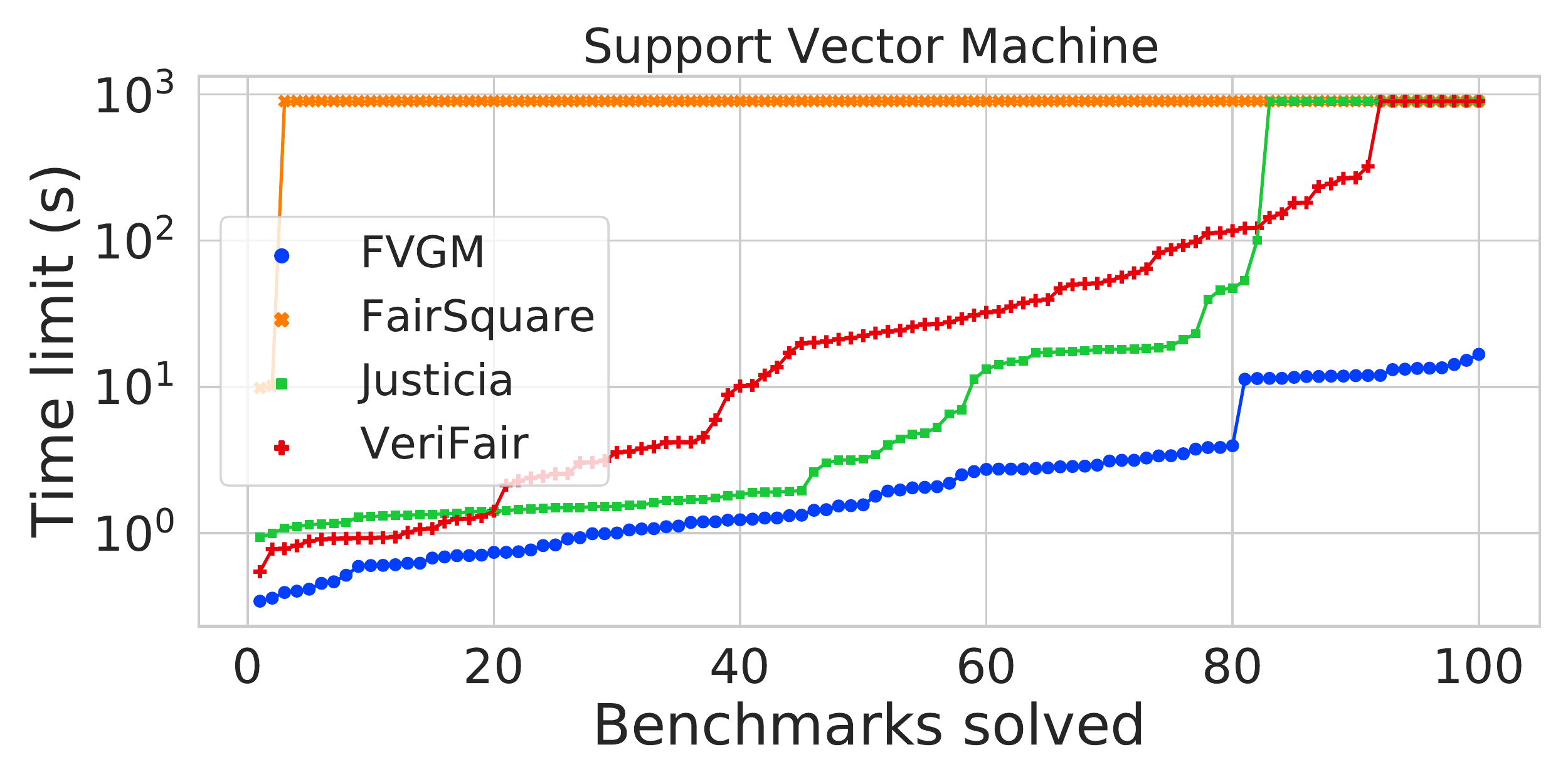}}		
	\end{center}
	\caption{A cactus plot to present the scalability of different fairness verifiers. The number of solved benchmarks are on the $ X $-axis and the required time is on the $ Y $-axis; a point $ (x,y) $ implies that a verifier takes less than or equal to $ y $ seconds to compute fairness metrics of $ x $ many benchmarks. We consider $ 100 $ benchmarks generated from $ 5 $ real-world datasets using $ 5 $-fold cross-validation. In each fold, we consider $\{25, 50, 75, 100\} $ percent of non-sensitive features.}\label{fig:scalability_exp}
\end{figure}

\textbf{Experimental Setup.}
We  implement a  prototype of {\framework} in Python (version 3.8).  
We deploy the Scikit-learn library for learning linear classifiers such as Logistic Regression (LR) and Support Vector Machine (SVM) with linear kernels. We perform five-fold cross-validation on a dataset. While the classifier is trained on continuous features, we discretize them to Boolean features to be invoked by {\framework}. During discretization, we apply a gird-search to estimate the best bin-size within a maximum bin of $ 10 $. To convert the coefficients of features into integers, we employ another grid-search to choose the best multiplier within $ \{1,2, \dots, 100\} $. For learning a Bayesian network on the converted Boolean data, we deploy the PGMPY library~\cite{ankan2015pgmpy}. For network learning, we apply a Hill-climbing search algorithm that learns a DAG structure by optimizing K2 score~\cite{koller2009probabilistic}. For estimating parameters of the network, we use Maximum Likelihood Estimation (MLE) algorithm. 

We compare {\framework} with three existing fairness verifiers: Justicia~\cite{ghosh2020justicia}, FairSquare~\cite{albarghouthi2017fairsquare}, and VeriFair~\cite{bastani2019probabilistic}. 

\subsection{Scalability Analysis}
\textbf{Benchmarks.} We perform the scalability analysis on five real-world datasets studied in fair ML literature: UCI Adult, German-credit~\cite{DK2017}, COMPAS~\cite{angwin2016machine}, Ricci~\cite{mcginley2010ricci}, and Titanic (\url{https://www.kaggle.com/c/titanic}). We consider $ 100 $ benchmarks generated from 5 real-world datasets and report the computation times (for DI and SP) of different verifiers.

\textbf{Results.} In Figure~\ref{fig:scalability_exp}, we present the scalability results of different verifiers. First, we observe that FairSquare often times out ($ =900 $ seconds) and can solve $ \le 5 $ benchmarks. This indicates that SMT-based reduction for linear classifiers cannot scale. Similarly, SSAT-based verifier Justicia that performs pseudo-Boolean to CNF translation for linear classifiers, times out for around $  20 $ out of $ 100 $ benchmarks. Sampling-based framework, VeriFair, has comparatively better scalability than SMT/SSAT based frameworks and can solve more than $ 90 $ benchmarks. Finally, {\framework} achieves impressive scalability by solving all $ 100 $ benchmarks with $ 1 $ to $ 2 $ orders of magnitude runtime improvements than existing verifiers. Therefore,\textit{ {\stochastic}-based framework {\framework} proves to be highly scalable in verifying fairness properties of linear classifiers than the state-of-the-art.} 

\begin{figure}[t!]
	\begin{center}	
		\subfloat{\includegraphics[scale=0.3]{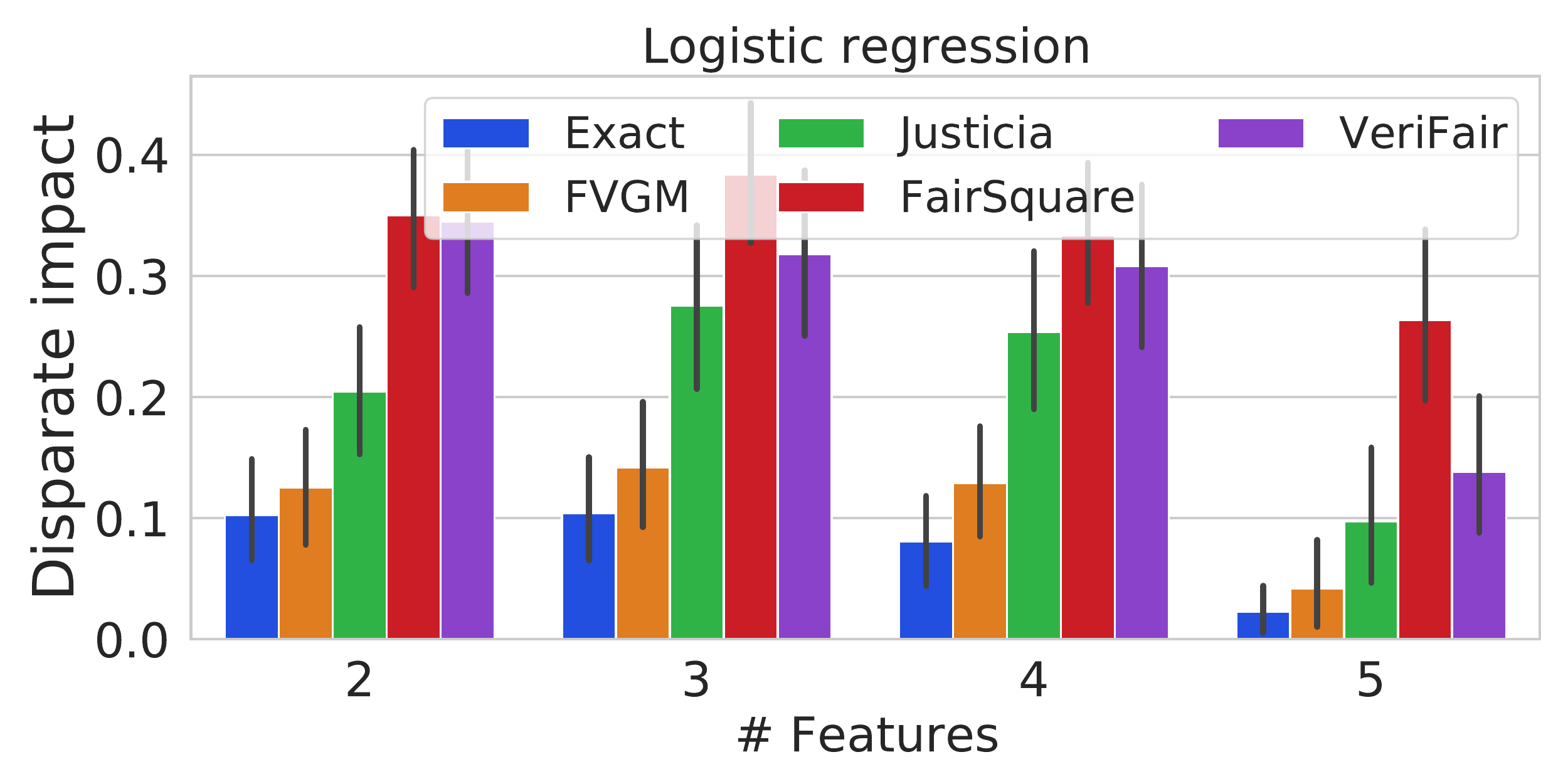}}
		\subfloat{\includegraphics[scale=0.3]{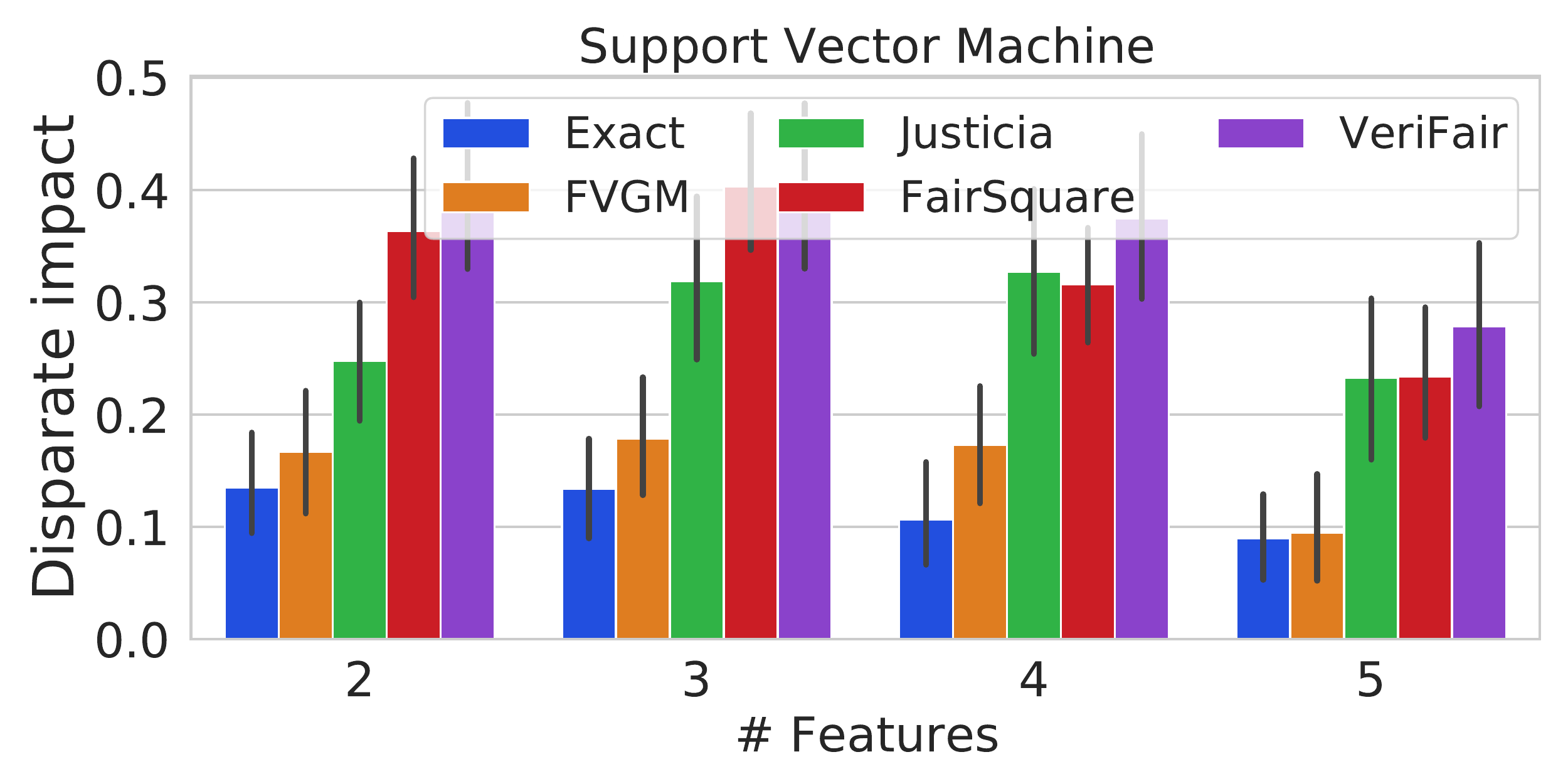}}		
		%
	\end{center}
	\caption{Comparing the average accuracy of different verifiers over 100 synthetic benchmarks while varying the number of features. {\framework} yields the closest estimation of the analytically calculated \textit{Exact} values of DI for LR and SVM classifiers.}\label{fig:sanity_exp}
\end{figure}
\subsection{Accuracy Analysis}
\noindent\textbf{Benchmark Generation.} To perform accuracy analysis, we require the ground truth, which is not available for real-world instances. Therefore, we focus on generating synthetic benchmarks for analytically computing the ground truth of different fairness metrics, such as DI, from the known distribution of features. In each benchmark, we consider $ n \in \{2, 3, 4, 5\} $ features including one Boolean sensitive feature, say $ A $, generated from a Bernoulli distribution with mean $ 0.5 $.  We generate non-sensitive features $ X_i $ from Gaussian distributions such that   $ \Pr[X_i | A = 1] \sim \mathcal{N}(\mu_i, \sigma^2) $ and $ \Pr[X_i | A = 0] \sim \mathcal{N}(\mu_i', \sigma^2) $, where $ \mu_i, \mu_i' \in [0,1] $, $ \sigma = 0.1 $, and $ \mu_i, \mu_i' $ are chosen from a uniform distribution in $ [0,1] $. Finally, we create label $ Y = \mathds{1}[ \sum_{i=1}^{n-1} X_i \ge 0.5 \sum_{i=1}^{n-1} (\mu_i + \mu_i')] $ such that $ Y $ does not directly depend on the sensitive feature. For each $ n $, we generate $ 100 $ random benchmarks, learn LR and SVM classifiers on them, and compute DI\footnote{  
To analytically compute DI, let the coefficients of the classifier be $ w_i $ for $ X_i $ and $ w_A $ for $ A $, and bias be $ \tau $. Since all non-sensitive features are from Gaussian distributions, we compute the probability of the predicted class $ \Pr[\hat{Y} | A = 1]  \sim \mathcal{N}(\sum_{i=1}^{n-1}w_i\mu_i, \sigma_{\hat{Y}}^2) $ and $ \Pr[\hat{Y} | A = 0]  \sim \mathcal{N}(\sum_{i=1}^{n-1}w_i\mu_i', \sigma_{\hat{Y}}^2) $ with $ \sigma_{\hat{Y}}^2 =   (\sum_{i=1}^{n-1}w_i^2)\sigma^2 $. Hence, the probability of positive prediction of the classifier  is $  1 - \mathsf{CDF}_{\hat{Y}| A =1}(\tau - w_A) $ for $ A = 1 $ and $  1 - \mathsf{CDF}_{\hat{Y}|A=0}(\tau) $ for $ A = 0 $, where $ \mathsf{CDF} $ is the cumulative distribution function. Finally, we compute DI by taking the ratio of the minimum and the maximum of the probability of positive prediction of the classifier.}
using different verifiers.

\textbf{Results.} 
 We  assess the accuracy of the competing verifiers in estimating fairness metrics, specifically DI with LR and SVM classifiers. In Figure~\ref{fig:sanity_exp}, {\framework} computes DI closest to the \textit{Exact} value for different number of features and both type of classifiers. In contrast, Justicia, FairSquare, and VeriFair measure DI far from the \textit{Exact} because of ignoring correlations among the features. For example, for SVM classifier with  $ n = 5 $ (right plot in Figure~\ref{fig:sanity_exp}), \textit{Exact} DI is $ 0.089 $ (average over 100 random benchmarks). Here, {\framework} computes DI as $ 0.094 $, while all other verifiers compute DI as at least $ 0.233 $. Therefore, \textit{{\framework} is more accurate than existing verifiers as it explicitly considers correlations among features}.

\begin{table}[!t]
		\centering

		\setlength{\tabcolsep}{0.45em}
		\begin{tabular}{lllrrrrrrrrrrrrr}
			
			\toprule
			Dataset & $ \sensitive $ & Algo. & $ \Delta $DI &  $ \Delta $PCF & $ \Delta $SP & $ \Delta $EO\\
			\midrule

			\multirow{4}{*}{\rotatebox[origin=c]{0}{Adult}}&\multirow{2}{*}{\rotatebox[origin=c]{0}{race}}&RW&$ \textbf{0.53} $&$ \textbf{-0.06} $&$ \textbf{-0.06} $&$ \textbf{-0.02} $\\
			&&OP&$ \textbf{0.57} $&$ \textbf{-0.07} $&$ \textbf{-0.07} $&$ \textbf{-0.02} $\\
			\cmidrule{2-7}
			&\multirow{2}{*}{\rotatebox[origin=c]{0}{sex}}&RW&$ \textbf{0.96} $&$ \textbf{-0.16} $&$ \textbf{-0.15} $&$ \textbf{-0.08} $\\
			&&OP&$ \textbf{0.43} $&$ \textbf{-0.08} $&$ \textbf{-0.08} $&$ 0.03 $\\

			\midrule
			\multirow{4}{*}{COMPAS}&\multirow{2}{*}{race}&RW&$ \textbf{0.13} $&$ \textbf{-0.07} $&$ \textbf{-0.07} $&$ \textbf{-0.06} $\\
			&&OP&$ \textbf{0.15} $&$ \textbf{-0.08} $&$ \textbf{-0.08} $&$ \textbf{-0.05} $\\
			\cmidrule{2-7}
			&\multirow{2}{*}{sex}&RW&$ \textbf{0.1} $&$ \textbf{-0.04} $&$ \textbf{-0.04} $&$ 0.04 $\\
			&&OP&$ \textbf{0.09} $&$ \textbf{-0.04} $&$ \textbf{-0.04} $&$ \textbf{-0.03} $\\
			
			\midrule
			\multirow{4}{*}{German}&\multirow{2}{*}{age}&RW&$ \textbf{0.52} $&$ \textbf{-0.53} $&$ \textbf{-0.52} $&$ \textbf{-0.47} $\\
			&&OP&$ \textbf{0.53} $&$ \textbf{-0.53} $&$ \textbf{-0.53} $&$ \textbf{-0.51} $\\
			\cmidrule{2-7}
			&\multirow{2}{*}{sex}&RW&$ -0.06 $&$ 0.06 $&$ 0.06 $&$ 0.02 $\\
			&&OP&$ -0.12 $&$ 0.12 $&$ 0.12 $&$ 0.07 $\\

			\bottomrule
		\end{tabular}
	
	\caption{Verification of fairness algorithms using {\framework}. $ \sensitive $ denotes sensitive features.  RW and OP refer to reweighing and optimized-preprocessing algorithms. Numbers in bold refer to fairness improvement.   }\label{tab:fair_algo_verification}

\end{table}

%
%
%
%
%
%
%
%
%

\section{Applications of {\framework}}\label{sec:applications}
In this section, we apply {\framework} for verifying fairness-enhancing algorithms and depreciation attacks. We also demonstrate that {\framework} facilitates computation of fairness influence functions by enabling the detection of bias due to individual features.

\paragraph{Verifying Fairness-enhancing Algorithms.} We deploy {\framework} in verifying the effectiveness of fairness-enhancing algorithms designed to ameliorate bias. For example, fairness 
pre-processing algorithms can be validated by applying {\framework} on the unprocessed and the processed data separately and comparing different fairness metrics. In Table~\ref{tab:fair_algo_verification}, we report the effect of fairness algorithms w.r.t. four fairness metrics: disparate impact (DI), path-specific causal fairness (PCF), statistical parity (SP), and equalized odds (EO). Note that, fairness is improved if DI increases and the rest of the metrics decrease. For instance, in most instances for Adult dataset, reweighing (RW)~\cite{kamiran2012data} and optimized pre-processing (OP)~\cite{calmon2017optimized} algorithms are successful in improving fairness. The exceptional case is the unfairness regarding the sensitive feature `sex', where OP algorithm fails in improving fairness metric EO. Thus, \textit{{\framework} verifies the enhancement and decrement in fairness by  fairness-enhancing algorithms. }

\begin{figure}
	\centering
	\subfloat{\includegraphics[scale=0.32]{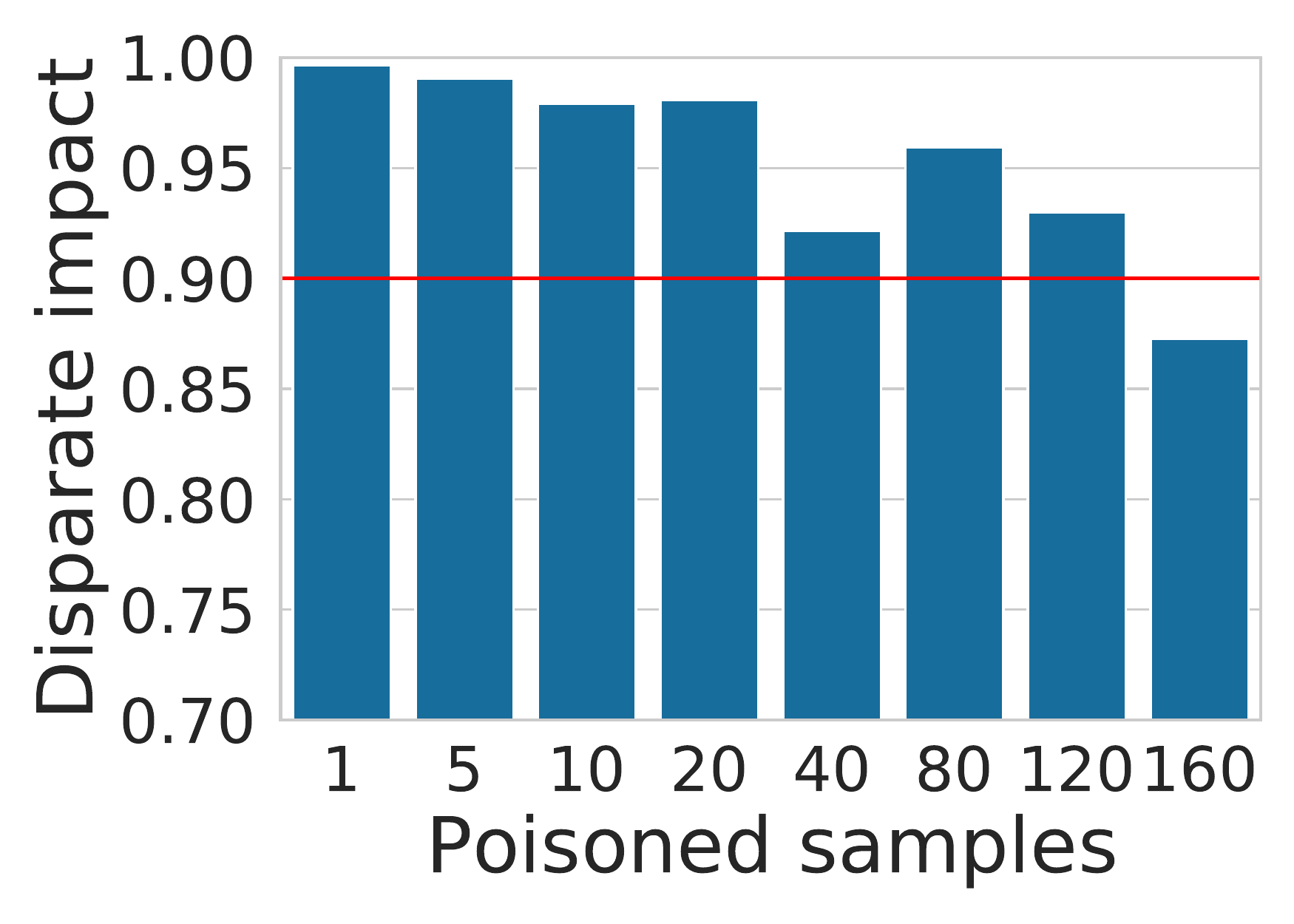}}
	\subfloat{\includegraphics[scale=0.32]{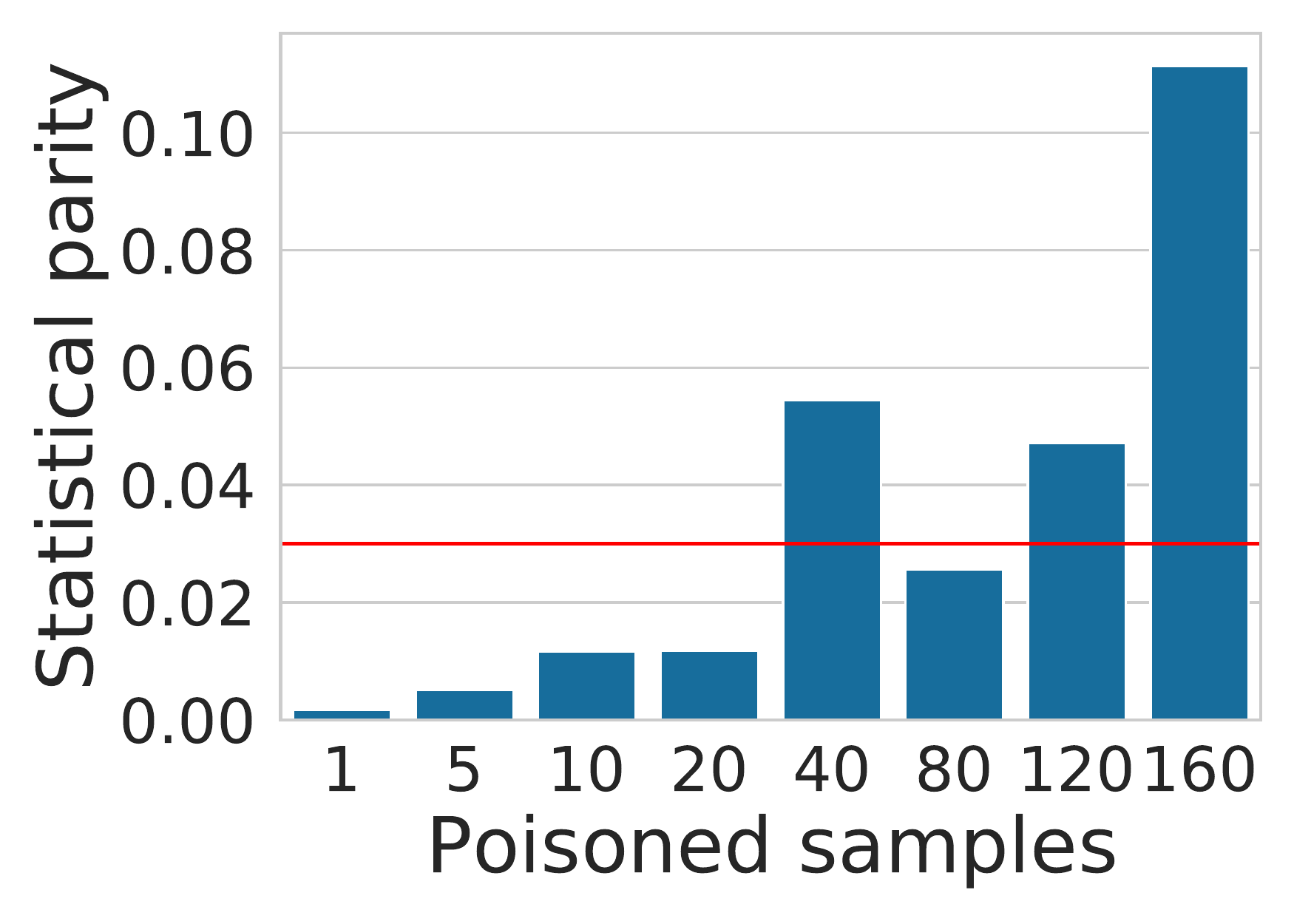}}
	
	\captionof{figure}{Verifying poisoning attack against fairness using {\framework}. The red line denotes the safety margin of the ML model against the attack.}\label{fig:fairness_attacks}
\end{figure}

\paragraph{Verifying Fairness Attacks.}
We apply {\framework} in verifying a fairness poisoning-attack algorithm. This algorithm injects a small fraction of poisoned samples into the training data such that the classifier becomes relatively unfair~\cite{solans2020poisoning}. We apply this attack to add $\{1, 5, \dots, 160\}$ poisoned samples  and measure the corresponding disparate impact and statistical parity. In Figure~\ref{fig:fairness_attacks},  {\framework} verifies that the disparate impact of the classifier decreases and statistical parity increases, i.e.\textit{ the classifier becomes more unfair},\textit{ as the number of poisoned samples increases}. Therefore, {\framework} shows the potential of being deployed in safety-critical applications to detect fairness attacks. For example, if we set $0.9$ as an acceptable threshold of disparate impact, {\framework} can raise an alarm once $160$ poisoned samples are added.

\paragraph{Fairness Influence Function (FIF): Tracing Sources of Unfairness.}
Another application of \framework{} as a fairness verifier is to quantify the effect of a subset of features on fairness. Thus, we define fairness influence function (FIF)  that computes the effect of a subset of non-sensitive features $\mathbf{S} \subseteq \nonsensitive$ on the probability of positive prediction of a classifier given a specific sensitive group $ \sensitive = \mathbf{a}$,
$
	\mathsf{FIF}(\mathbf{S}) \triangleq \Pr[\hat{Y} = 1 | \sensitive = \mathbf{a}, \mathcal{D}] - \Pr[\hat{Y} = 1 | \sensitive = \mathbf{a},  \mathcal{D}_{-\mathbf{S}}].
$
FIF allows us to explain the sources of unfairness in the classifier. In practice, we compute FIF of $\mathbf{S}$ by replacing the probability distribution of $\mathbf{S}$ with a uniformly random distribution, referred to as  $ \mathcal{D}_{-\mathbf{S}} $, and reporting differences in the conditional probability of positive prediction of the classifier. 

\begin{figure}[t!]	
	\centering
	\vspace{-2ex}
	\subfloat{\includegraphics[scale=0.32]{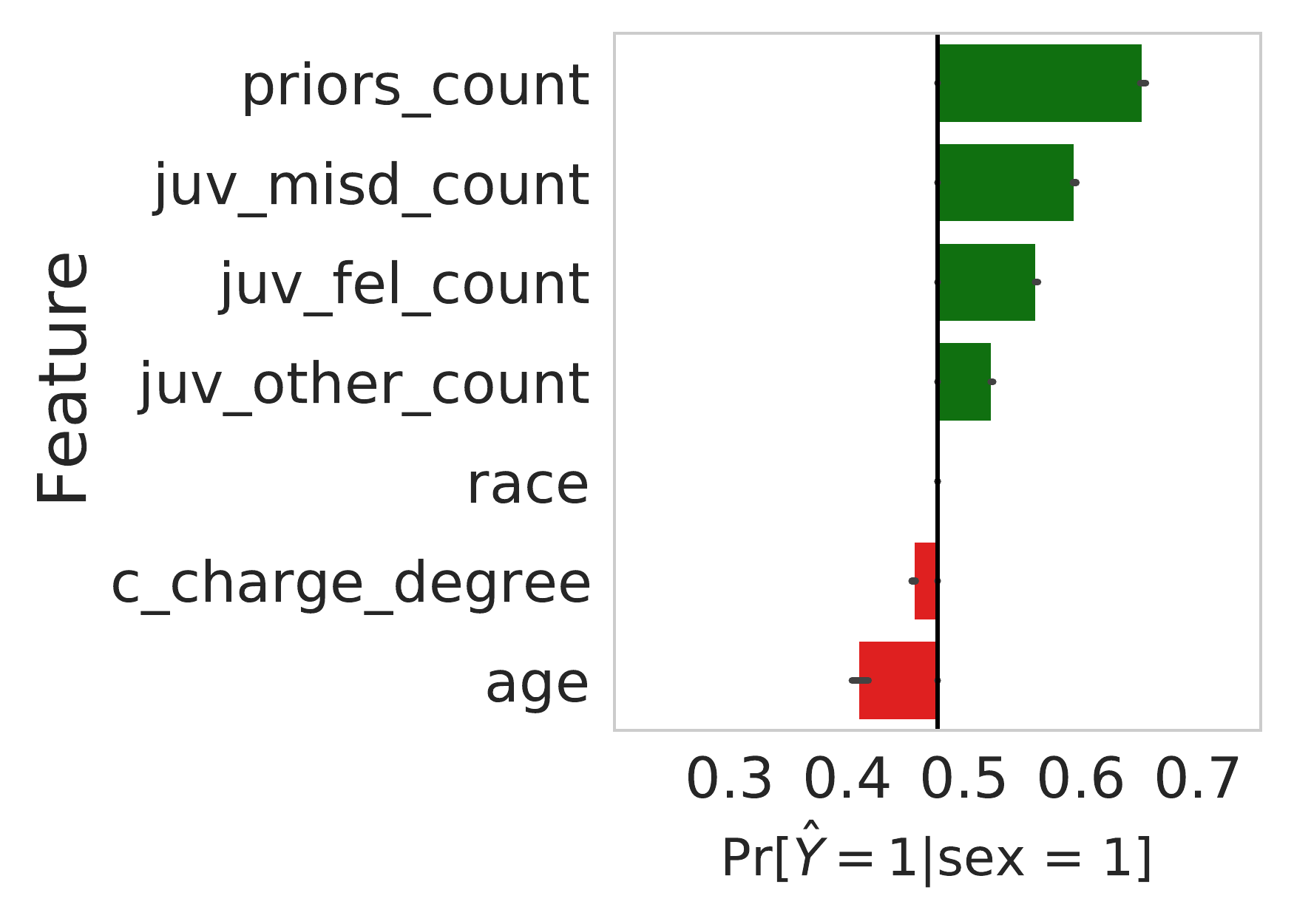}}
	\subfloat{\includegraphics[scale=0.32]{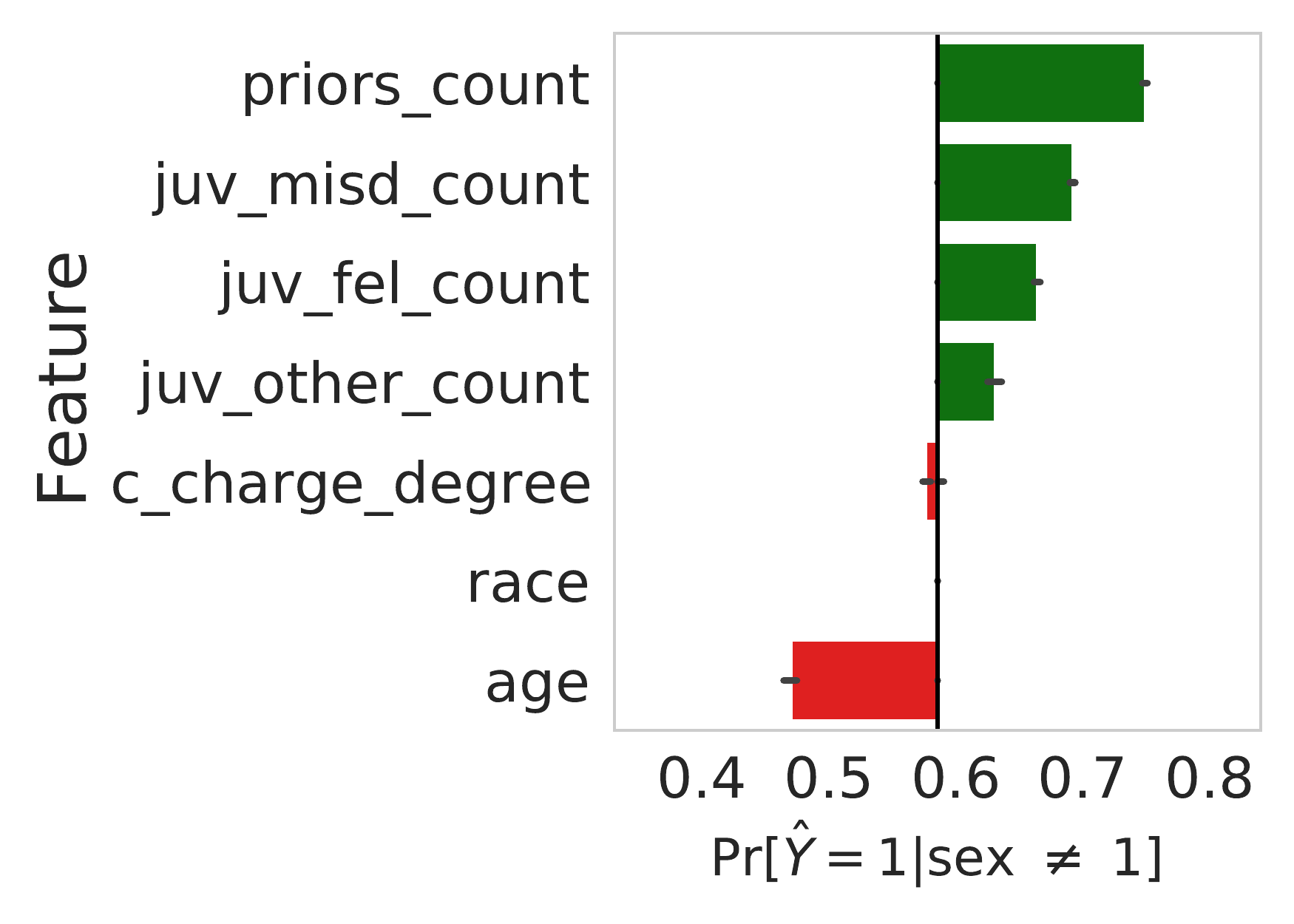}}
	\caption{FIF computation for COMPAS dataset. For Female (left plot) and Male, influence of `age' decreases and the probability of positive prediction of the classifier increases  by different magnitudes.}\label{fig:influence function}
\end{figure}

In Figure~\ref{fig:influence function}, we compute FIF for all features in COMPAS dataset, separately for two sensitive groups: Female (`sex' $ = 1 $) and Male. This dataset concerns the likelihood of a person of re-offending crimes within the next two years. We first observe that the base values of the probability of positive prediction are different for the two groups ($ 0.46 $ vs $ 0.61 $ for Female and Male), thereby showing Male as more probable to re-offend crimes than Female. Moreover, FIF of the feature `age' is comparatively higher in magnitude for Male than Female. 
This implies that while deciding recidivism the algorithm assumes that Female individuals across different ages re-offend crimes with almost the same probability and the probability of re-offending for Male individuals highly depends on age. Thus, applying {\framework} to compute FIF provides us insights about sources of bias and thus, indicators to improve fairness.
$\Pr[\hat{Y} = 1|$age\_0 $\ge$ 0.5]

\section{Conclusion}
In this paper, we propose {\framework}, an efficient fairness verification framework for linear classifiers based on a novel stochastic subset-sum problem. {\framework} encodes a graphical model of feature-correlations, represented as a Bayesian Network, and computes multiple group and causal fairness metrics accurately. We experimentally demonstrate that {\framework} is not only more accurate and scalable than the existing verifiers but also applicable in practical fairness tasks, such as verifying fairness attacks and enhancing algorithms, and computing the fairness influence functions. 
As a future work, we aim to design fairness-enhancing algorithms certified by fairness verifiers, such as {\framework}. 
Since {\framework} serves as an accurate and scalable fairness verifier for linear classifiers, it will be interesting to design such verifiers for other ML models.

\section*{Acknowledgments}
This work was supported in part by National Research Foundation, Singapore under its NRF Fellowship Programme [NRF-NRFFAI1-2019-0004 ] and AI Singapore Programme [AISG-RP-2018-005],  and NUS ODPRT Grant [R-252-000-685-13]. The computational work for this paper was performed on resources of Max Planck Institute for Software Systems, Germany and the National Supercomputing Centre, Singapore (\url{https://www.nscc.sg}).

\bibliography{main.bib}

\clearpage
\clearpage
\appendix

\clearpage

		\section{Broader Impact}
		We have presented an efficient framework for verifying fairness of linear classifiers. This work does not bear any negative societal impact. The sole purpose of our framework is to detect bias of machine learning models that are deployed in high-stake decision making.

		\section{Background}
		
		\subsection{Fairness Metrics}
		In this section, we state the fairness metrics verified in this paper in further details.
		\subsubsection{Group Fairness.} Group fairness is categorized into three families: independence, separation and sufficiency, of which {\framework} verifies independence and separation metrics. 
		The independence metrics state that the predicition of the classifier should be independent of compound sensitive groups. Formally, independence notion specifies an equal probability of positive prediction across all sensitive groups for a classifier $\alg$, i.e., $\Pr[\hat{Y} =1 | \mathbf{A} =  \mathbf{a}]  =  \Pr[\hat{Y} =1 | \mathbf{A} =  \mathbf{a}'] , \forall \mathbf{a}, \mathbf{a}' \in A$.
		Since satisfying independence exactly is hard, relaxations of independence fairness metrics, such as \textit{disparate impact} and \textit{statistical parity}~\cite{dwork2012fairness,feldman2015certifying}, are proposed. 
		
		\textit{Disparate impact} (DI)~\cite{feldman2015certifying} measures the ratio of probabilities of positive prediction between the most favored group and least favored group, and prescribe it to be close to $1$. Formally, a classifier satisfies $(1 - \epsilon)$-disparate impact if, for $\epsilon \in [0,1] $,
		\[
		\min_{\mathbf{a}} \Pr[\hat{Y} =1 | \mathbf{A} =  \mathbf{a}]  \ge (1 - \epsilon) \max_{\mathbf{a}'} \Pr[\hat{Y} =1 | \mathbf{A} =  \mathbf{a}'].
		\]
		Another popular relaxation of independence metrics  is \textit{statistical parity} (SP) that measures the difference of probability of positive prediction among sensitive groups, and prescribe this to be near zero. Formally, an algorithm satisfies $\epsilon$-statistical parity if, for $\epsilon \in [0,1] $, 
		\[
		\max_{\mathbf{a}, \mathbf{a}'}|\Pr[\hat{Y} =1 | \mathbf{A} = \mathbf{a}] - \Pr [\hat{Y} = 1| \mathbf{A} = \mathbf{a}']| \le \epsilon.
		\]
		For both disparate impact and statistical parity, lower value of $\epsilon$ indicates higher group fairness of the classifier $\alg$.

		In the \textit{separation (or classification parity)} notion of fairness, the predicted label $\hat{Y}$ of a classifier $\alg$ is independent of the sensitive features $\sensitive$ given the class labels $Y$. In case of binary classifiers, a popular separation metric is \textit{equalized odds} (EO)~\cite{hardt2016equality} that computes the difference of false positive rates (FPR) and the difference of true positive rates (TPR) among all compound sensitive groups. 
		Lower value of equalized odds indicates better fairness.
		A classifier $\alg$ satisfies $\epsilon$-equalized odds if, for all compound sensitive groups $\mathbf{a}, \mathbf{a}' \in A$,
		$ \max_{\mathbf{a}, \mathbf{a}'} |\Pr[\hat{Y} =1 |\mathbf{A}= \mathbf{a}, Y= 0  ] - \Pr [\hat{Y} = 1|\mathbf{A}= \mathbf{a}', Y = 0]| \le \epsilon, $ and $
		\max_{\mathbf{a}, \mathbf{a}'}|\Pr[\hat{Y} =1 |\mathbf{A}= \mathbf{a}, Y= 1  ] - \Pr [\hat{Y} = 1|\mathbf{A}= \mathbf{a}', Y = 1]| \le \epsilon.
		$

		\subsubsection{Path-specific Causal Fairness.}
		Let $ \mathbf{a}_{\max}  \triangleq \argmax_{ \mathbf{a}} \Pr[\hat{Y} =1 |\mathbf{A}=  \mathbf{a}] $. We consider mediator features $ \mediator \subseteq \nonsensitive $ sampled from the conditional distribution $ {\mathcal{Z}_{|\mathbf{A} = \mathbf{a}_{\max}}} $. This emulates the fact that mediator variables have the same sensitive features $ \mathbf{a}_{\max} $.  For $ \epsilon \in [0,1] $,  path-specific causal fairness is defined as 
		$
		\max_{\mathbf{a}, \mathbf{a}'} |\Pr[\hat{Y} = 1 | \sensitive =  \mathbf{a}, \mediator] - \Pr[\hat{Y} = 1 | \sensitive = \mathbf{a}', \mediator ]| \le \epsilon
		$.
		
		Therefore, PCF constrains that $ \hat{Y} $ is not directly dependent of $ \sensitive $ while $ \sensitive $ may indirectly affects $ \hat{Y} $ only through $ \mediator $. PCF is a variation of counterfactual fairness and causal fairness without mediator features~\cite{bastani2019probabilistic}.

		\begin{example}
			Following~\cite{bastani2019probabilistic}, we consider a classifier that decides the hiring of employees based on three features: gender (sensitive), years of experience (non-sensitive), and college-participation (mediator). It is practical to consider that gender $ \in $ \{male, female\} can affect the college-participation of individuals, and all three features are determining factors for the hiring process. Let `male' be the most favored group by the classifier, for instance. Path-specific causal fairness (PCF) ensures that a female candidate should be given a job offer with similar probability as a male candidate (by constraining $ \epsilon \approx 0 $). She,  however,  went to (participated in) college as if she were a male candidate while other non-mediator features such as  `years of experience' are the same.  Therefore, PCF measures the effect of gender on job offer, but ignores the effect of gender on whether candidates went to college.
		\end{example}

		\section{Proofs of Theoretical Results}

		\begin{lemmarep}
			Let $ n' $ be the number of existential and universal variables in $ \mathbf{B} $. Let $ w_{\exists} = \sum_{B_i \in \mathbf{B} | q_i = \exists} \max\{w_i, 0\}$  and $ w_{\forall} = \sum_{B_i \in \mathbf{B} | q_i = \forall} \min\{w_i, 0\}$ be the considered sum of weights of existential and universal variables, respectively. We can exactly solve {\stochastic} using dynamic programming with time complexity $ \mathcal{O}((n - n')(\tau + |w_{neg}| - w_{\exists} - w_{\forall}) + n') $. The space complexity is  $ \mathcal{O}((n - n')(\tau + |w_{neg}| - w_{\exists} - w_{\forall})) $.
		\end{lemmarep}
	
		\begin{proof}
			\textbf{Case 1: All $n$ variables in $\mathbf{B}$ have randomized quantifiers.}
			
			At step-$i$ of the dynamic programming (Eq.~\eqref{eq:dp_recurse}), we modify the residual threshold of that step, namely $\tau_i$, either by subtracting $w_i$ or by retaining it.
			Now, we observe that the residual threshold $\tau_i$ for any $i \in \lbrace1,\ldots,n\rbrace$ will be in $[0, \tau+|w_{neg}|]$. This holds because if $\tau_i$ crosses these bounds, the dynamic programming is terminated as shown in Equation~\eqref{eq:dp_terminus}.
			Since all weights of $\{w_i\}_{i=1}^{n}$ are integers, the maximum number of values that the residual threshold can take, is $ (\tau + |w_{neg}|) $.
			Thus, we need to store at most $ n(\tau + |w_{neg}|)$ values in the memory for performing dynamic programming with $n$ variables and $ (\tau + |w_{neg}|) $ number of possible weights.
			Thus, the space complexity  is $ \mathcal{O}(n(\tau + |w_{neg}|)) $.  
			
			In order to construct the dynamic programming table, we have to call the $\mathsf{dp}$ function $ \mathcal{O}(n(\tau + |w_{neg}|)) $ times, in the worst-case.
			Thus, the time complexity of our method is $ \mathcal{O}(n(\tau + |w_{neg}|) $.

			\textbf{Case 2: $n'$ variables have existential or universal quantifiers and $n-n'$ variables have randomized quantifiers in $\mathbf{B}$.}

			According to Eq.~\eqref{eq:dp_recurse}, $ w_\exists $ and $ w_\forall $ are the fixed weights of all existential and universal variables, respectively. Therefore, we need to consider at most $ \tau + |w_{neg}| -  w_\exists - w_\forall $ values of residual weights for random variables. By applying analysis in Case 1, the space and time complexity is derived as $ \mathcal{O}((n - n')(\tau + |w_{neg}| - w_{\exists} - w_{\forall})) $. 
			
			We note that there is an additional time complexity of $ \mathcal{O}(n') $ for existential and universal variables in Eq.~\eqref{eq:dp_recurse}. Thus the time complexity becomes $ \mathcal{O}((n - n')(\tau + |w_{neg}| - w_{\exists} - w_{\forall}) + n') $. We, however, do not require to store any entry for existential and universal variables in $ \mathsf{dp} $  function and thus, the space complexity remains the same as $ \mathcal{O}((n - n')(\tau + |w_{neg}| - w_{\exists} - w_{\forall})) $.
		\end{proof}
	
		\begin{lemmarep}
			Let $ \mathbf{V} \subseteq \mathbf{B} $ be the set of vertices in the Bayesian network and $ n'' $ be the number of existential and universal variables in $ \mathbf{B} \setminus \mathbf{V} $. Let $ w'_{\exists} = \sum_{B_i \in \mathbf{B} \setminus \mathbf{V} | q_i = \exists} \max\{w_i, 0\}$  and $ w'_{\forall} = \sum_{B_i \in \mathbf{B} \setminus \mathbf{V} | q_i = \forall} \min\{w_i, 0\}$ be the sum of considered weights of existential and universal variables, respectively that only appear in $ \mathbf{B} \setminus \mathbf{V} $. To exactly compute {\stochastic} with correlated variables in dynamic programming approach,  time complexity is $ \mathcal{O}(2^{|\mathbf{V}|} + (n - n'' - |\mathbf{V}|)(\tau + |w_{neg}| - w'_{\exists} - w'_{\forall}) + n'') $ and space complexity is $ \mathcal{O}((n - n'' - |\mathbf{V}|)(\tau + |w_{neg}| - w'_{\exists} - w'_{\forall})) $.
		\end{lemmarep}

		\begin{figure}
	\begin{center}
		\subfloat[]{\includegraphics[scale=0.45]{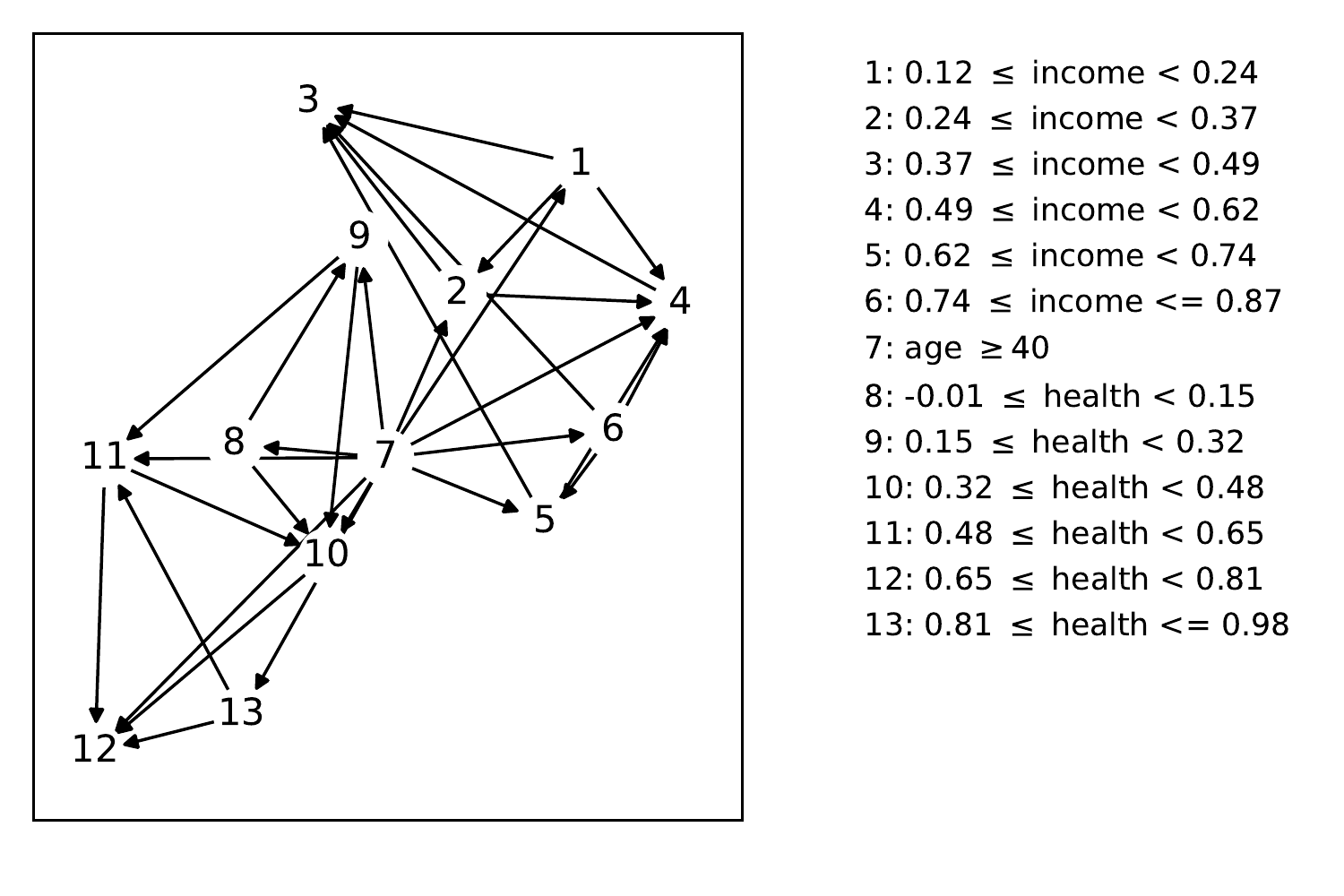}\label{fig:synthetic_bn}}\\
		\vspace{-1em}		
%
%
%
		\subfloat[]{\includegraphics[scale=0.33]{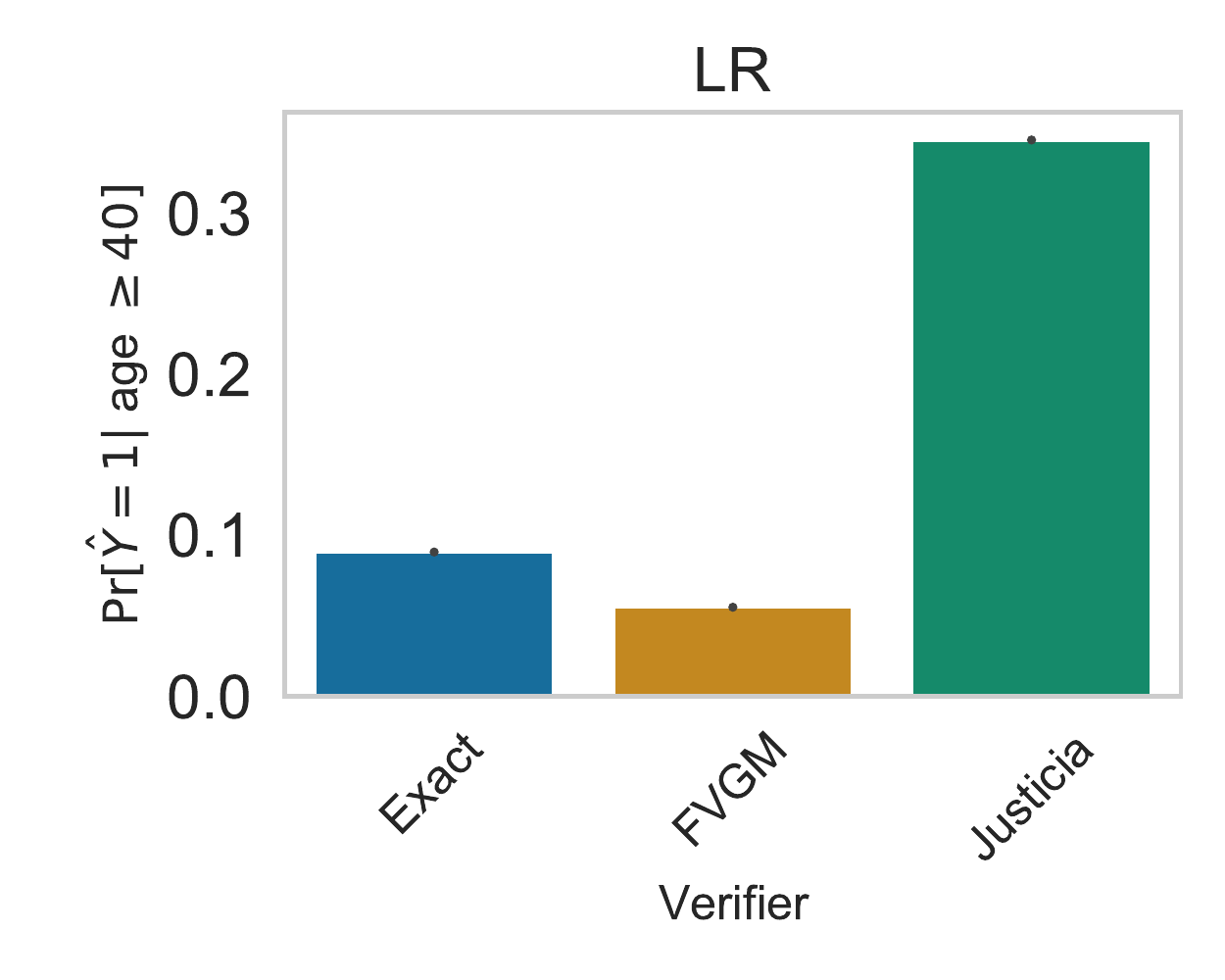}\label{fig:synthetic_ppv_max_LR}}
		\subfloat[]{\includegraphics[scale=0.33]{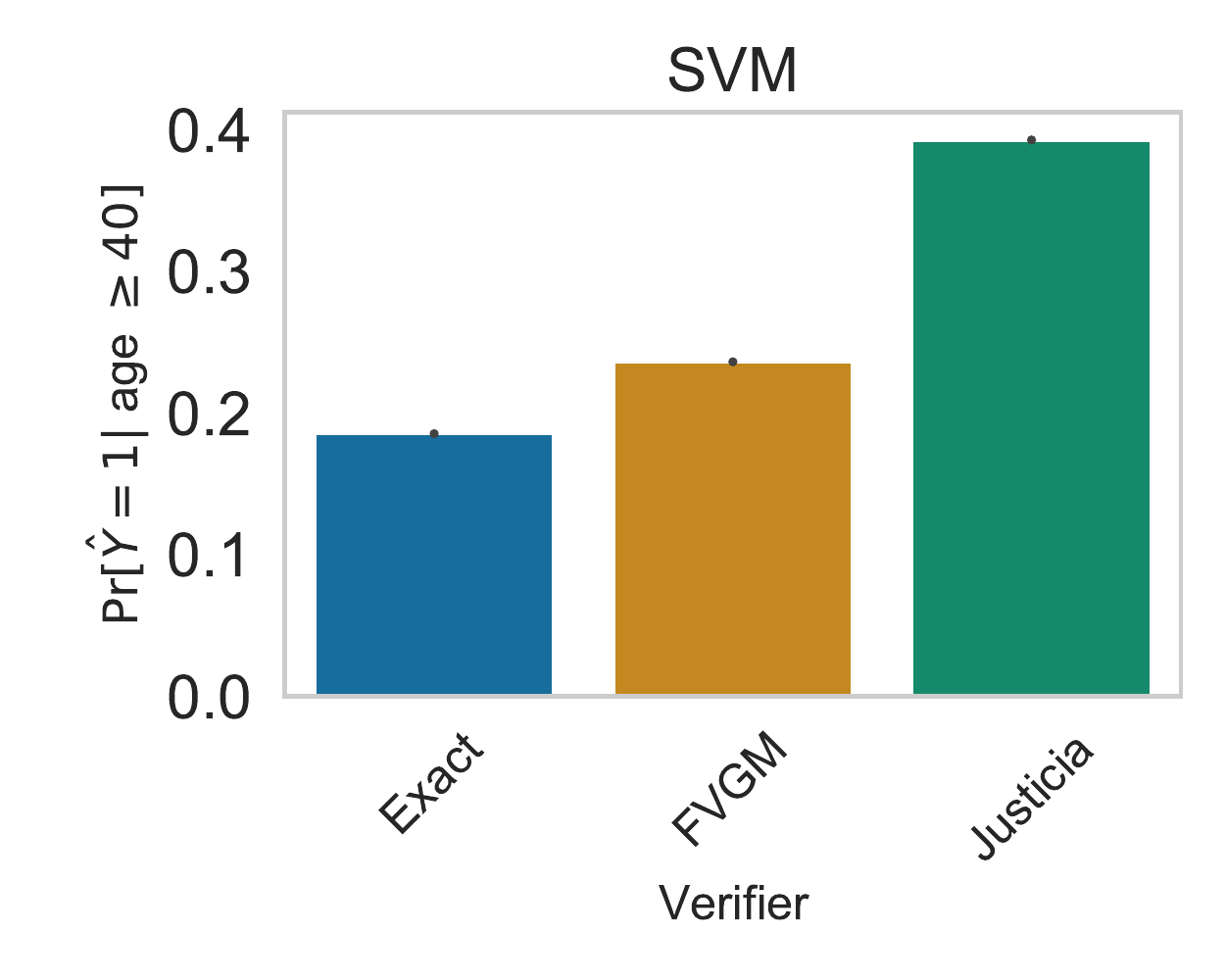}}\\
		\vspace{-1em}
		\subfloat[]{\includegraphics[scale=0.33]{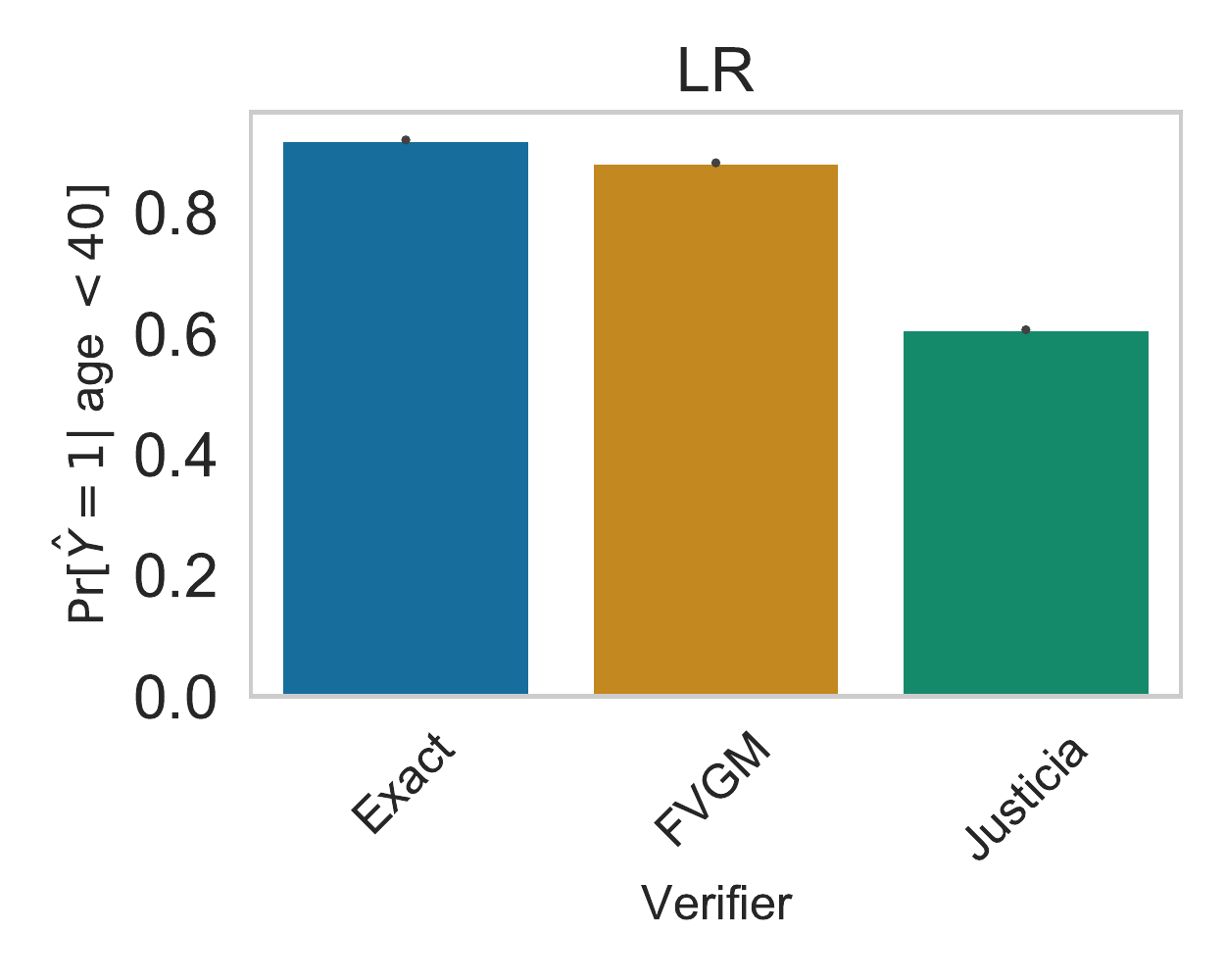}}
		\subfloat[]{\includegraphics[scale=0.33]{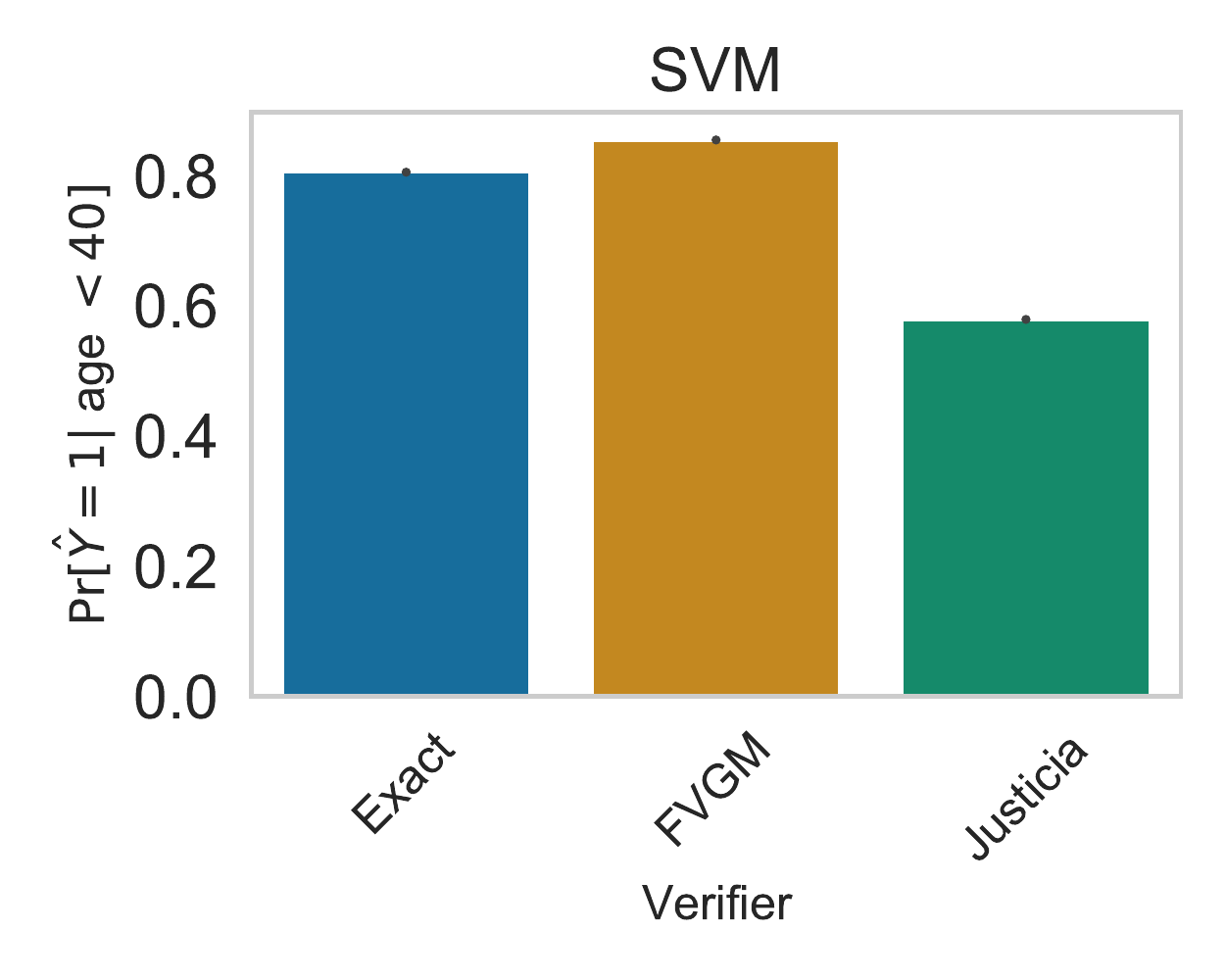}\label{fig:synthetic_ppv_min_SVM}}
		\vspace{-1em}
		
		\subfloat[]{\includegraphics[scale=0.33]{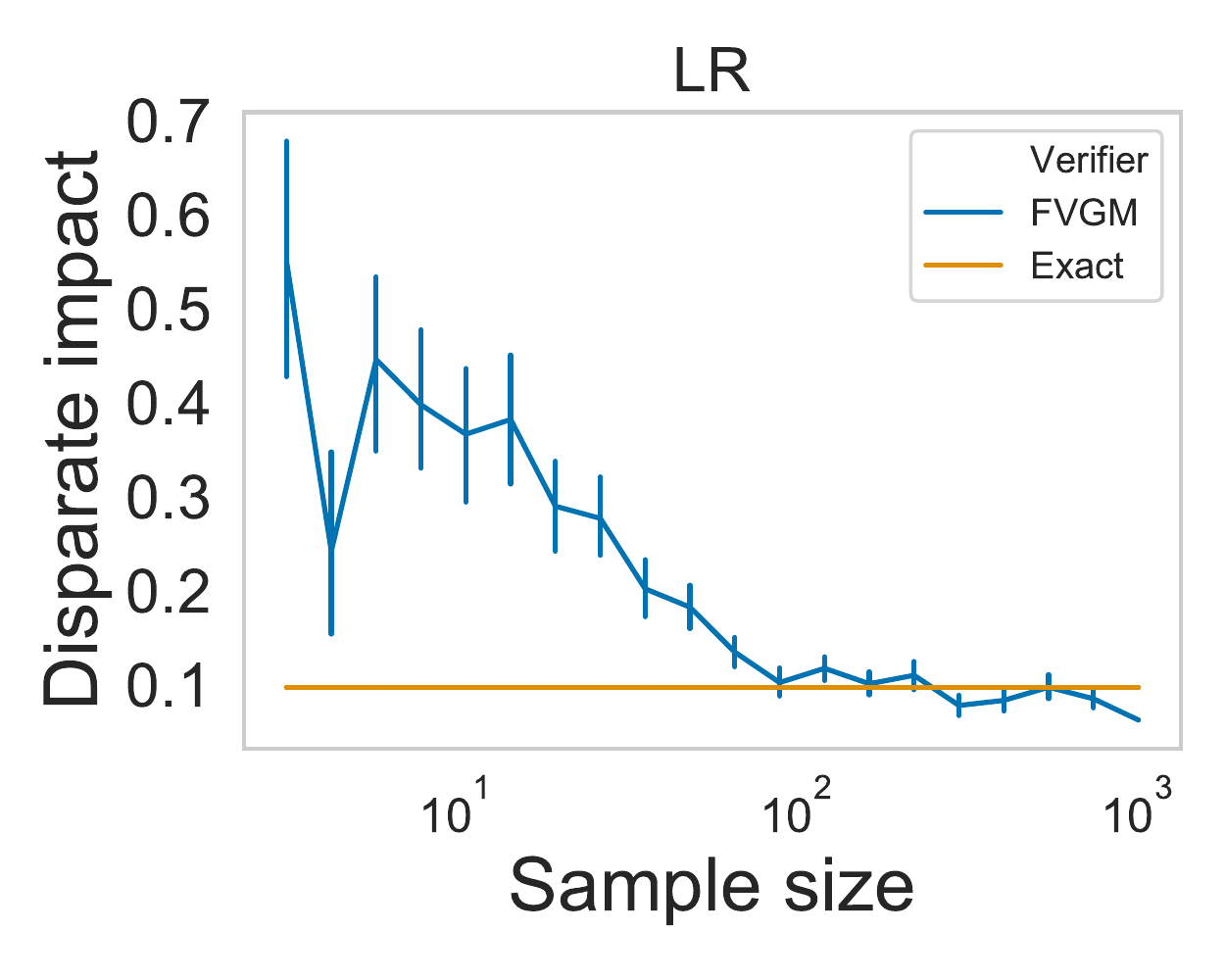}\label{fig:synthetic_sample_size_LR}}
		\subfloat[]{\includegraphics[scale=0.33]{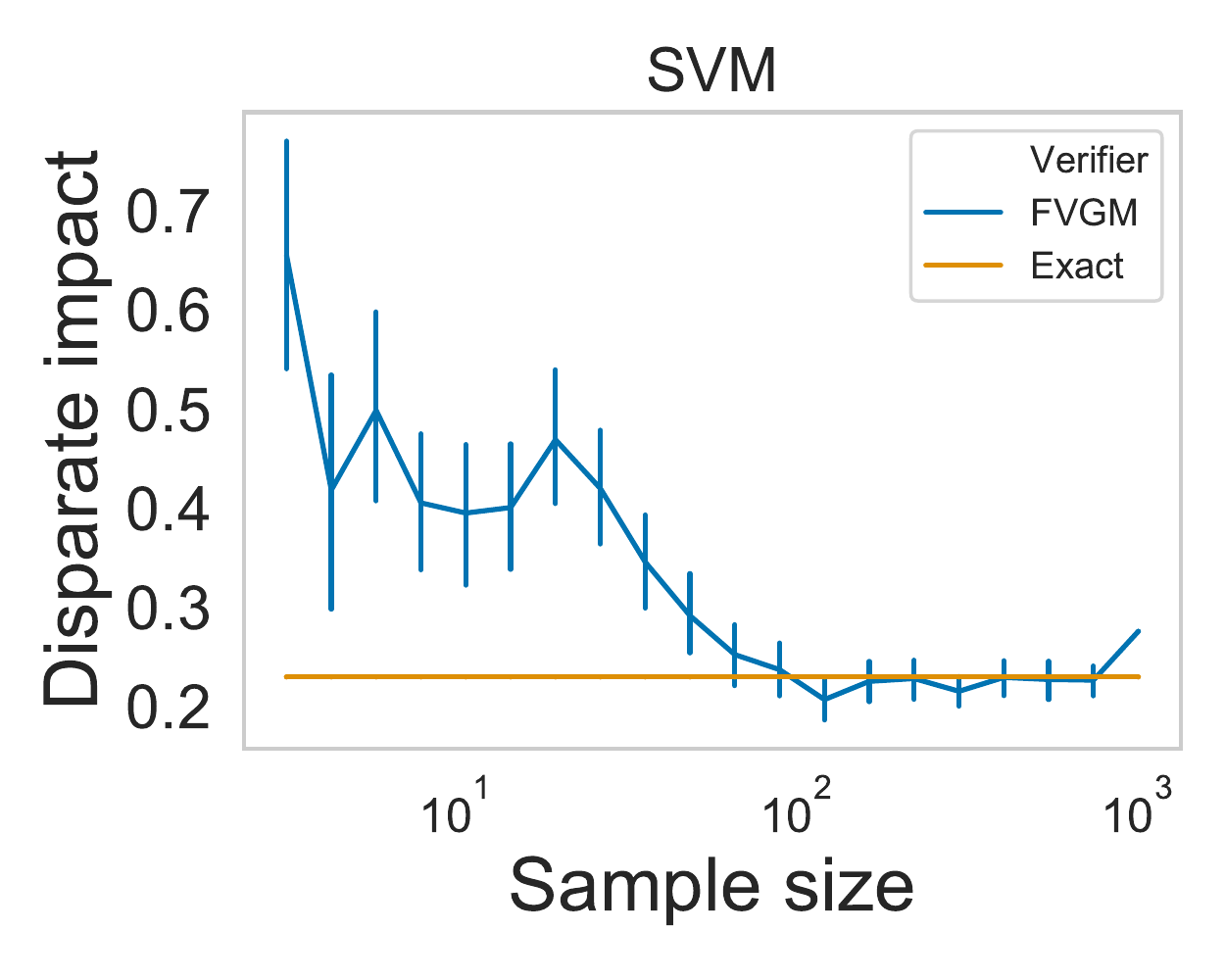}}\\
		\vspace{-1em}
		\subfloat[]{\includegraphics[scale=0.33]{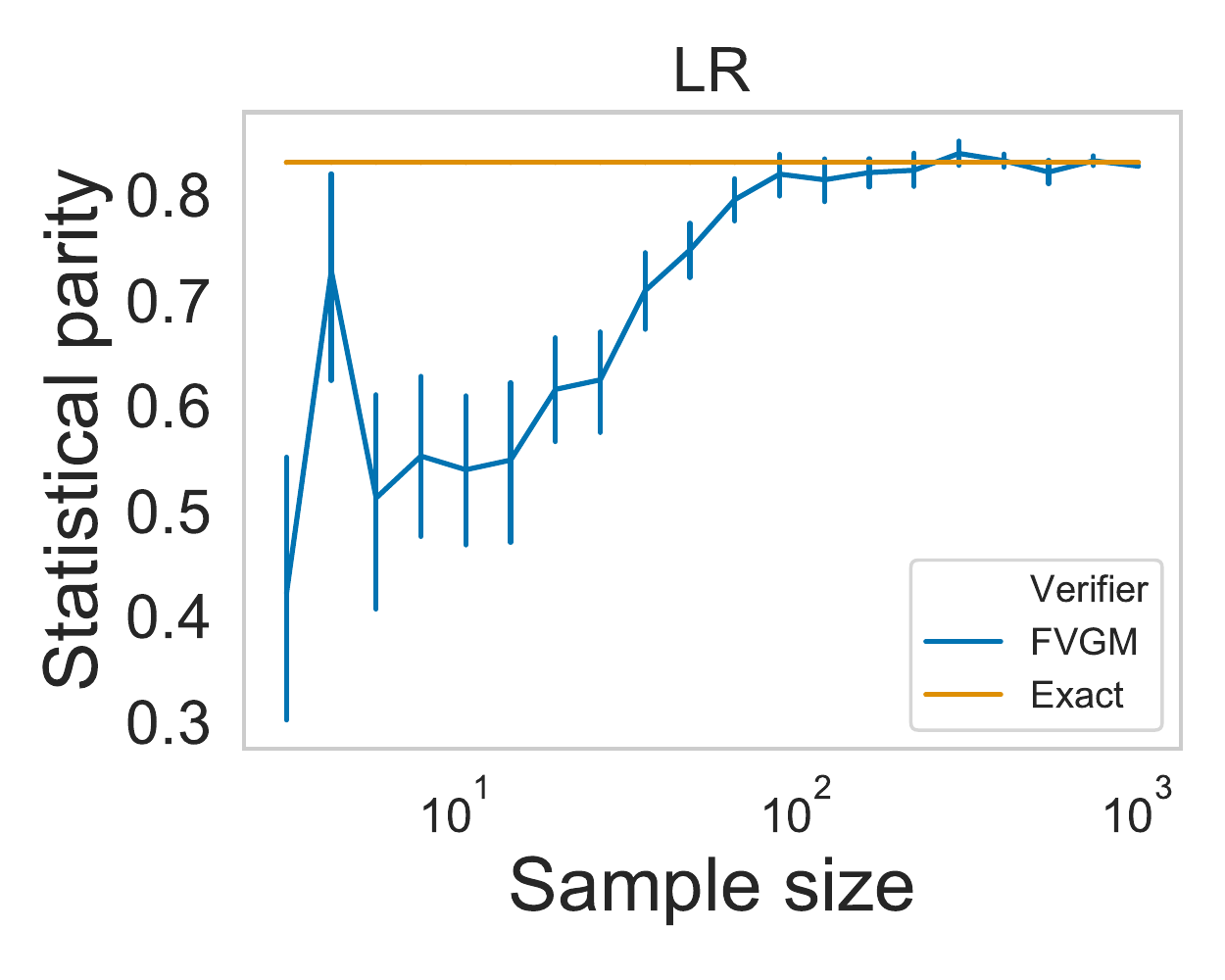}}
		\subfloat[]{\includegraphics[scale=0.33]{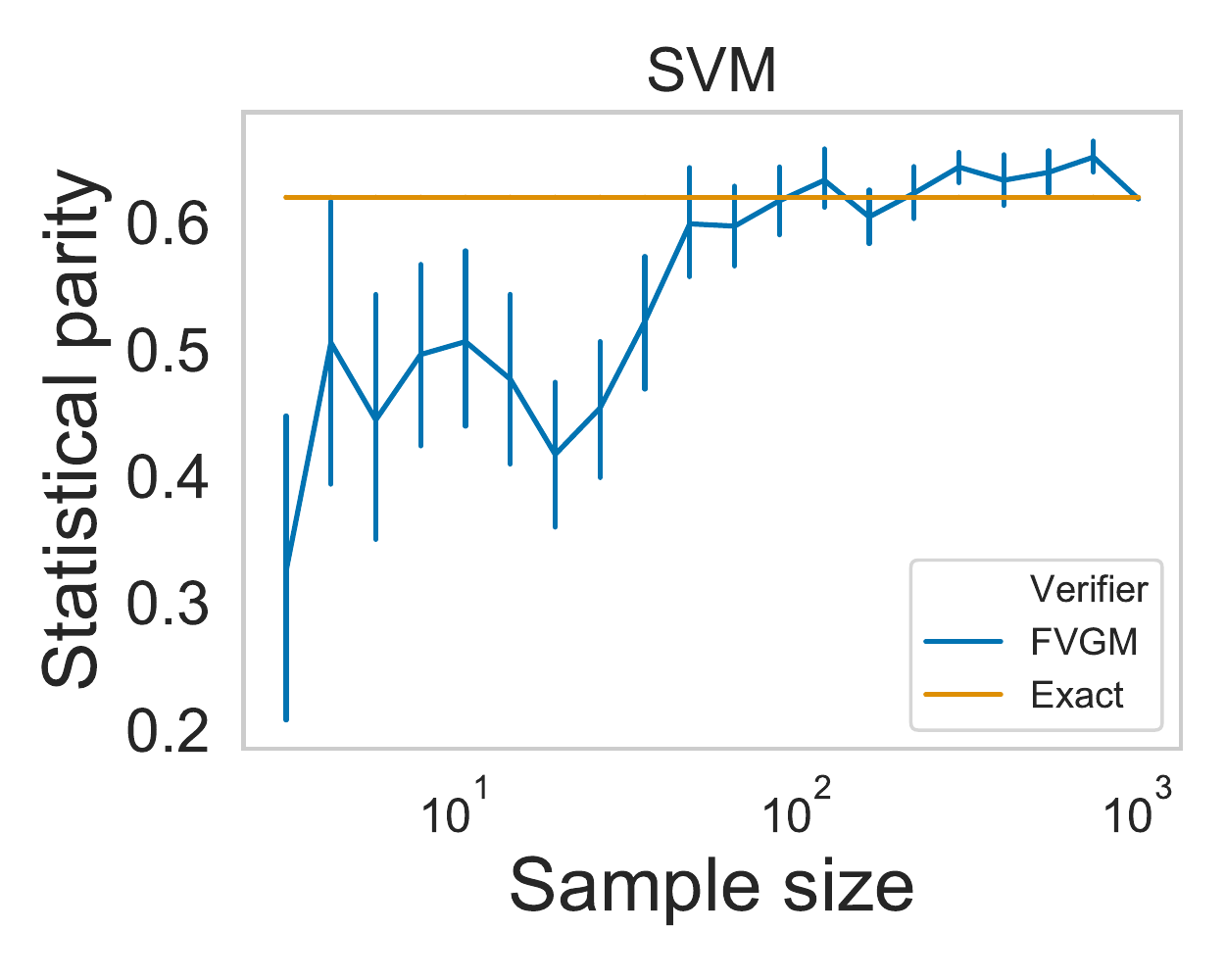}\label{fig:synthetic_sample_size_SVM}}
		\vspace{-1.5em}

	\end{center}
	
	\caption{
		Measuring accuracy of different fairness verifiers for Example~\ref{example:intro}. LR, SVM, DT refers to Logistic Regression classifier, Support Vector Machine, and Decision Tree, respectively.}
	\label{fig:synthetic_results_appendix}
\end{figure}

		\begin{proof}
			We first separate analysis of space and time complexity for variables in $ \mathbf{V} $ and in $\mathbf{B}\setminus \mathbf{V} $. For each Boolean variable in $ \mathbf{V} $, we enumerate all assignments, which has time complexity of $ 2^{|\mathbf{V}|} $ and there is no space complexity as discussed in Section~\ref{sec:dp_with_BN}.

			For variables in $\mathbf{B}\setminus \mathbf{V} $, we apply analysis from  Lemma~\ref{lemma:complexity_sss}, where we consider $ (n - n'' - |\mathbf{V}|) $ random variables, $ n'' $ existential/universal variables, and residual weights can take at most $ (\tau + |w_{neg}| - w'_{\exists} - w'_{\forall}) $ values. Hence, time complexity is $ \mathcal{O}((n - n'' - |\mathbf{V}|)(\tau + |w_{neg}| - w'_{\exists} - w'_{\forall}) + n'') $, and space complexity is $ \mathcal{O}((n - n'' - |\mathbf{V}|)(\tau + |w_{neg}| - w'_{\exists} - w'_{\forall})) $ 
			
			Combining two cases, overall time complexity is $ \mathcal{O}(2^{|\mathbf{V}|} + (n - n'' - |\mathbf{V}|)(\tau + |w_{neg}| - w'_{\exists} - w'_{\forall}) + n'') $ and space complexity is $ \mathcal{O}((n - n'' - |\mathbf{V}|)(\tau + |w_{neg}| - w'_{\exists} - w'_{\forall})) $. 
		\end{proof}

	\section{Extended Experimental Evaluations}
	\label{appendix:experiments}
	Each experiment is performed on Intel Xeon E$ 7-8857 $ v$2 $ CPUs with $ 16 $GB memory, $ 64 $bit Linux distribution based on Debian OS and clock speed $ 3 $ GHz. In the following, we discuss extended experimental results.

	\subsection{Accuracy Comparison Among Different Verifiers}
	We have considered a synthetic problem for comparing accuracy among different verifiers. For Example~\ref{example:intro}, we consider `age $ \ge 40 $' as a Bernoulli random  variable with probability $ 0.5 $. For `income' feature ($ I $), we consider two Gaussian distributions $ \Pr[I | \text{age} \ge 40] \sim \mathcal{N}(0.6, 0.1) $ and $ \Pr[I | \text{age} < 40] \sim \mathcal{N}(0.4, 0.1) $ separated by two age groups. Moreover, for `fitness' feature ($ F $), we consider two Gaussian distributions $ \Pr[F | \text{age} \ge 40] \sim \mathcal{N}(0.7, 0.1) $ and $ \Pr[F | \text{age} < 40] \sim \mathcal{N}(0.3, 0.1) $. On this data, the trained LR and SVM classifier has decision boundary as $ 7.26I + 7.4F - 1.34A \ge 6.62 $ and $ 9.37I + 9.75F - 0.34A \ge 9.4 $, respectively.

	In Figure~\ref{fig:synthetic_bn} we show the Bayesian Network on discretized features, in particular for income and fitness features. In Figure~\ref{fig:synthetic_ppv_max_LR} \ to Figure~\ref{fig:synthetic_ppv_min_SVM}, we show the probability of positive prediction of different classifiers computed by different verifiers, where  {\framework} outputs closest to exactly computed values, in comparison with Justicia. In Figure~\ref{fig:synthetic_sample_size_LR} to Figure~\ref{fig:synthetic_sample_size_SVM}, we show the effect of sample size on {\framework} in measuring fairness metrics: disparate impact and statistical parity, where with increasing sample size, the estimate becomes more accurate.

		\begin{figure}[!t]
		\begin{center}
			\subfloat{\includegraphics[scale=0.3]{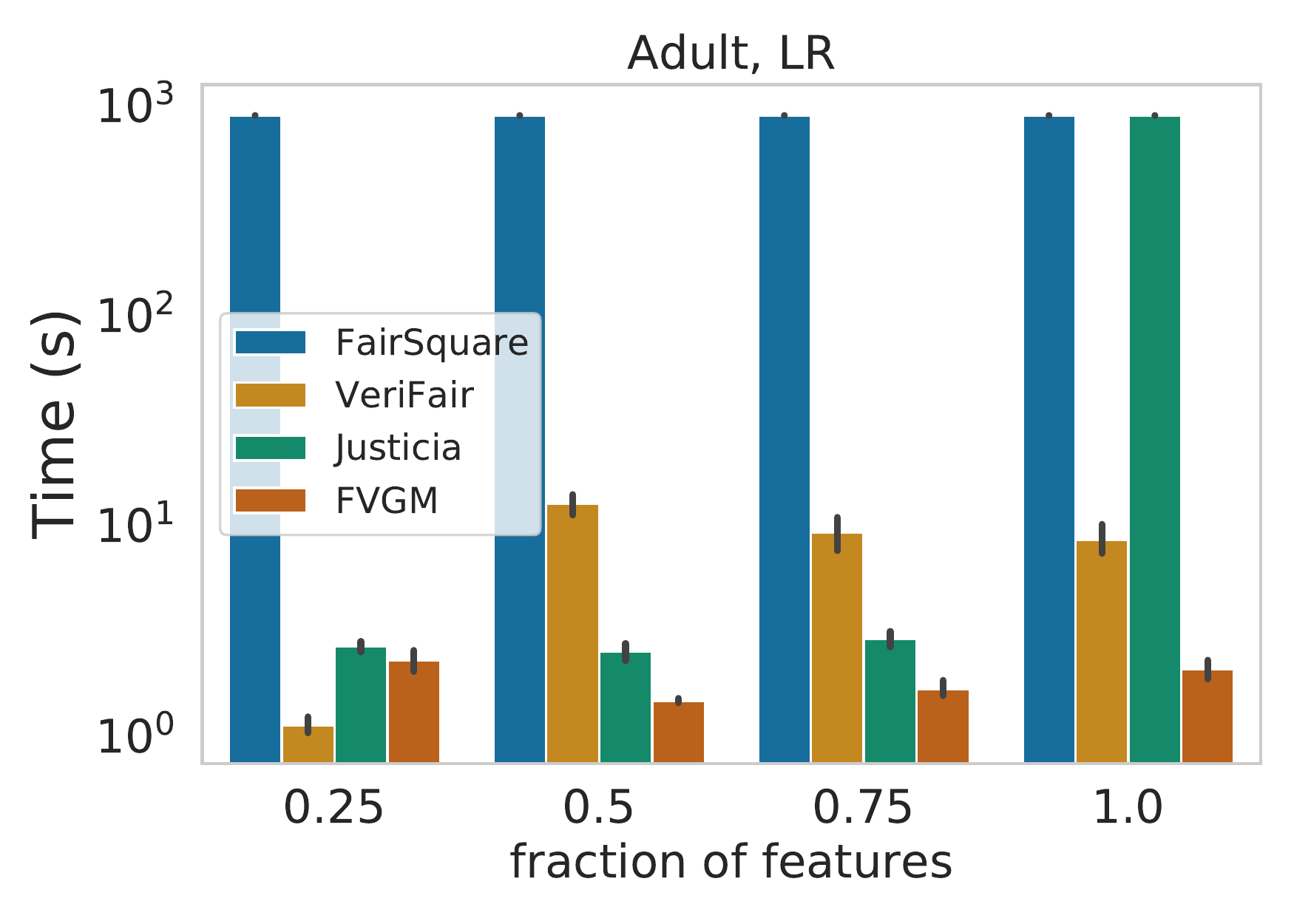}}
			\subfloat{\includegraphics[scale=0.3]{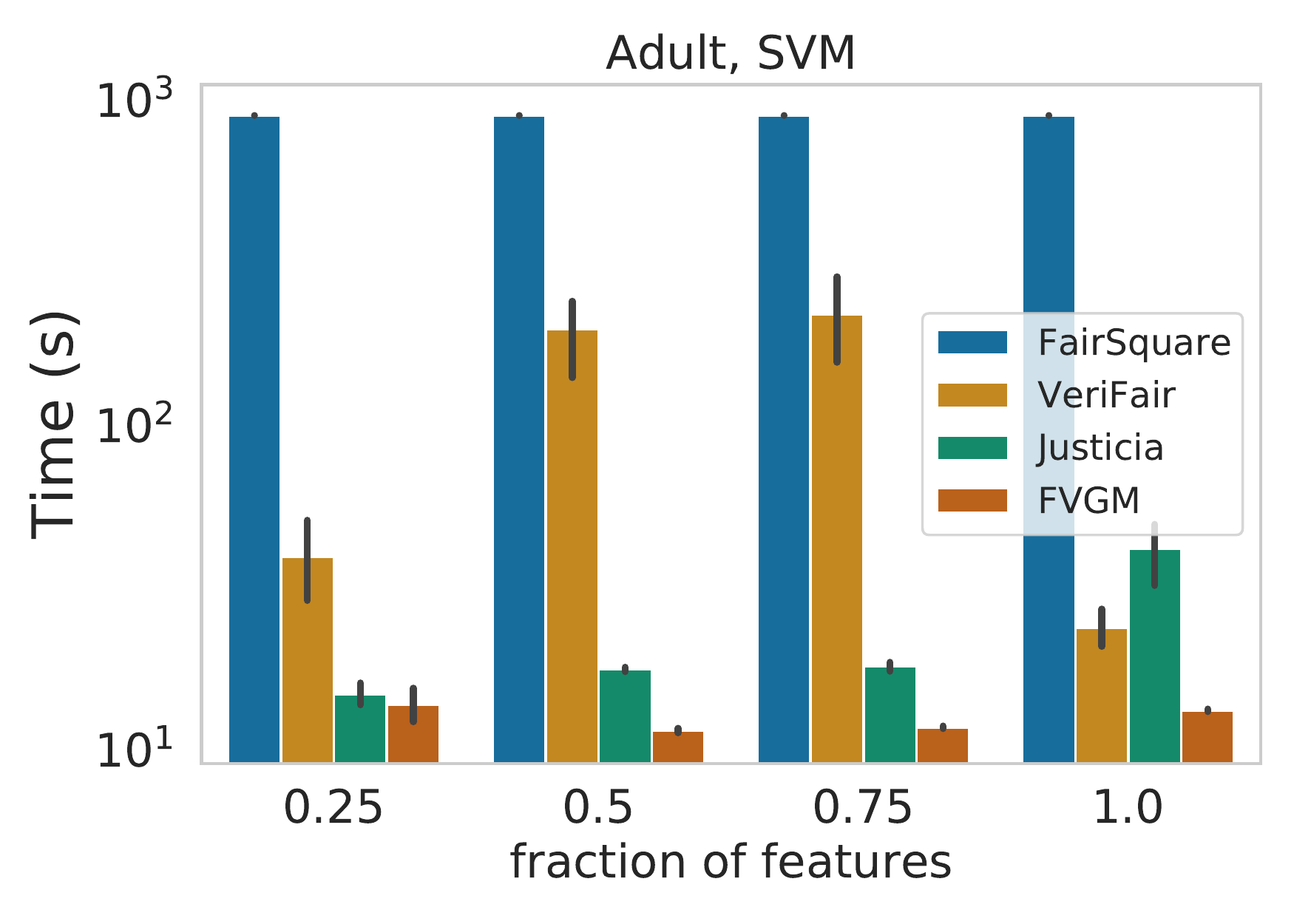}}	\\		
			\subfloat{\includegraphics[scale=0.3]{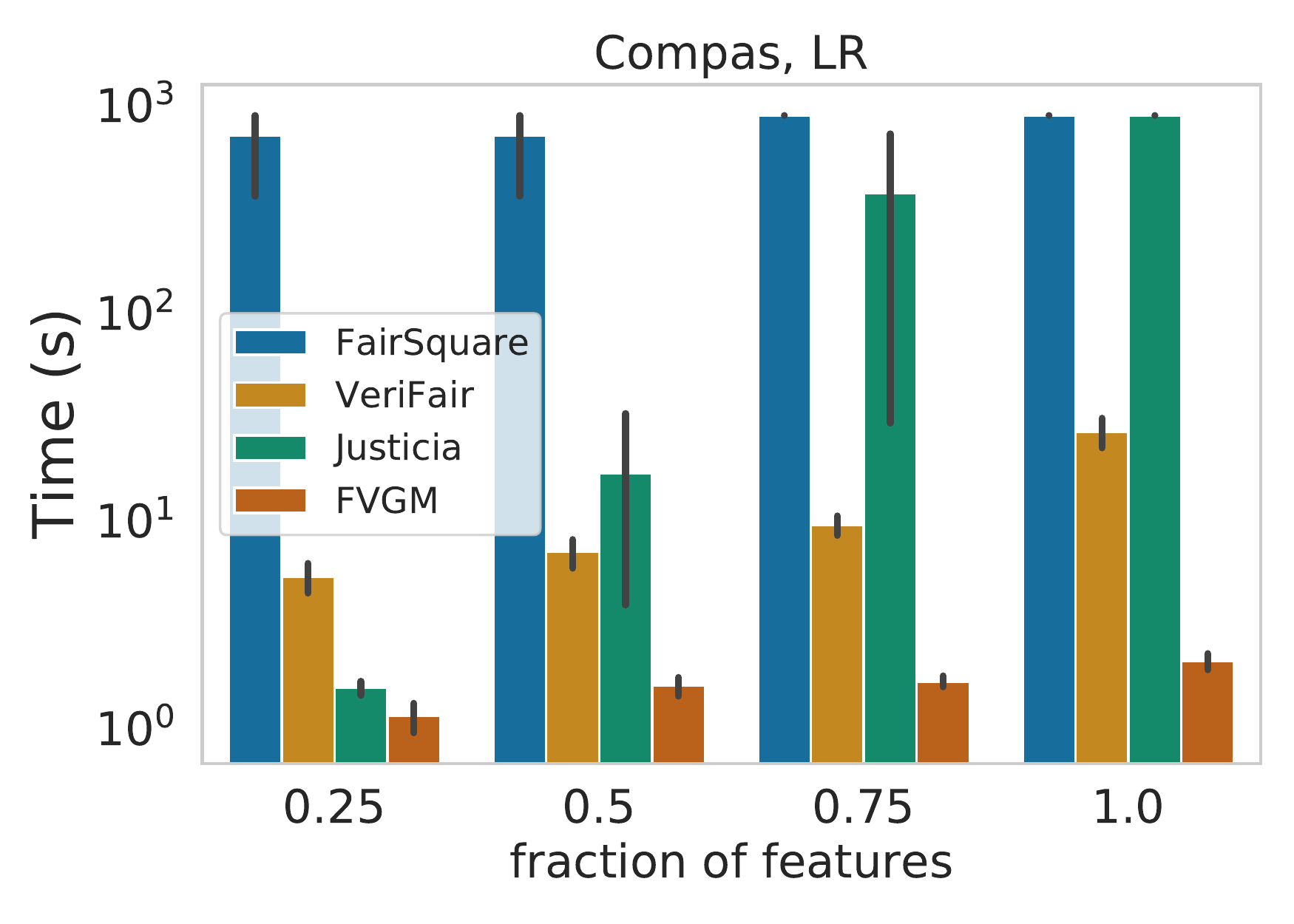}}
			\subfloat{\includegraphics[scale=0.3]{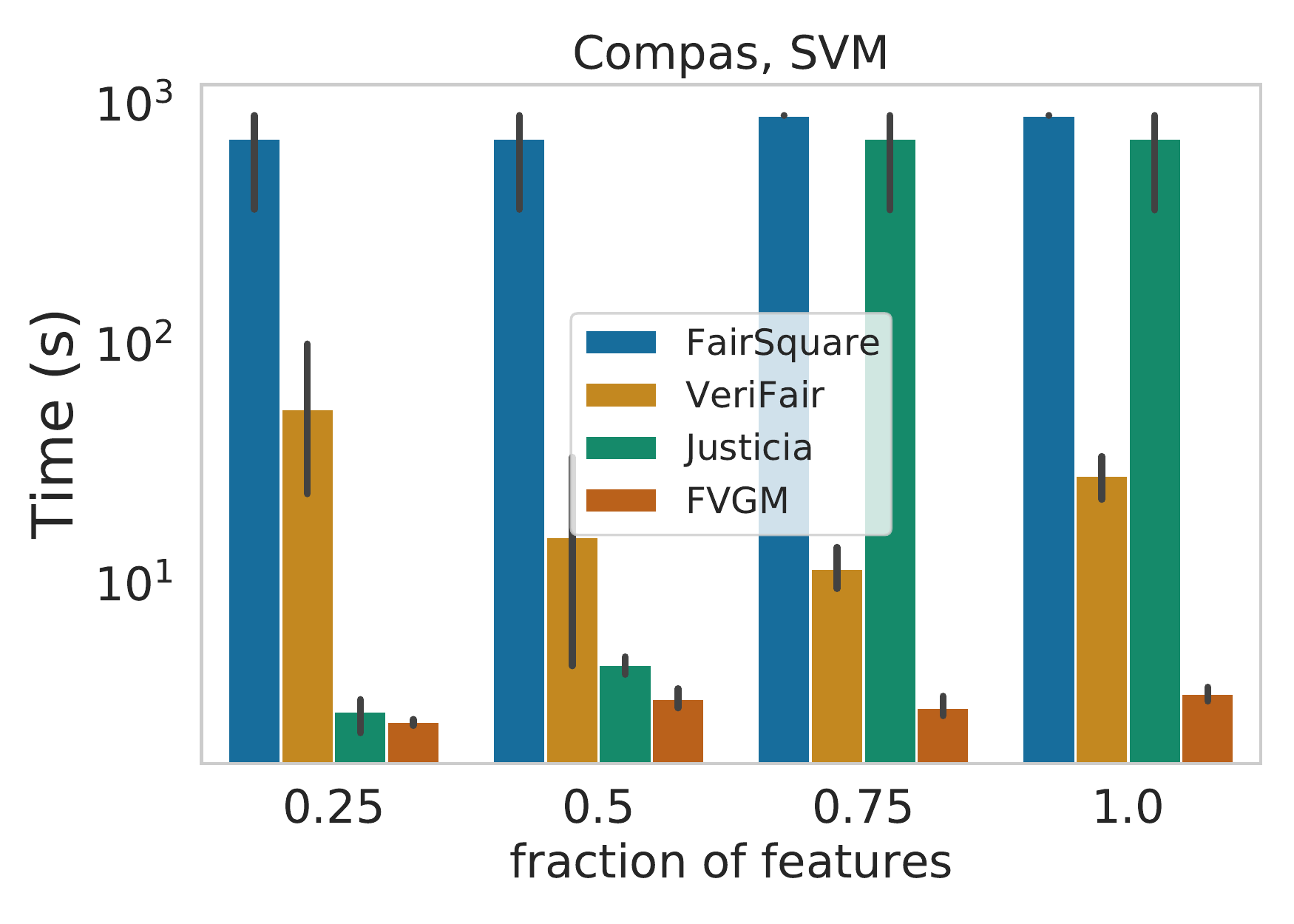}}\\
			
			\subfloat{\includegraphics[scale=0.3]{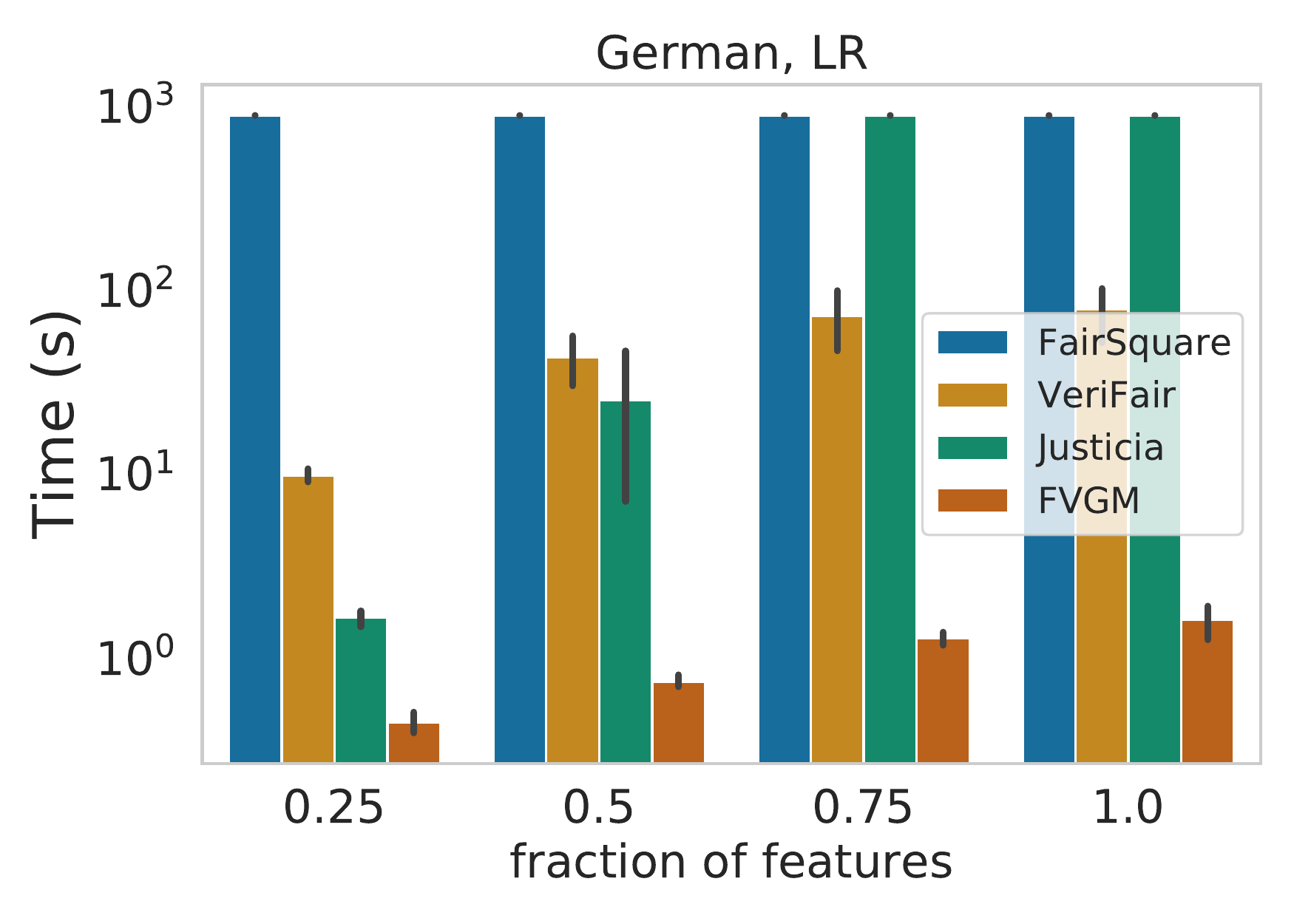}}
			\subfloat{\includegraphics[scale=0.3]{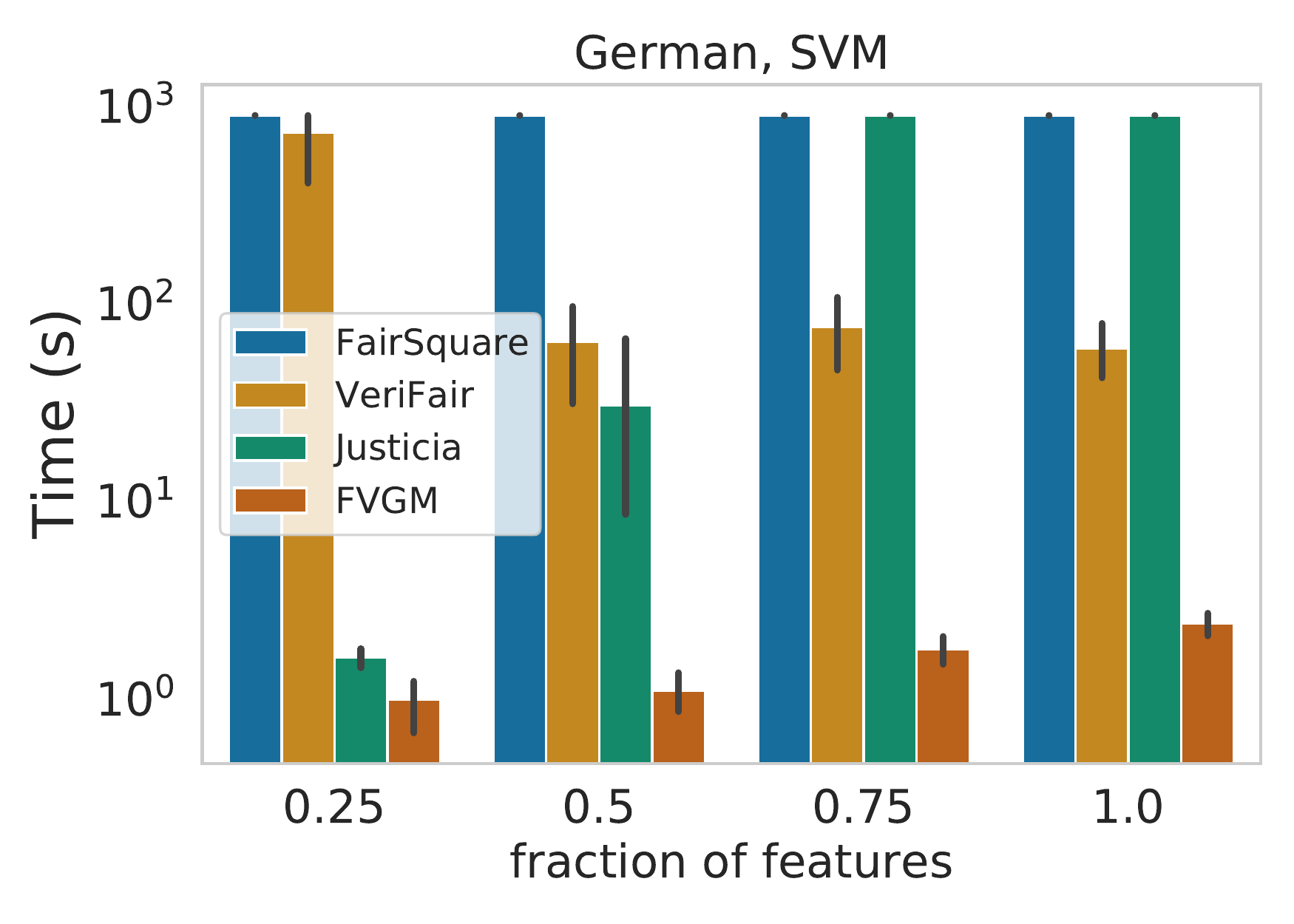}}		\\
			\subfloat{\includegraphics[scale=0.3]{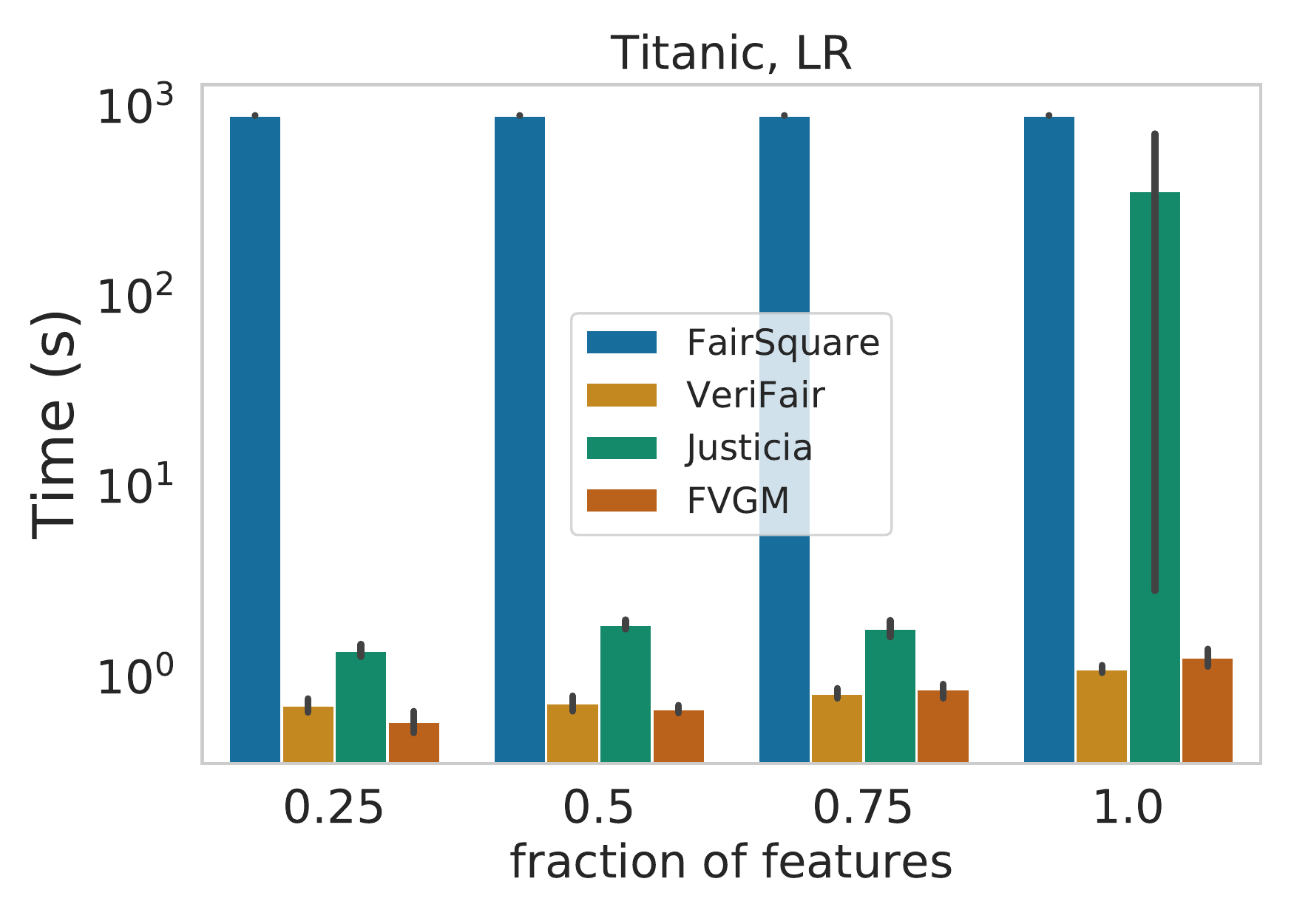}}
			\subfloat{\includegraphics[scale=0.3]{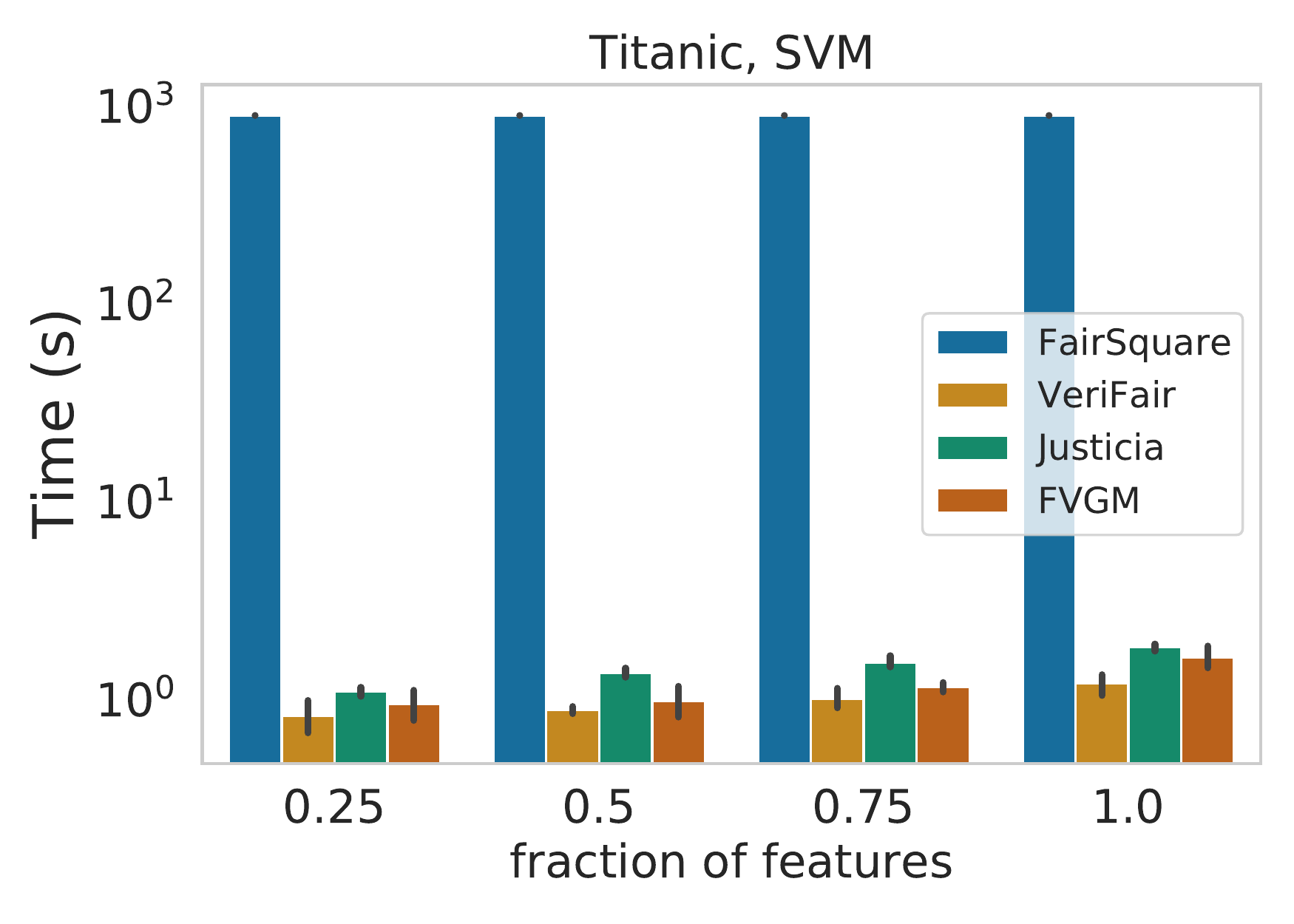}}\\
			
			\subfloat{\includegraphics[scale=0.3]{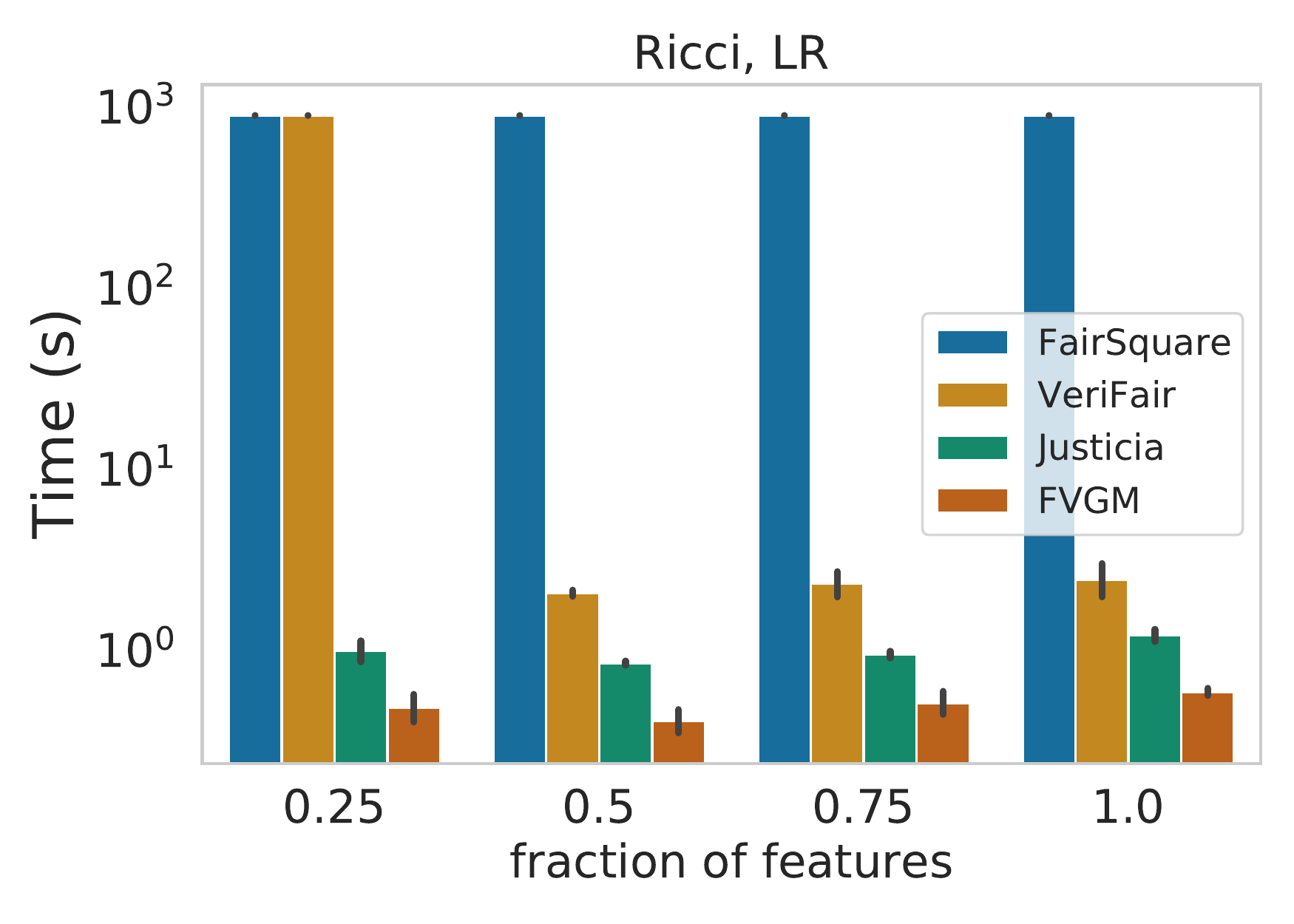}}
			\subfloat{\includegraphics[scale=0.3]{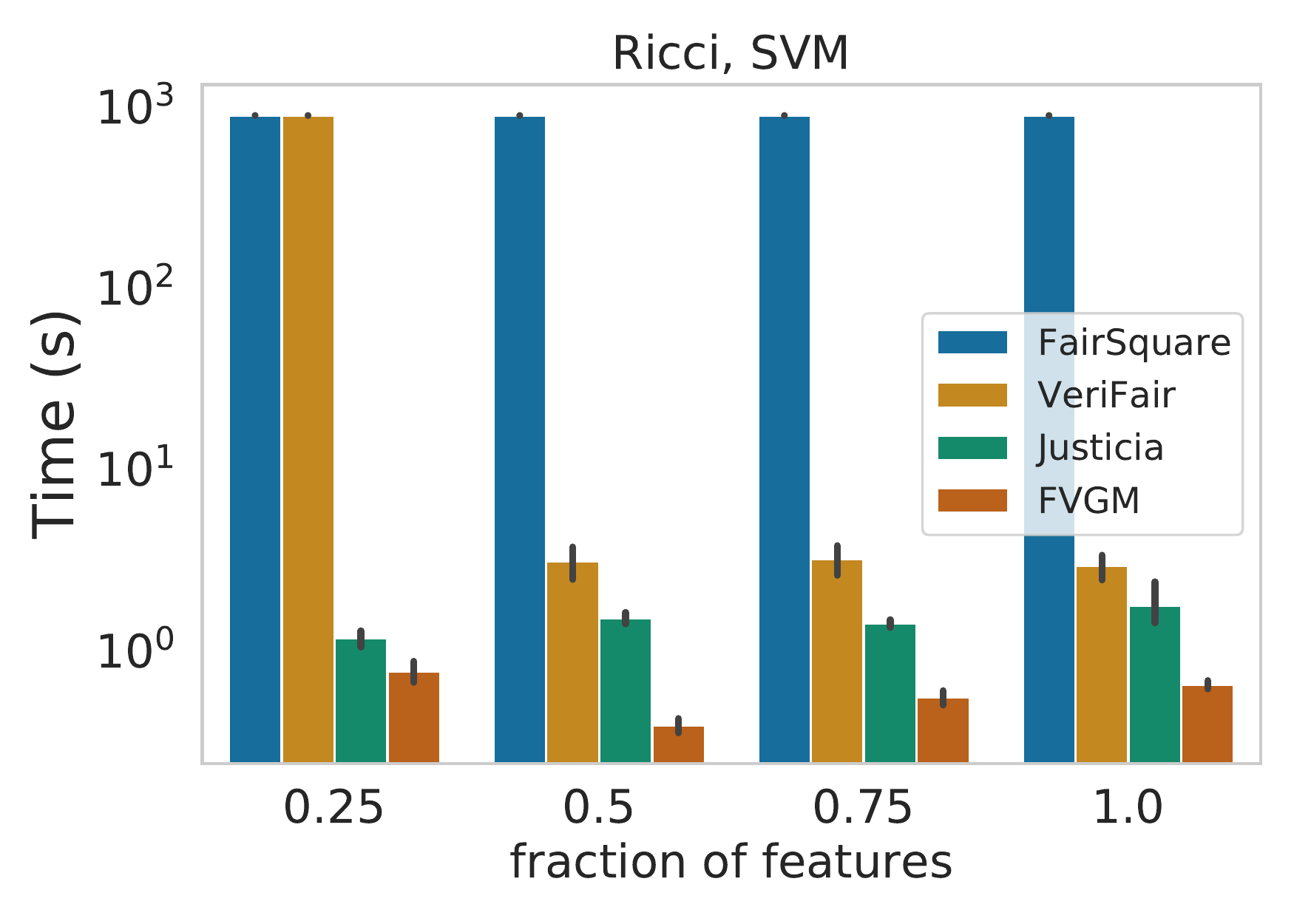}}
			
			\caption{Effect of number of features on the runtime of different datasets for LR and SVM classifiers.}
			\label{fig:time_vary_features}

		\end{center}

	\end{figure}

	\subsection{Scalability Comparison Among Different Verifiers}
	
	In Figure~\ref{fig:time_vary_features}, we present the runtime of different fairness verifiers while varying the number of features in different datasets. We observe that with an increase of features, the runtime increases in general.

	\begin{figure}[!t]
	\begin{center}
		\subfloat{\includegraphics[scale=0.3]{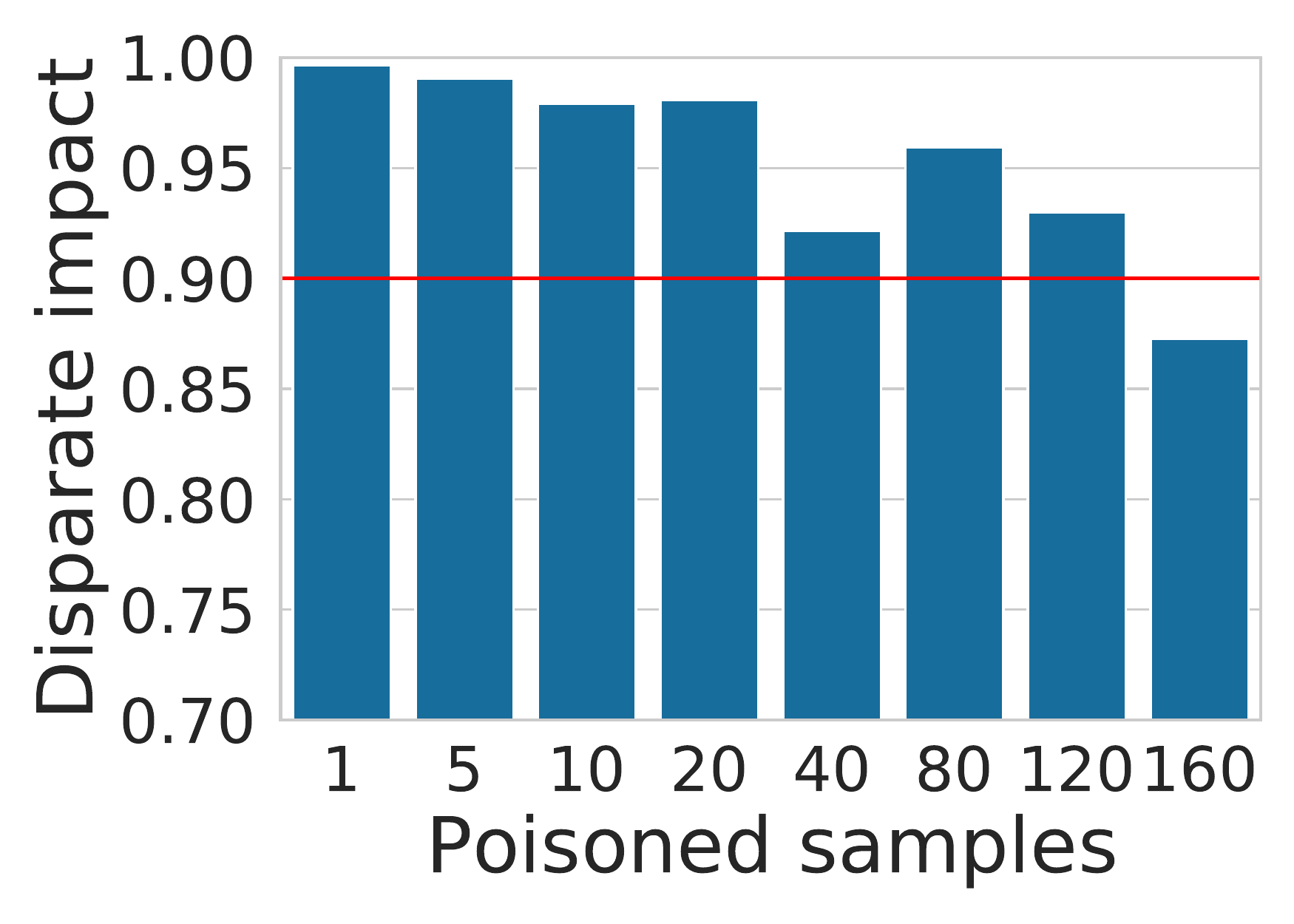}}
		\subfloat{\includegraphics[scale=0.3]{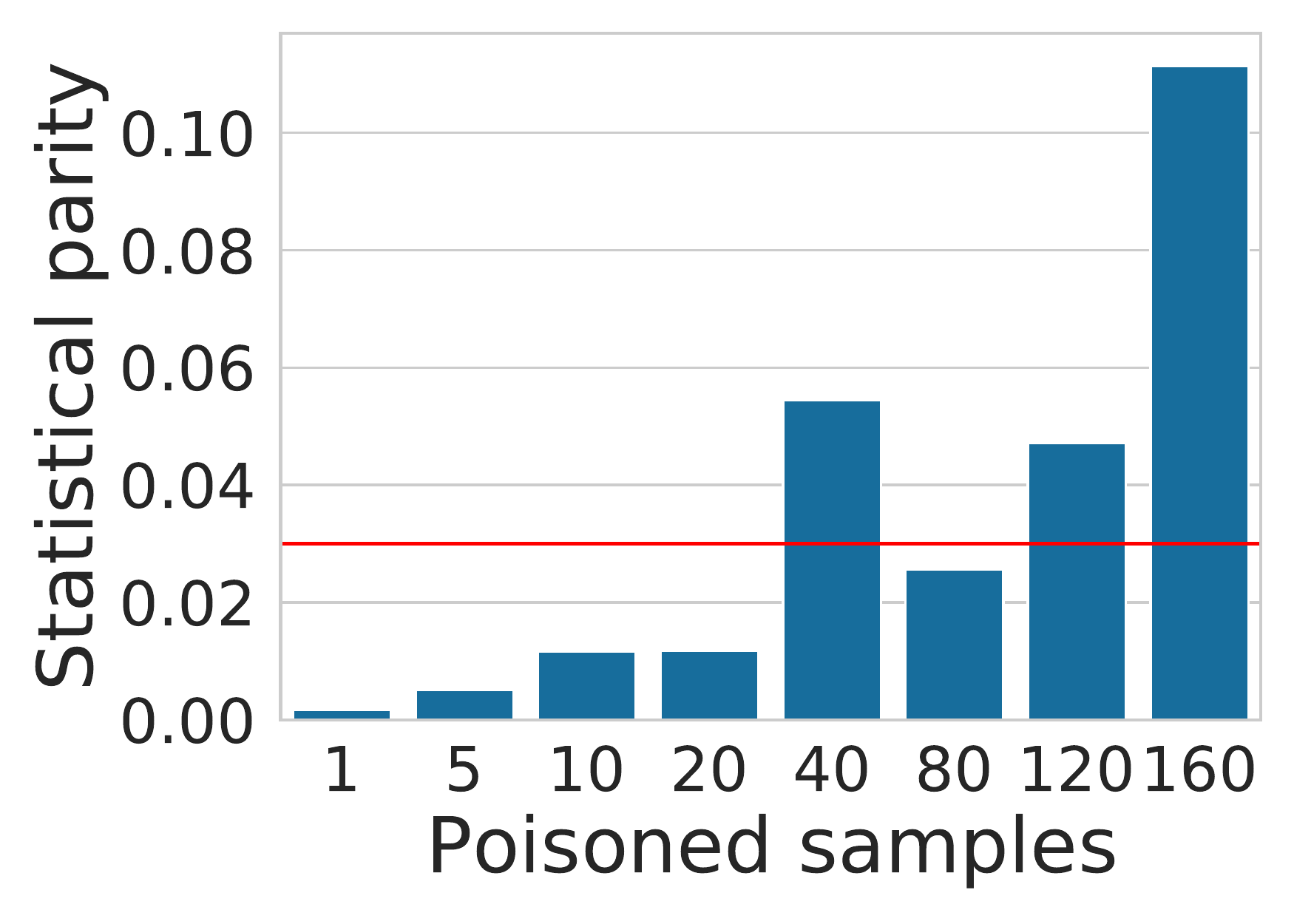}}\\
	\end{center}
	\caption{Verifying fairness poisoning attack using {\framework}. The red line denotes safety margin, which being exceeded denotes system-vulnerability by the attack algorithm. As the number of poisoned samples increase, disparate impact (DI) decreases and statistical parity (SP)  increases.}
	\label{fig:attack_extended}
\end{figure}

	\begin{figure*}[!t]
	\begin{center}
		\subfloat{\includegraphics[scale=0.33]{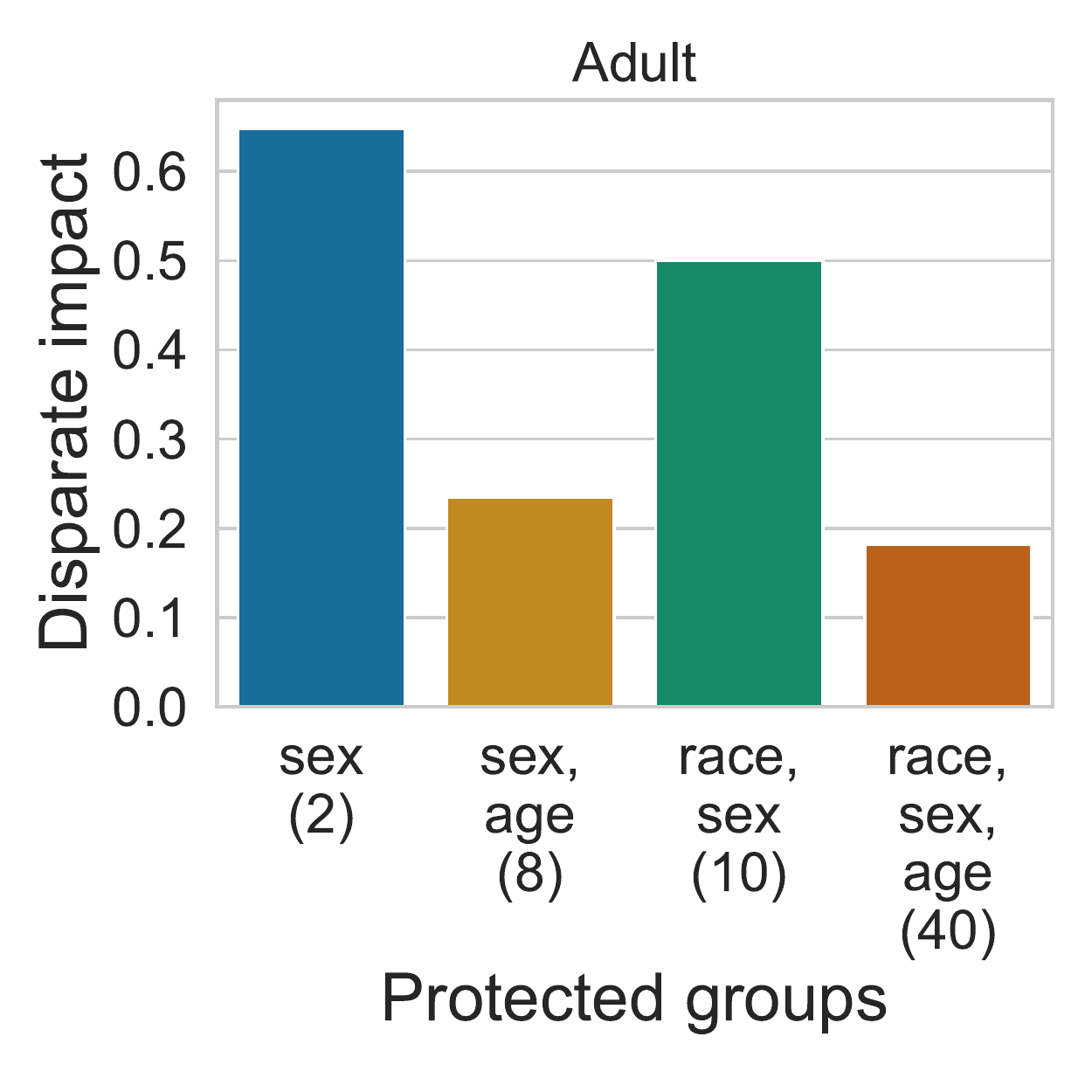}}
		\subfloat{\includegraphics[scale=0.33]{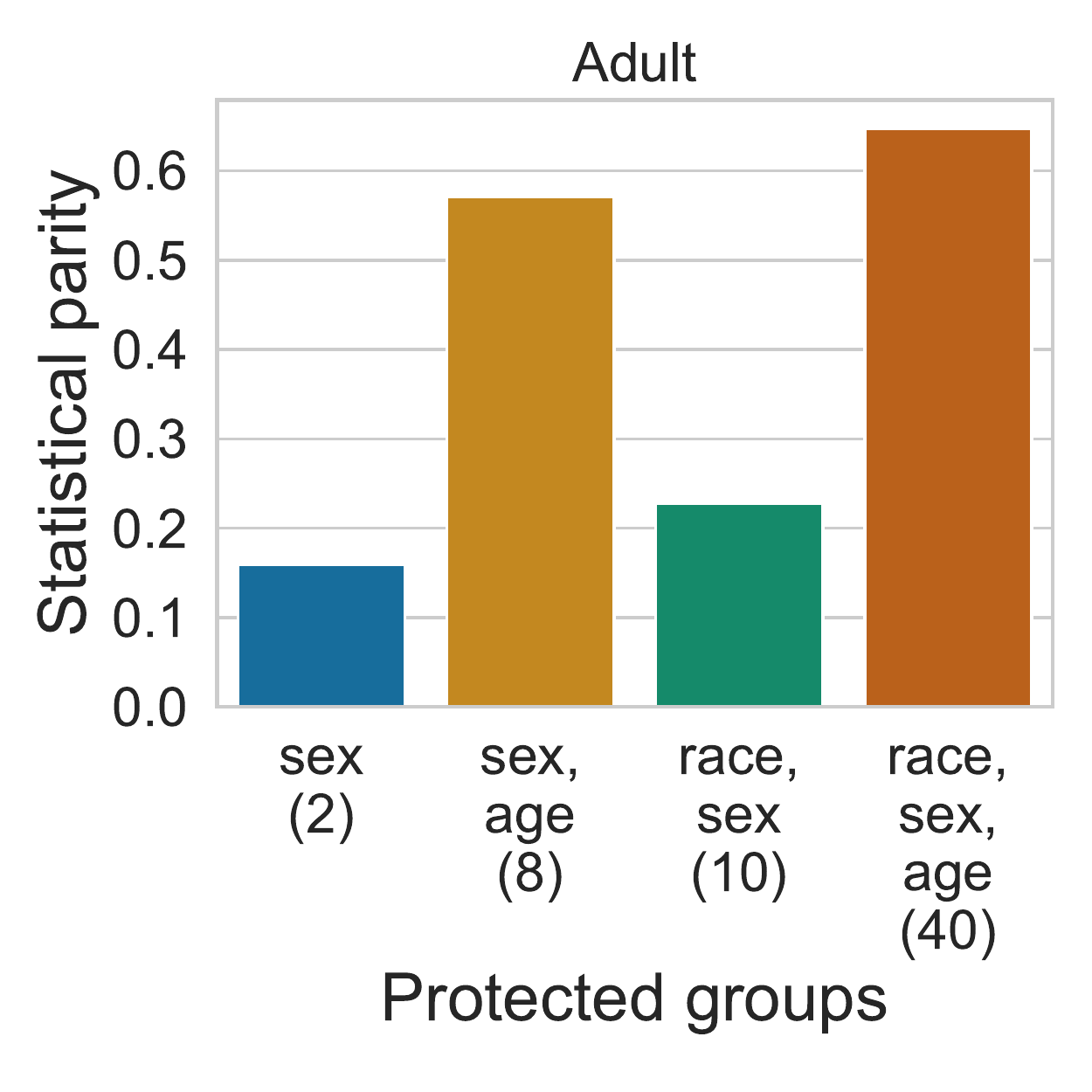}}\\
		\subfloat{\includegraphics[scale=0.33]{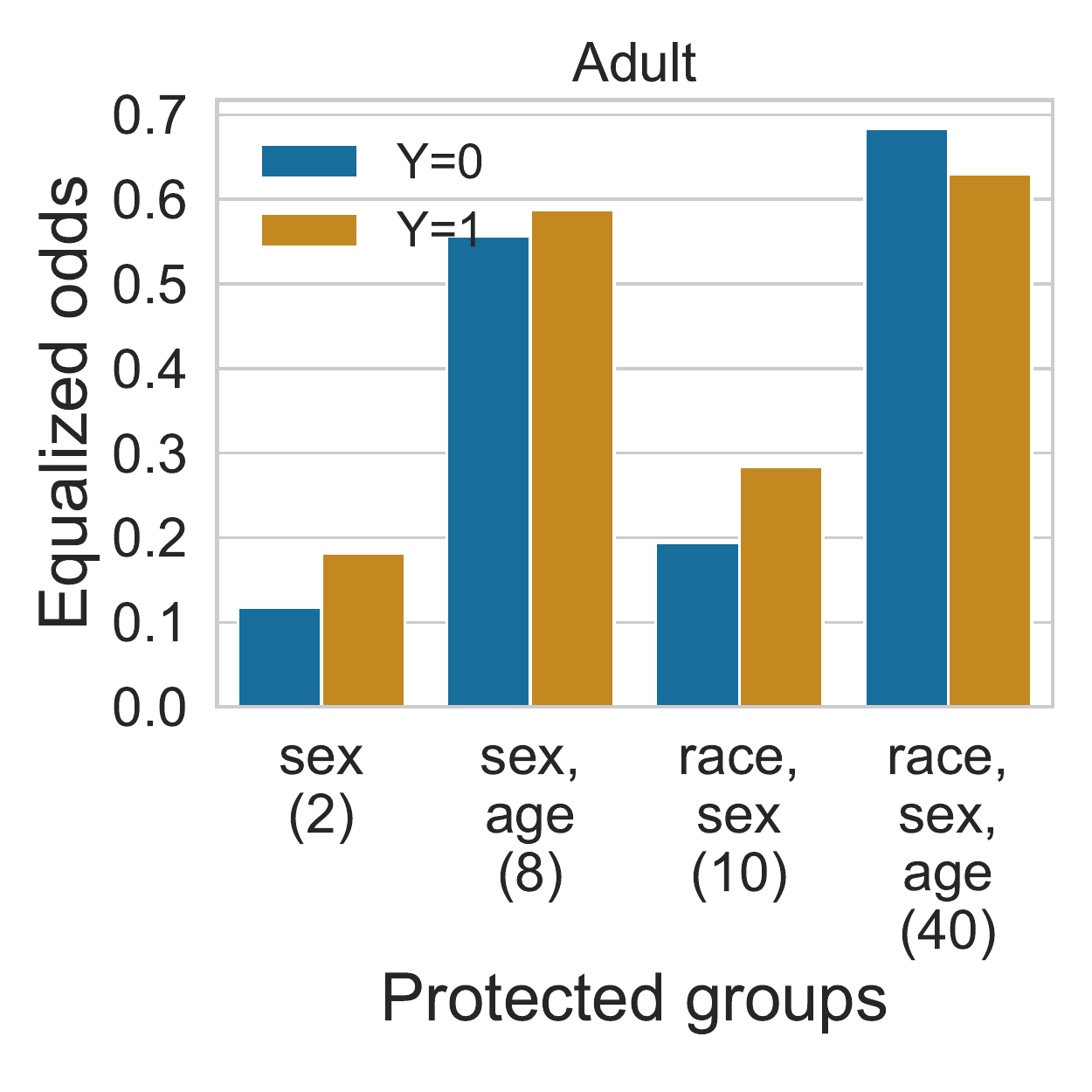}}
		\subfloat{\includegraphics[scale=0.33]{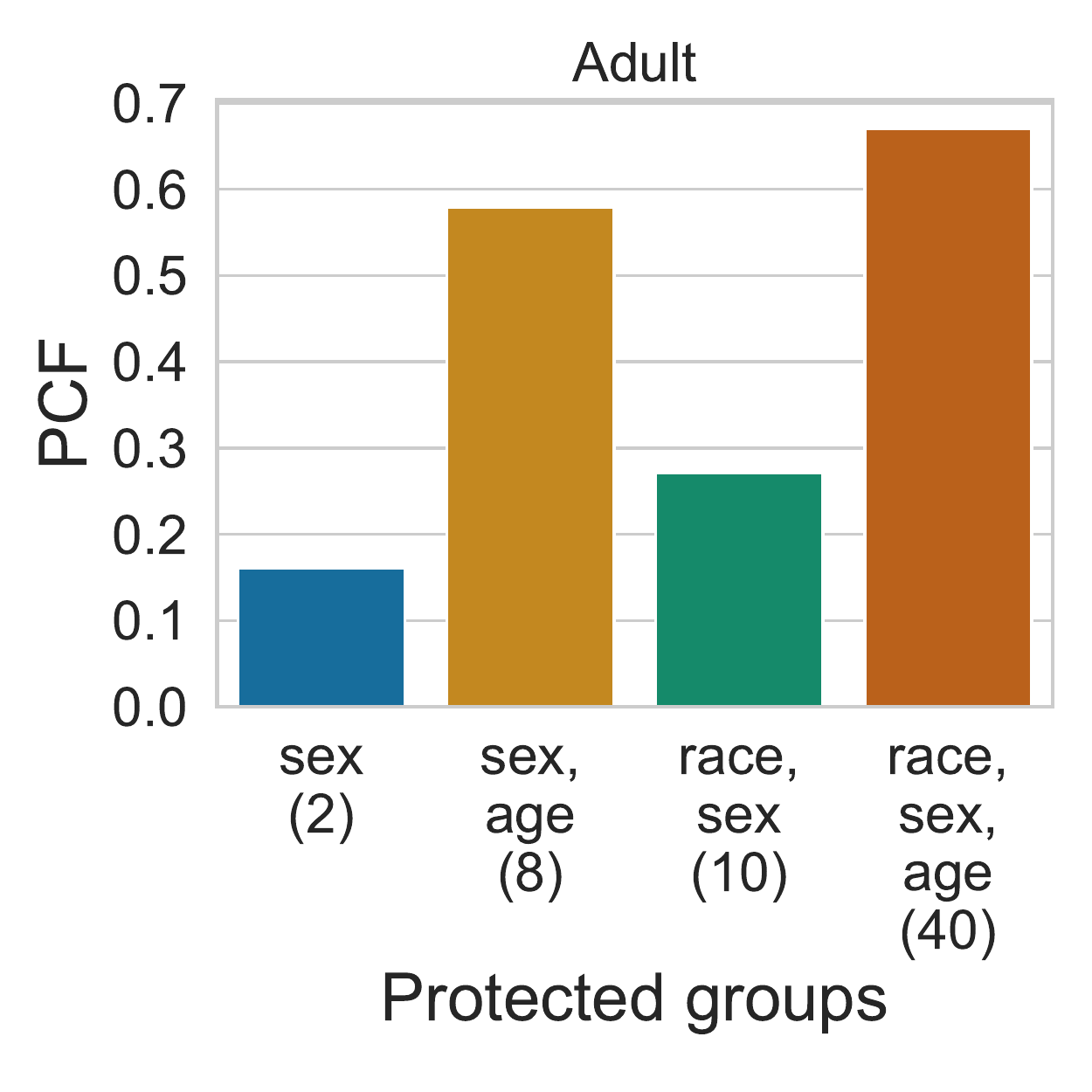}}	
		\\
		\subfloat{\includegraphics[scale=0.33]{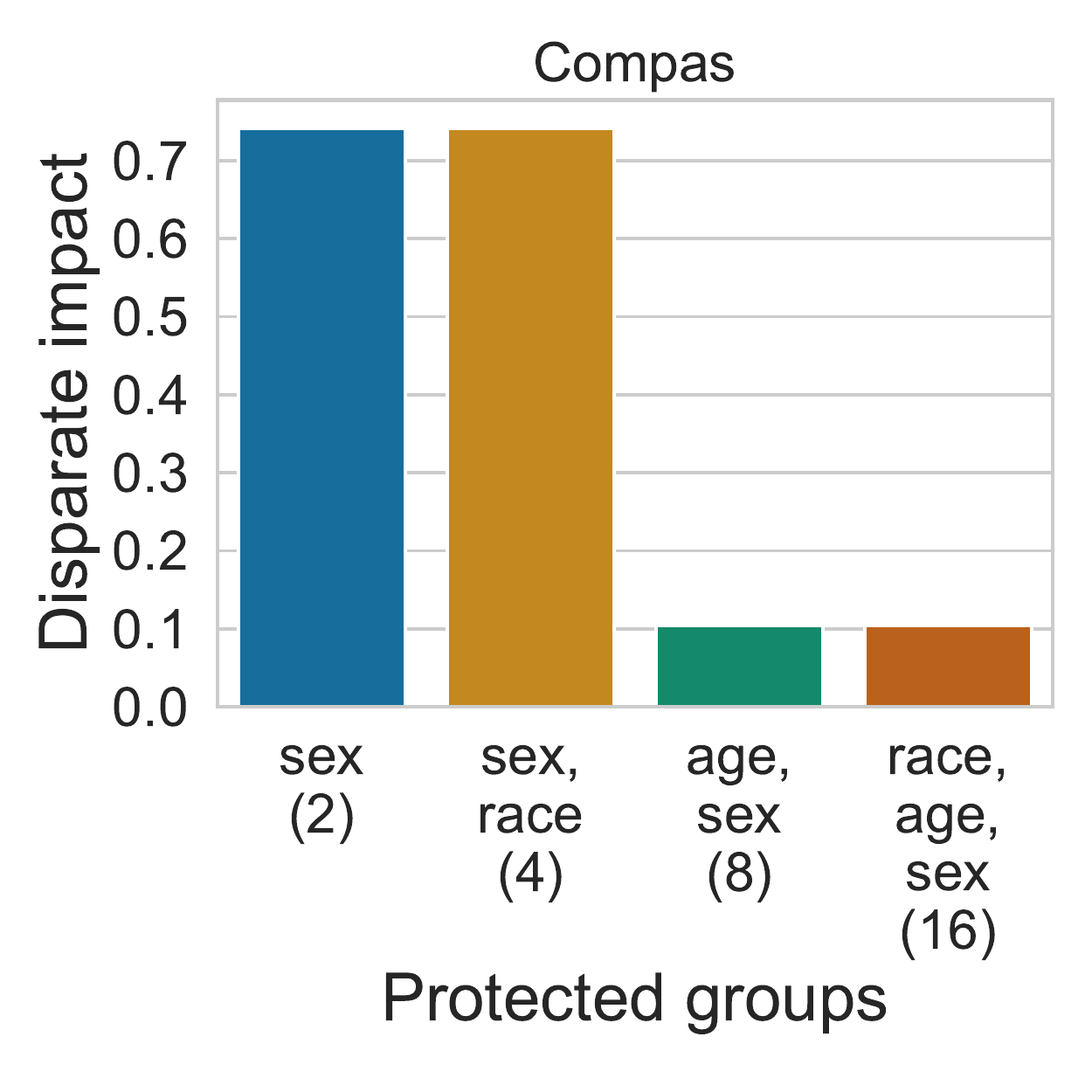}}			
		\subfloat{\includegraphics[scale=0.33]{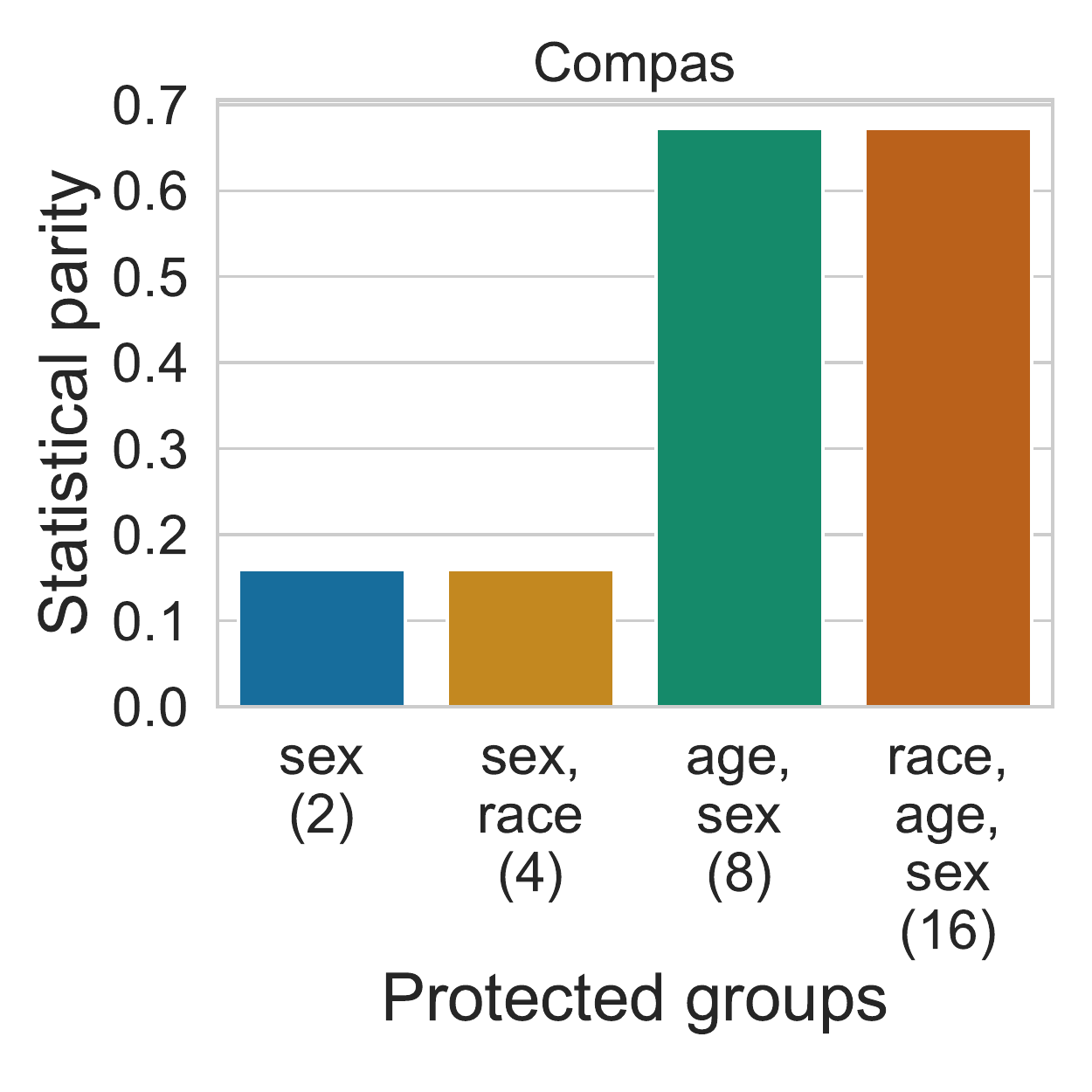}}\\
		\subfloat{\includegraphics[scale=0.33]{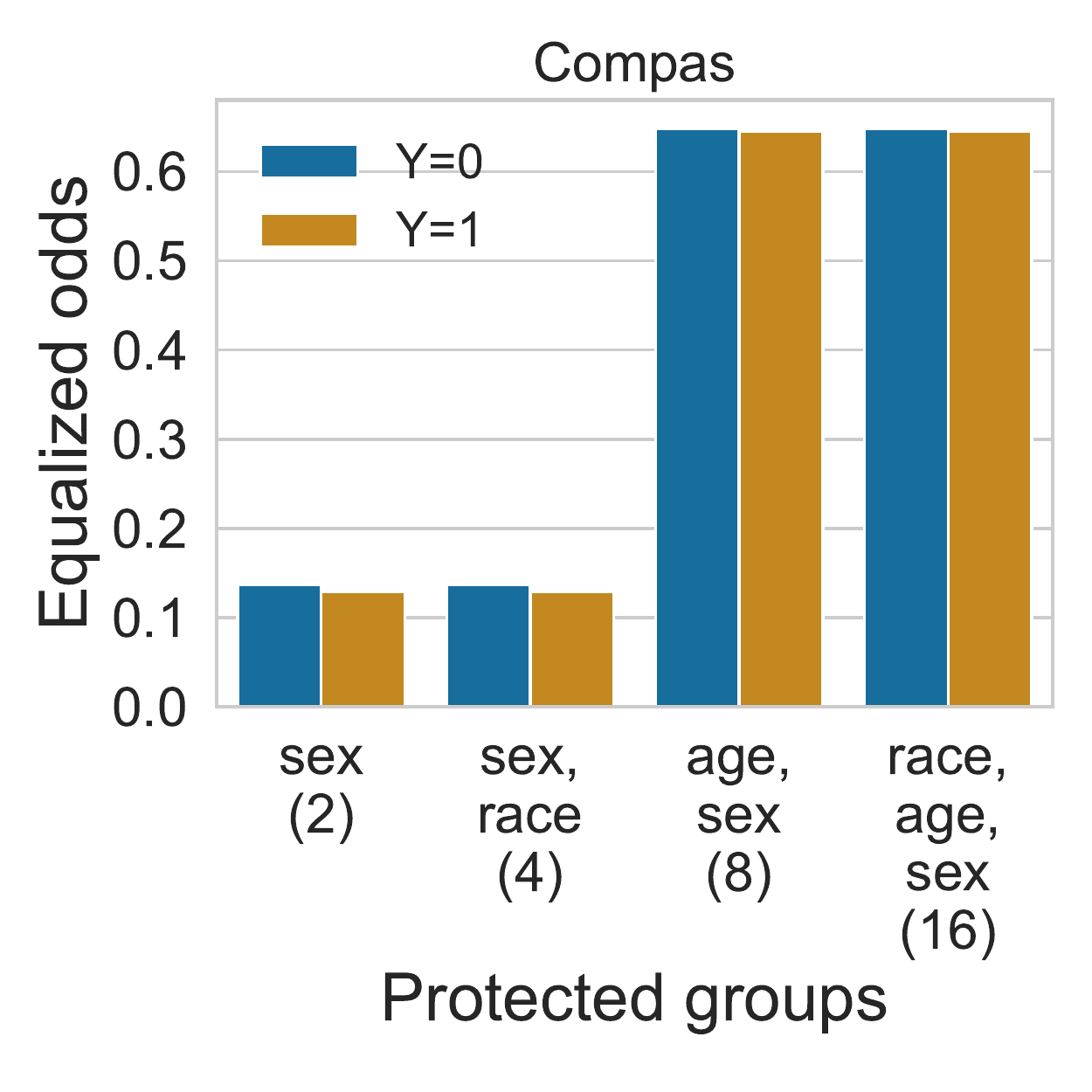}}
		\subfloat{\includegraphics[scale=0.33]{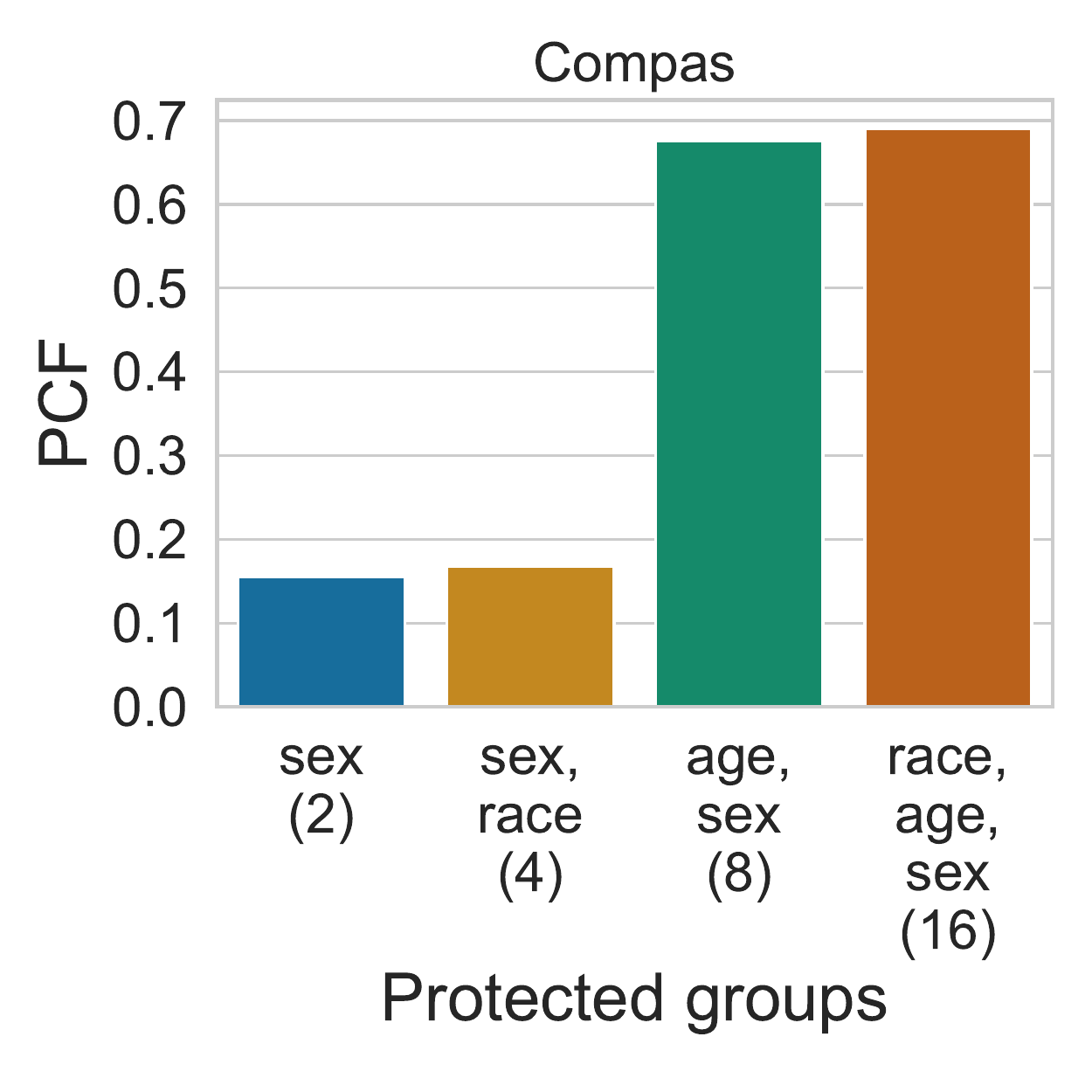}}	
		
	\end{center}
		\caption{Verifying compound sensitive groups with respect to multiple fairness metrics. In each plot, the $ X $-axis shows sensitive features with the number of compound groups (within parenthesis) and $ Y $-axis shows computed group and causal fairness metrics. }
		\label{fig:multiple_metrics}

\end{figure*}

	\begin{figure}[!t]
		\begin{center}
%
%
%
%
%
	
	\subfloat{\includegraphics[scale=0.3]{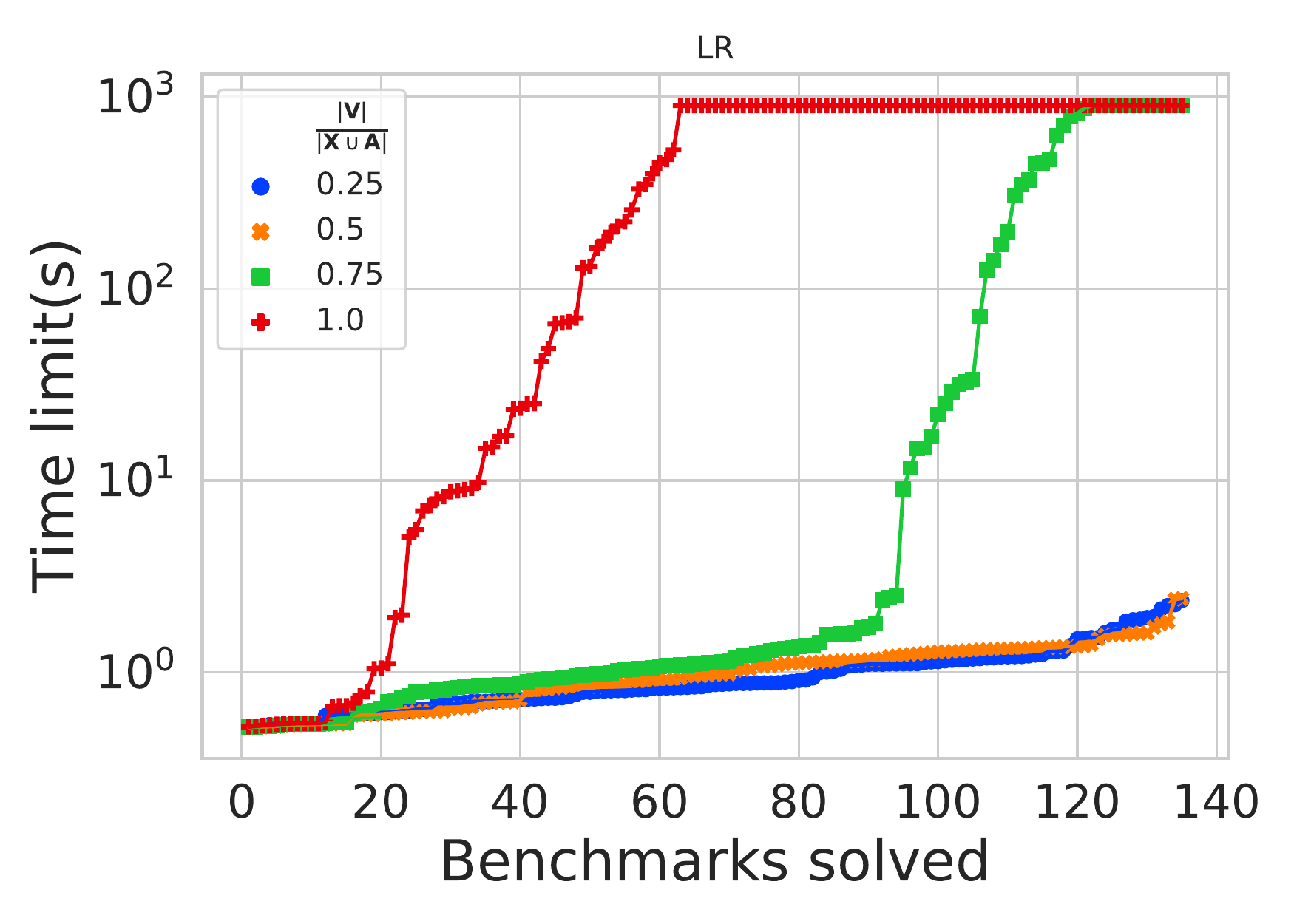}}
	\subfloat{\includegraphics[scale=0.3]{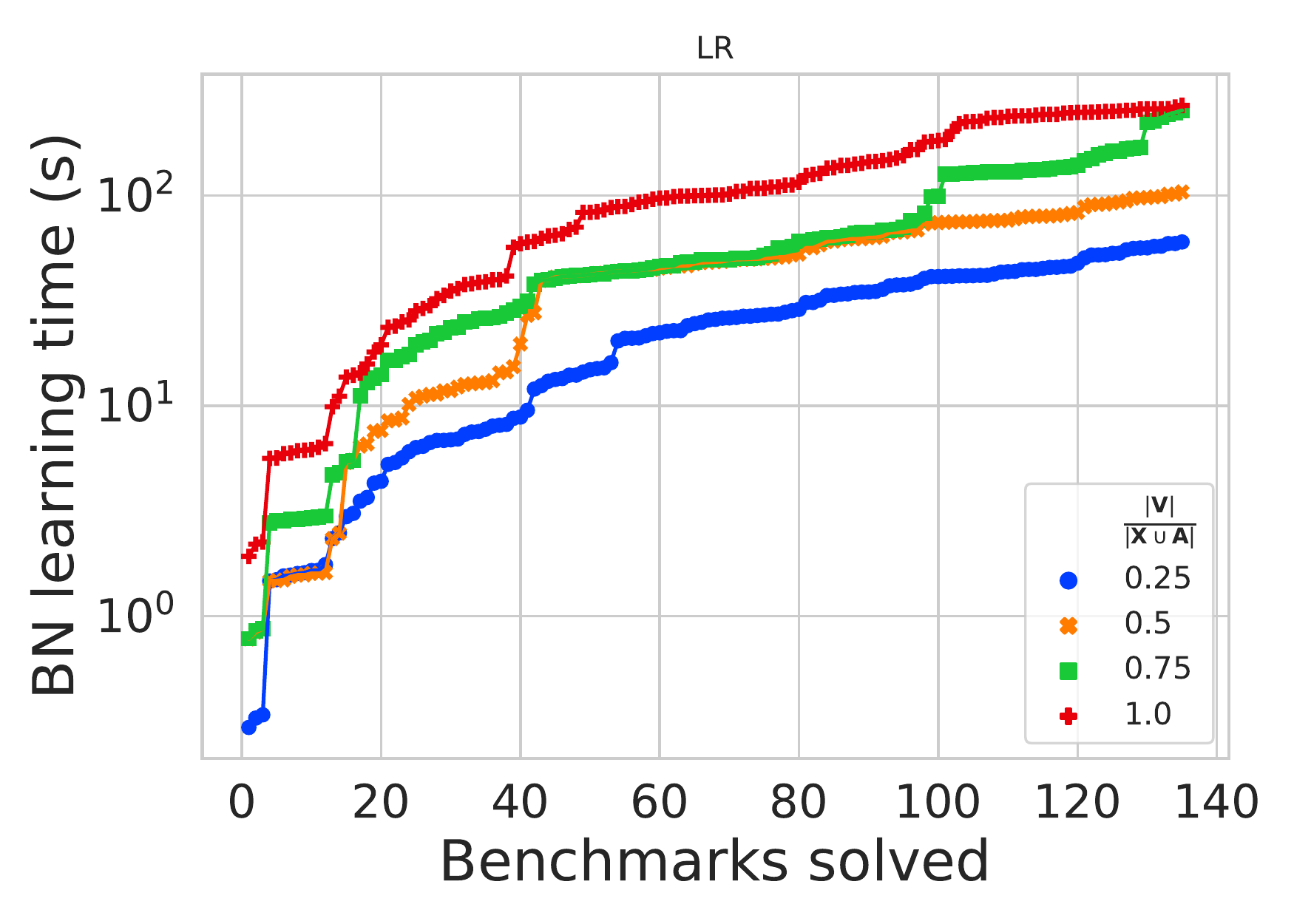}}
			
		\end{center}
		
		\caption{Effect of number of variables in the learned Bayesian Network on computation time of {\framework}. In both plots, we vary $ \frac{|\mathbf{V}|}{|\nonsensitive \cup \sensitive|} $, that is the ratio between the number of variables in the Bayesian Network to the number of features. We observe that as this ratio increases to $ 1 $, both runtime of {\framework} (left plot) and network learning time (right plot) increase. } 
		\label{fig:results_DAG_complexity}
\end{figure}

	\subsection{Verifying Fairness Algorithms on Multiple Fairness Metrics}
	
	We show extended results on verifying fairness attack in Figure~\ref{fig:attack_extended} for two fairness metrics: disparate impact (DI) and statistical parity (SP). We observe that {\framework} can detect poisoning attack for both metrics. 
	
	In Figure~\ref{fig:multiple_metrics} we show verification results on compound sensitive groups with respect to multiple fairness metrics. In this figure, we observe that with an increase in the number of groups, fairness metrics worsens\textemdash disparate impact decreases and other three metrics increases. Additionally, in Table~\ref{fig:extended_verifying_fairness_algorithms}, we show extended results for German and COMPAS dataset on verifying fairness algorithms: reweighing and optimized pre-processing algorithms. 
		
	\subsection{Performance Analysis of Bayesian Network}
	In Figure~\ref{fig:results_DAG_complexity}, we analyze the performance of encoding Bayesian Networks of differing complexity. We define the complexity of the network as  $ \frac{|V|}{|\nonsensitive \cup \sensitive|} $, which is the ratio between the number of features appearing in the network and total features. In this figure, as the ratio increases, both computation time of {\framework} and learning time of Bayesian Network increase.

	\begin{table}[!t]
	
	\centering
	\small
	\caption{Extended results for verification of different fairness enhancing algorithms using {\framework}. Numbers in bold refer to fairness improvement w.r.t. different fairness metrics. RW and OP refer to reweighing and optimized-preprocessing algorithms respectively.  }
	\label{fig:extended_verifying_fairness_algorithms}
	\setlength{\tabcolsep}{.5em}
	
	\begin{tabular}{lllrrrrrrrrrrrrr}
		
		\toprule
		Dataset & Sensitive & Algo. & $ \Delta $DI &  $ \Delta $PCF & $ \Delta $SP & $ \Delta$ EO\\
		\midrule

		\multirow{4}{*}{Adult}&\multirow{2}{*}{race}&RW&$ \textbf{0.53} $&$ \textbf{-0.06} $&$ \textbf{-0.06} $&$ \textbf{-0.02} $\\
		&&OP&$ \textbf{0.57} $&$ \textbf{-0.07} $&$ \textbf{-0.07} $&$ \textbf{-0.02} $\\
		\cmidrule{2-7}
		&\multirow{2}{*}{sex}&RW&$ \textbf{0.96} $&$ \textbf{-0.16} $&$ \textbf{-0.15} $&$ \textbf{-0.08} $\\
		&&OP&$ \textbf{0.43} $&$ \textbf{-0.08} $&$ \textbf{-0.08} $&$ 0.03 $\\
		
		\midrule
		\multirow{4}{*}{COMPAS}&\multirow{2}{*}{race}&RW&$ \textbf{0.13} $&$ \textbf{-0.07} $&$ \textbf{-0.07} $&$ \textbf{-0.06} $\\
		&&OP&$ \textbf{0.15} $&$ \textbf{-0.08} $&$ \textbf{-0.08} $&$ \textbf{-0.05} $\\
		\cmidrule{2-7}
		&\multirow{2}{*}{sex}&RW&$ \textbf{0.1} $&$ \textbf{-0.04} $&$ \textbf{-0.04} $&$ 0.04 $\\
		&&OP&$ \textbf{0.09} $&$ \textbf{-0.04} $&$ \textbf{-0.04} $&$ \textbf{-0.03} $\\
		
		\midrule
		\multirow{4}{*}{German}&\multirow{2}{*}{age}&RW&$ \textbf{0.52} $&$ \textbf{-0.53} $&$ \textbf{-0.52} $&$ \textbf{-0.47} $\\
		&&OP&$ \textbf{0.53} $&$ \textbf{-0.53} $&$ \textbf{-0.53} $&$ \textbf{-0.51} $\\
		\cmidrule{2-7}
		&\multirow{2}{*}{sex}&RW&$ -0.06 $&$ 0.06 $&$ 0.06 $&$ 0.02 $\\
		&&OP&$ -0.12 $&$ 0.12 $&$ 0.12 $&$ 0.07 $\\

		\bottomrule
	\end{tabular}
\end{table}

\section{Fairness Influence Functions (FIF)}
Now, we present an elaborate discussion on computing fairness influence functions.
A Fairness Influence Function, denoted as $ \mathsf{FIF}(\cdot) $, is computed with respect to a \textit{quantity of interest}, for example, the  probability of positive prediction of the classifier or different fairness metrics such as DI and SP. Let $ \mathbf{S}  \subseteq \nonsensitive $ be a set of non-sensitive features, for which we are interested in computing their influence. A general approach to compute $ \mathsf{FIF}(\mathbf{S}) $ is to replace each feature in $ \mathbf{S} $ with random values and report differences in the quantity of interest~\cite{datta2016algorithmic}. Let $ \mathcal{X}'_i $ denote the modified marginal distribution where we replace feature $ X_i \in \mathbf{S} $ with random values. For example, $ \mathcal{X}'_i $ can be viewed as an uniform distribution within the support of $ X_i $. We then define the modified product distribution, denoted as $ \mathcal{D}_{-\mathbf{S}} $ by combining all marginal distributions, where each feature in $ \mathbf{S} $ has a modified distribution. 
\[
\mathcal{D}_{-\mathbf{S}} = \prod_{i | X_i \in \mathbf{S}} \mathcal{X}'_i \prod_{i | X_i \in X \setminus \mathbf{S}} \mathcal{X}_i \prod_{j=1}^{n} \mathcal{A}_j 
\]
We first give a general definition of influence function of $ \mathbf{S} $ on a quantity, say $ Q $, in the following. 
\[
\mathsf{FIF}(\mathbf{S}) \triangleq Q(\mathcal{D}) - Q(\mathcal{D}_{-\mathbf{S}})
\]
Intuitively, influence of $ \mathbf{S} $ is the difference in a quantity computed for the original distribution $ \mathcal{D} $ and the modified distribution $ \mathcal{D}_{-\mathbf{S}} $. In the following, we define influence function in terms of the probability of positive prediction of the classifier. 
\[
\mathsf{FIF}(\mathbf{S}) = \Pr[\hat{Y} = 1 | \mathcal{D}] - \Pr[\hat{Y} = 1 | \mathcal{D}_{-\mathbf{S}}]
\]
Influence function also generalizes to the probability of positive prediction specific to compound sensitive groups. For a sensitive group $ \mathbf{a} \in \sensitive $, we define group-specific influence as follows. 

\[
\mathsf{FIF}_{\mathbf{a}}(\mathbf{S}) \triangleq \Pr[\hat{Y} = 1 | \sensitive = \mathbf{a}, \mathcal{D}] - \Pr[\hat{Y} = 1 | \sensitive = \mathbf{a},  \mathcal{D}_{-\mathbf{S}}]
\]

\paragraph{Experimental Analysis.} We empirically instantiate computation of FIF and its consequences for a logistic regression classifier on COMPAS and Adult datasets.

\begin{figure}[t!]
	\begin{center}
	
		\subfloat[COMPAS]{\includegraphics[scale=0.3]{figures/dependency_exp_Learn-dependency_compas_lr_sex_0}\label{fig:fif_a}}
		\subfloat[COMPAS]{\includegraphics[scale=0.3]{figures/dependency_exp_Learn-dependency_compas_lr_sex_1}\label{fig:fif_b}}\\
		\subfloat[Adult]{\includegraphics[scale=0.3]{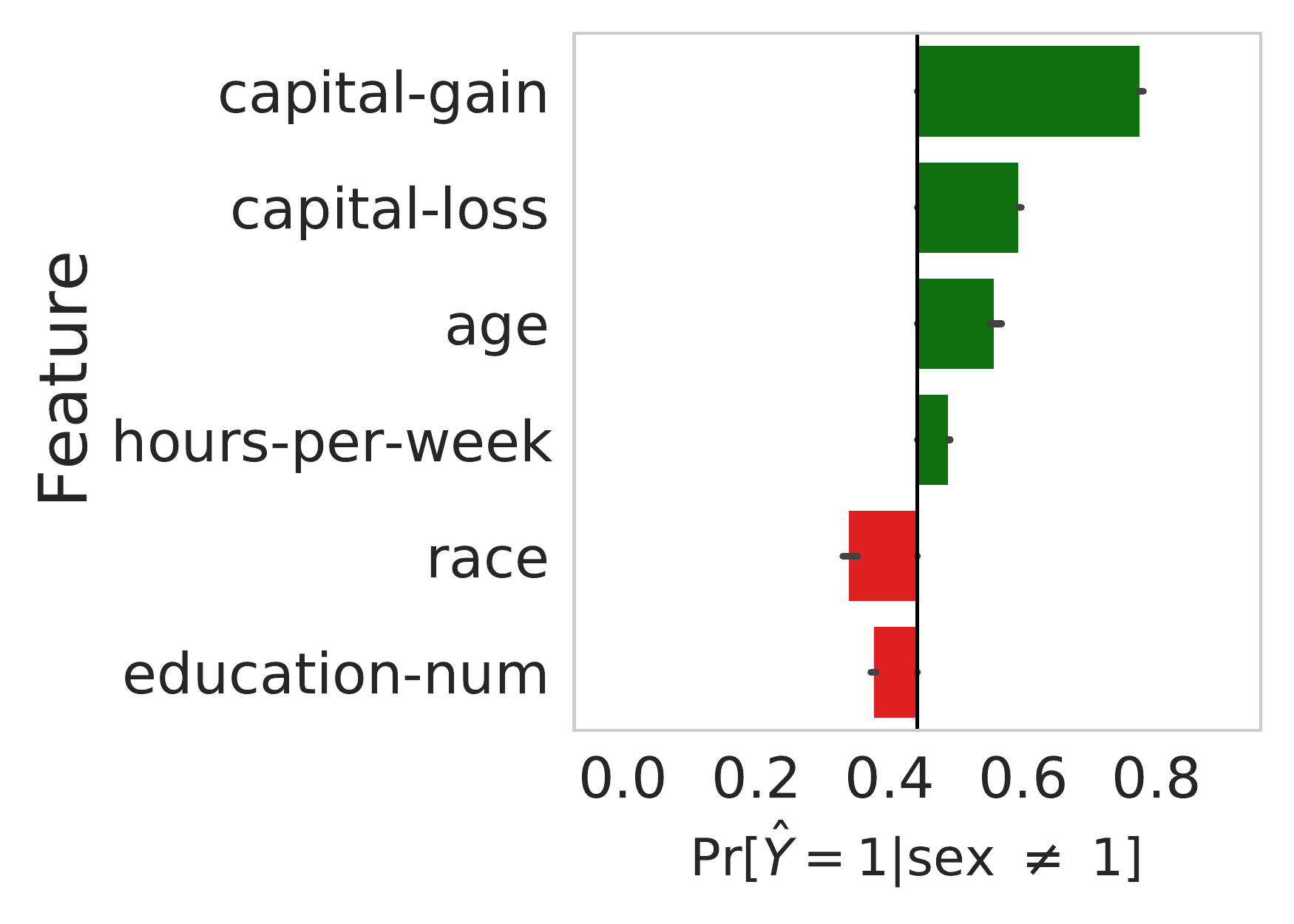}\label{fig:fif_d}}
		\subfloat[Adult]{\includegraphics[scale=0.3]{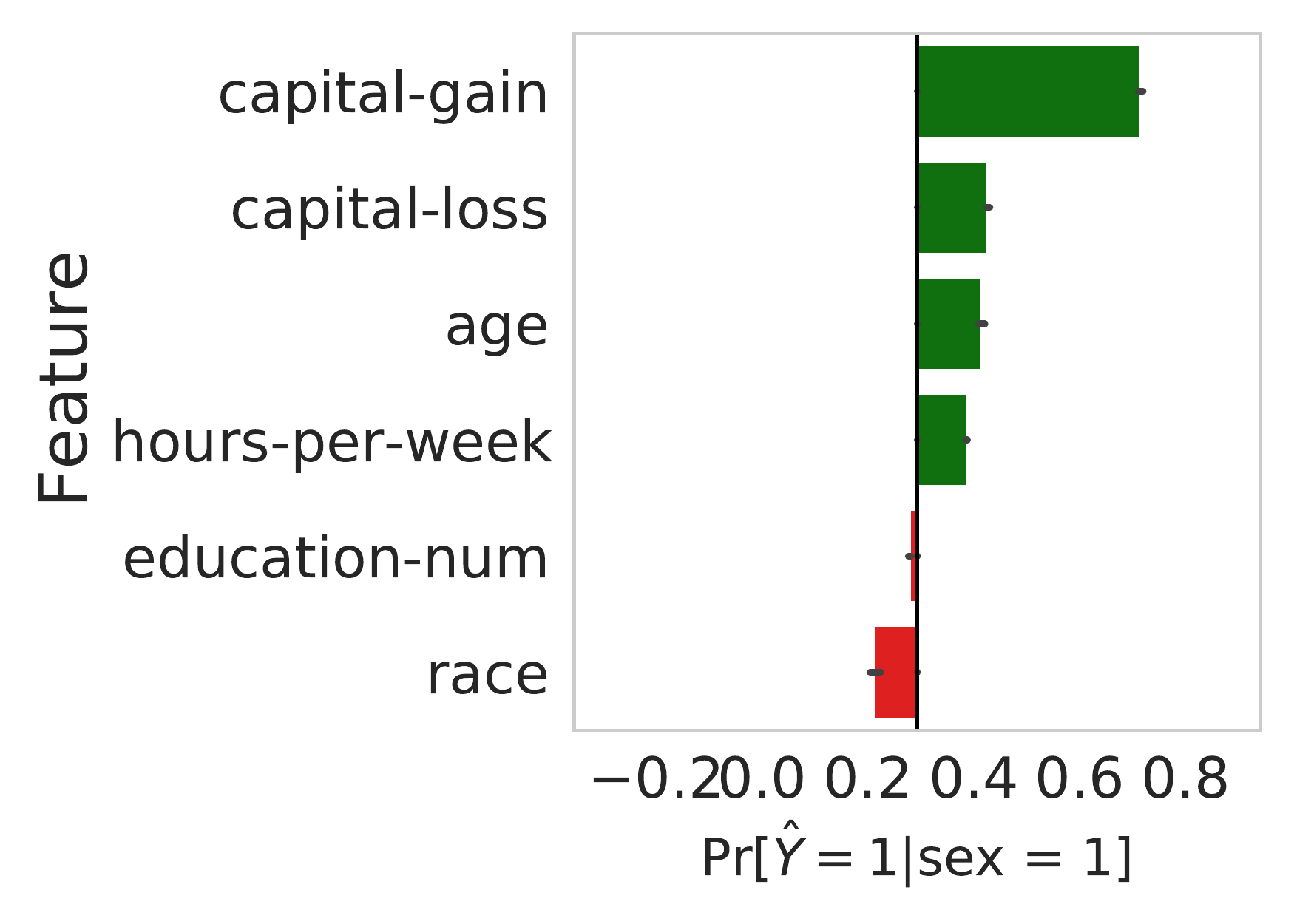}\label{fig:fif_e}}\\
		\subfloat[COMPAS]{\includegraphics[scale=0.3]{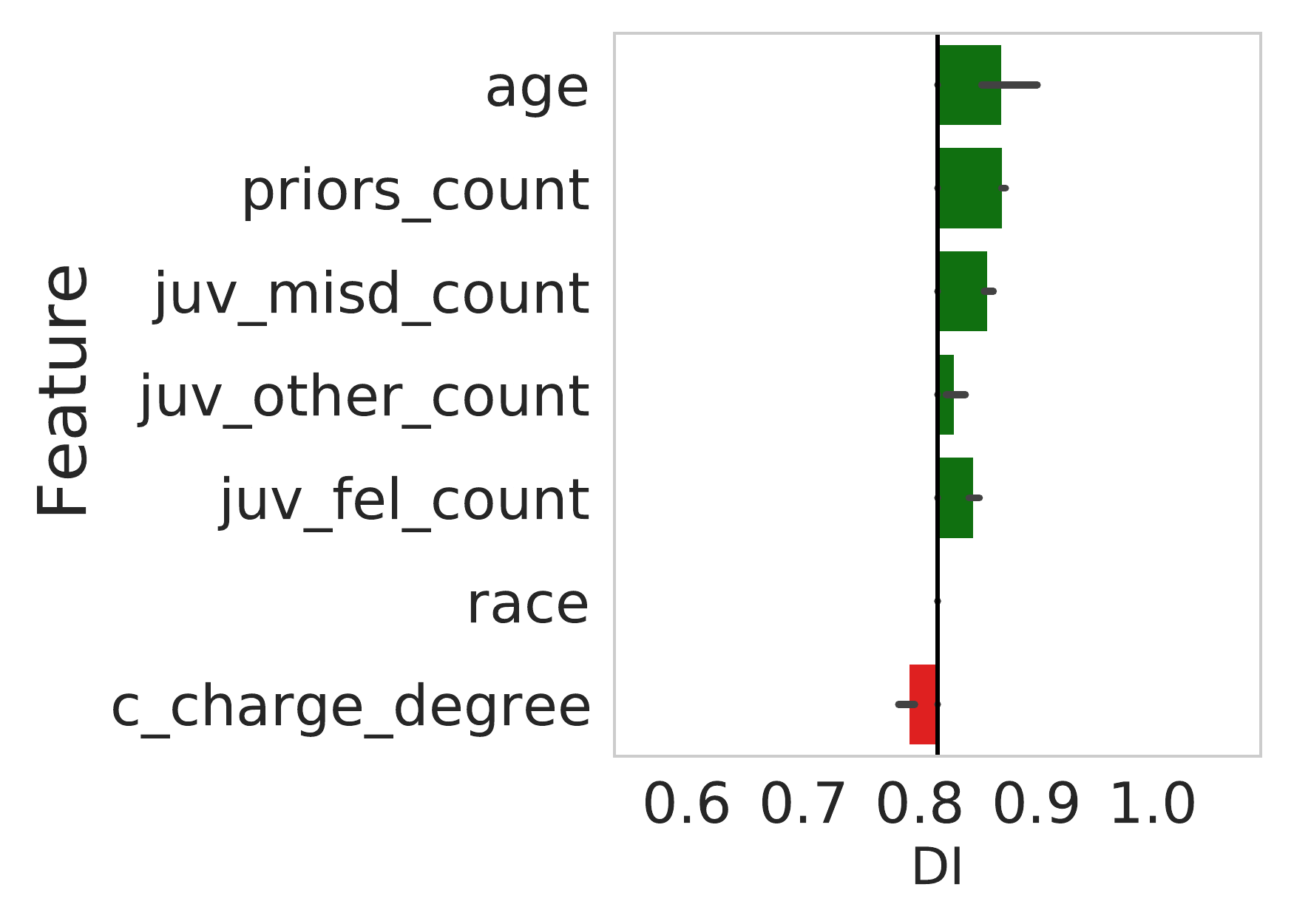}\label{fig:fif_c}}
		\subfloat[Adult]{\includegraphics[scale=0.3]{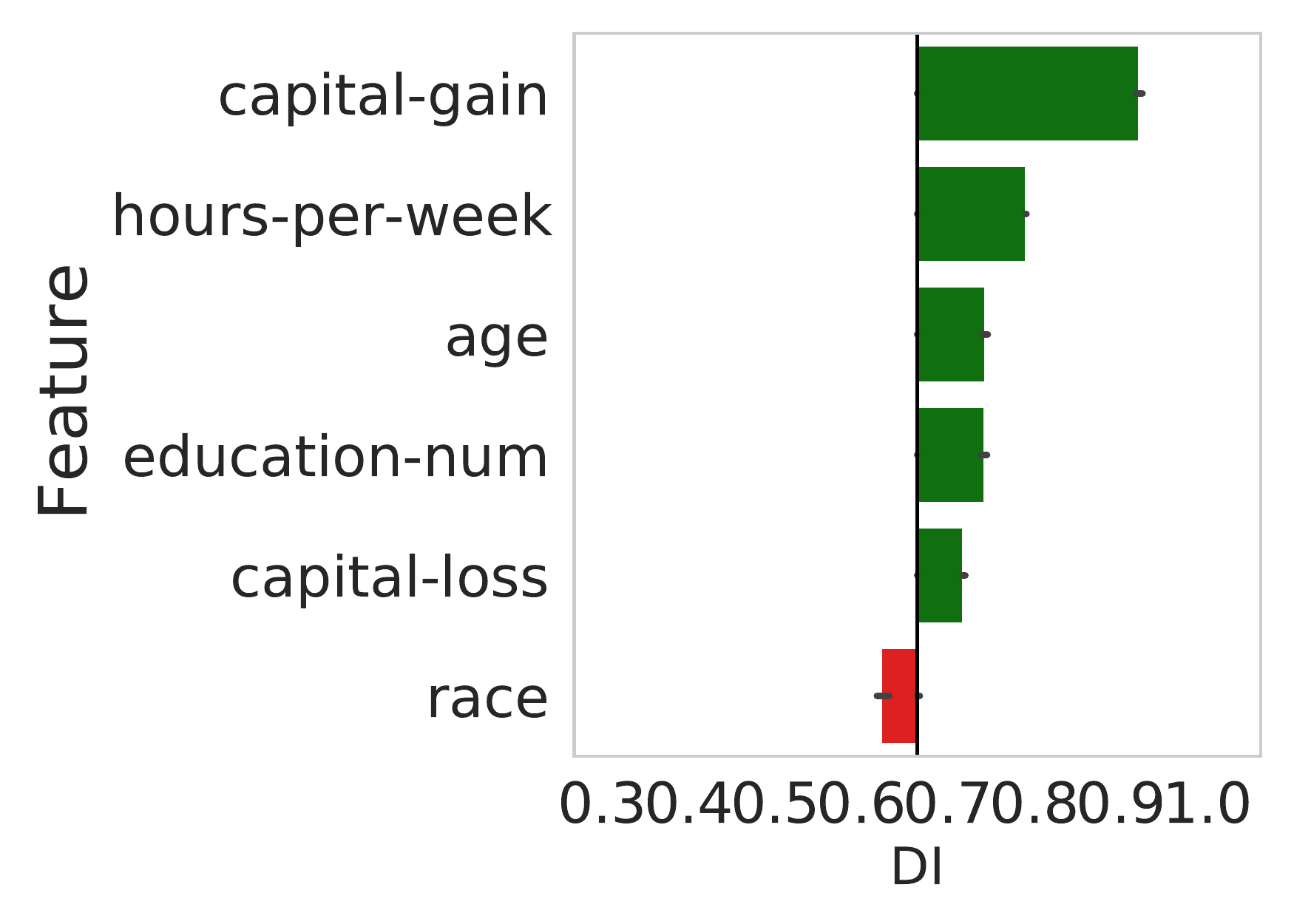}\label{fig:fif_f}}\\

	\end{center}
	\caption{Extended results on computing feature influence functions (FIF) for COMPAS  and Adult dataset.}\label{fig:fif}
\end{figure}

For both the datasets, we consider biological `sex' as the sensitive features.
In both cases, we denote the sensitive groups `male' and `female' as `sex=0' and `sex=1', respectively. 
FIFs computed for these sensitive groups show influence of different features and their relevant disparities. We illustrate the results in Figure~\ref{fig:fif}.

Figure~\ref{fig:fif_a} illustrates that the probability of positive prediction  for the `male' population in the dataset is $0.61$. If we now replace the feature `age' with uniformly random values, the probability of positive prediction for the same group becomes $0.46$, Thus, the FIF of the feature `age' for `male' group is $0.15$.
Similarly, all the red bars in Figures~\ref{fig:fif_a}-\ref{fig:fif_b}, ~\ref{fig:fif_c}-\ref{fig:fif_d} indicate a positive influence of that feature and the green bars indicate a negative influence of the feature on the individual getting classified to $\hat{Y}=1$.

In Figures~\ref{fig:fif_c} and~\ref{fig:fif_f}, we illustrate the influence of different features on the disparate impact between the two groups `male' and `female'. 
The green bar indicates that removing a feature causes an increase in DI, i.e. fairness in classification, and the red bar indicates the opposite. Alternatively, we can conclude that bigger is the green bar for a feature, higher is the bias-inducing effect of it.

Thus, from Figure~\ref{fig:fif_c}, we conclude that `age' is the most bias-inducing feature among the two groups `male' and `female'.
From Fig~\ref{fig:fif_a} and~\ref{fig:fif_b}, we observe that `age' is a decisive feature for `male' while it is comparatively insignificant for `female'.

In case of Adult dataset, the probability of positive prediction for `male' and `female' groups are 0.45 and 0.3 respectively. Among all the features, we observe that `race' has a positive impact on the classification for both the groups. This means removing the `race' information decreases the chance of getting classified to higher economic group. On the other hand, `capital-gain' has the highest influence on the classification for both groups. It is also the most biased-inducing feature as it varies most disparately among the two groups as the `capital-gain' differs significantly between two groups. 

\section{Fairness Verification of CNF Classifiers with Feature-correlations} 
\label{sec:CNF_feature_correlation}
Existing verifiers such as Justicia focuses mainly on classifiers expressible as a CNF formula. But none of the existing verifiers consider correlation among the features while computing fairness metrics. In this section,  we address fairness verification of CNF-based classifiers with explicit consideration of correlations among features. In particular, we present how to encode a Bayesian network, which captures correlated features, into a SSAT-based formulation that is tailored for verifying CNF classifiers. To this end, we first discuss the basics of SSAT followed by an elaboration of the proposed methodology.

\subsection{Background: Stochastic Boolean Satisfiability (SSAT)}\label{sec:ssat}
Let $ \mathbf{B}  = \{\bool_1, \dots, \bool_m\}  $ be a set of Boolean variables taking assignment in $ \{0,1\}^m $. A \textit{literal} $ l_i $ is a variable $ \bool_i $ or its complement $ \neg \bool_i $. A propositional formula $\phi$ defined over $\mathbf{B}$ is in CNF if $\phi \triangleq \wedge_i C_i $   is  a conjunction of clauses and each clause $ C_i \triangleq \vee_j l_j $ is a disjunction of literals. A CNF formula $ \phi $ is satisfied (called SAT) if there is an assignment $ \sigma $ over $\mathbf{B}$ that evaluates formula $ \phi $ to $ 1 $ (or true)\textemdash at least one literal in each clause is true and all clauses are true. We additionally consider a quantification over each Boolean variable $ B_i $, denoted by $ q_i \in \{\exists, \forall, \R^{p_i}\}$, where $ \exists $ is existential, $ \forall $ is universal and $ \R $ is a randomized quantifier with $ p_i = \Pr[B_i = 1] $. The SSAT problem takes \textit{an ordered set of quantifiers} over  $\mathbf{B}$ and a CNF formula $ \phi $ and computes the \textit{probability of satisfaction} of $ \phi $ given quantifiers.  Formally, a SSAT formula $ \Phi \triangleq (\{q_i\}_{i=1}^{m}, \phi) $ and the SSAT problem computes $ \Pr[\Phi]  \in [0,1]$. The semantics of SSAT formula is in the following.

\begin{enumerate}
	\item $ \Pr[\text{true}] = 1 $,  $ \Pr[\text{false}] = 0 $, 
	\item $ \Pr [\Phi] = \max_{\bool_1} \{\Pr[\Phi|_{\bool_1}], \Pr[\Phi|_{\neg \bool_1}]\}$ if $ \bool_1 $ is existentially quantified ($ \exists $), 
	\item $ \Pr [\Phi] = \min_{\bool_1} \{\Pr[\Phi|_{\bool_1}], \Pr[\Phi|_{\neg \bool_1}]\} $ if $ \bool_1 $ is universally quantified ($ \forall $), 
	\item $ \Pr [\Phi] = p\Pr[\Phi|_{\bool_1}] + (1-p) \Pr[\Phi|_{\neg \bool_1}] $ if $ \bool_1 $ is randomized quantified ($\R^{p}$) with probability $p = \Pr[\bool_1 = 1]$,
\end{enumerate}

In SSAT, we recursively solve for each Boolean variable in $\mathbf{B}$ starting with the first variable $ B_1 $
where $ \Phi|_{\bool_1} $ (resp.\ $ \Phi|_{\neg \bool_1} $) is the SSAT formula with CNF $ \phi $ substituted by an assignment of $ \bool_1 $ as true (resp.\ false) and quantifiers $ \{q_i\}_{i=2}^{m} $. 	

In this paper, we are interested in two specific SSAT formulations: exists-random (ER) SSAT formulas and universal-random (UR) SSAT formulas. In ER-SSAT (resp.\ UR-SSAT), $ \{q_i\} $ is constructed such that  existential (resp.\ universal) quantified variables is followed by randomized quantified variables. For more details on SSAT formulas,  we refer to \cite{ghosh2020justicia,lee2017solving, lee2018solving}.


\subsection{Methodology}
For CNF classifiers~\cite{GMM20}, SSAT is a natural choice as it computes the probability of a CNF formula  given quantifiers of variables\textemdash equivalently, the probability of positive prediction of a CNF classifier. \cite{ghosh2020justicia} has proposed a SSAT based formulation for verifying a CNF classifier $ \phi_{\hat{Y}} $ with a fundamental limitation of ignoring feature-correlations, which we address now. Let $ \phi_\BN $ be a CNF formula that encodes the Bayesian Network $ \BN $. Following~\cite{ghosh2020justicia}, we discuss an encoding that compute the  probability of positive prediction of the classifier for the most favored sensitive group with additional consideration of features-correlation. Let $\mathbf{Q}$ denote an ordered set of quantifiers over variables in the conjoined CNF $ \phi_{\hat{Y}} \wedge \phi_\BN $. We then construct a SSAT formula $ \Phi =  (\mathbf{Q},  \phi_{\hat{Y}} \wedge \phi_\BN) $ and solve it for computing $ \max_{\mathbf{a}} \Pr[\hat{Y} = 1 | \sensitive = \mathbf{a}] $. For computing $ \min_{\mathbf{a}} \Pr[\hat{Y} = 1 | \sensitive = \mathbf{a}] $ for the least favored group ,only difference is in the construction of $\mathbf{Q}$. We next discuss the construction of both $ \phi_\BN $ and $\mathbf{Q}$.

\paragraph{Encoding a Bayesian Network as a CNF Formula.}\label{sec:BN_to_CNF}
Our goal is to encode the Bayesian network $ \BN = (\graph, \factors) $ into a CNF formula $ \phi_\BN $ such that \textit{the weighted model count} of $ \phi_\BN $ exactly computes a joint probability distribution~\cite{chavira2008probabilistic}.  We note that SSAT does not allow conditional probabilities of randomized quantified variables trivially. Hence, $ \phi_\BN $ contains \textit{additional variables} to capture the conditional probabilities, as discussed next.

Let  $ G = (\mathbf{V}, E) $ where $  $  $ \mathbf{V} \subseteq \nonsensitive \cup \sensitive $ and $ \mathbf{E} \subseteq \mathbf{V} \times \mathbf{V} $. 	For each network variable $ V_i \in \mathbf{V} $, we define a Boolean \textit{indicator}  variable $ \lambda_{V_i} $ such that $ \Pr[\lambda_{V_i}] \triangleq \Pr[V_i] $. We add following constraint in $ \phi_\BN $ to establish the relation between $ \lambda_{V_i} $ and $ V_i $. 
\begin{align}
	\lambda_{V_i} \leftrightarrow V_i,
	\label{eq:indicator_constraint}
\end{align}
Intuitively, both $ \lambda_{V_i} $ and $ V_i $ are either true or false. This constraint can be trivially translated to clauses in CNF using the equivalence rule $ a \leftrightarrow b\equiv (\neg a \vee b)  \wedge (a \vee \neg a) $ for Boolean variables $ a, b$.

We now show encoding of conditional probabilities induced by parameter $ \theta $. Let $ V_i \in  \mathbf{V}  $  be a vertex in $ G $ where $ \parent(V_i) \ne \emptyset $ be $ V_i $'s parents and $ |\parent(V_i)| = k $. Additionally, let $ v $ and $ \mathbf{u} \triangleq [u_1,.., u_k] $ be an assignment of  $ V_i $ and $ \parent(V_i)  $, respectively.  To encode $ \Pr[V_i = v| \parent(V_i) = \mathbf{u}]$, we introduce auxiliary variable $ \lambda_{v,\mathbf{u}} $ and add following constraints in $ \phi_{\BN} $.
\begin{align}
	\lambda_{v,\mathbf{u}}  \wedge \bigwedge_{j=1}^{k} \lambda_{u_j} \rightarrow \lambda_v
	\label{eq:factor_pos}
\end{align}
\begin{align}
	\neg \lambda_{v,\mathbf{u}}  \wedge \bigwedge_{j=1}^{k} \lambda_{u_j} \rightarrow \neg \lambda_v
	\label{eq:factor_neg}
\end{align}

where $ \lambda_v \equiv \lambda_{V_i} $. Moreover, $ \lambda_{u_j} $ is the indicator variable corresponding to the $ j^\text{th} $ parent in $ \parent(V_i) $. In the above two constraints,	for a fixed assignment $ \mathbf{u} $ of parents $ \parent(V_i) $, both $ \lambda_v $ and $ \lambda_{v,\mathbf{u}} $ are either true or false.  Hence, these two constraints encode the conditional probability of $ V_i = v $ given $ \parent(V_i) = \mathbf{u} $ using $ \Pr[\lambda_{v,\mathbf{u}}] = \Pr[V_i = v| \parent(V_i) = \mathbf{u}]$. Both constraints can be translated to CNF clauses trivially. For example, Eq.~\ref{eq:factor_pos} is translated as $ \neg \lambda_{v,\mathbf{u}}  \vee \bigvee_{j=1}^{k} \neg \lambda_{u_j} \vee \lambda_v $. We next analyze the complexity of $ \phi_\BN $ in terms of the number of variables and clauses.

\begin{lemma}
	For a Bayesian network $ \BN = (\graph, \factors) $ defined over $ n $ Boolean network variables, the encoded CNF formula $ \phi_\BN $ has $ n + |\factors| $  variables and $ 2(n + |\factors|) $ clauses. 
\end{lemma}
\begin{proof}
	Since the DAG in the Bayesian network has $ n $ vertices, we consider $ n $ indicator variables. Moreover, for encoding conditional probabilities, we consider $ |\theta| $ auxiliary variables where parameter $ \theta $ denotes the number of distinct conditional probabilities in the network. Hence, total variables in $ \phi_\BN $ is $ n + |\factors|  $.
	
	According to Eq.~\eqref{eq:indicator_constraint},~\eqref{eq:factor_pos},~\eqref{eq:factor_neg}, there are $2( n + |\theta|) $ clauses in $ \phi_\BN $ as discussed in Section~\ref{sec:BN_to_CNF}.

\end{proof}

\begin{figure}[!t]
	\begin{center}
		\subfloat[]{\includegraphics[scale=0.4]{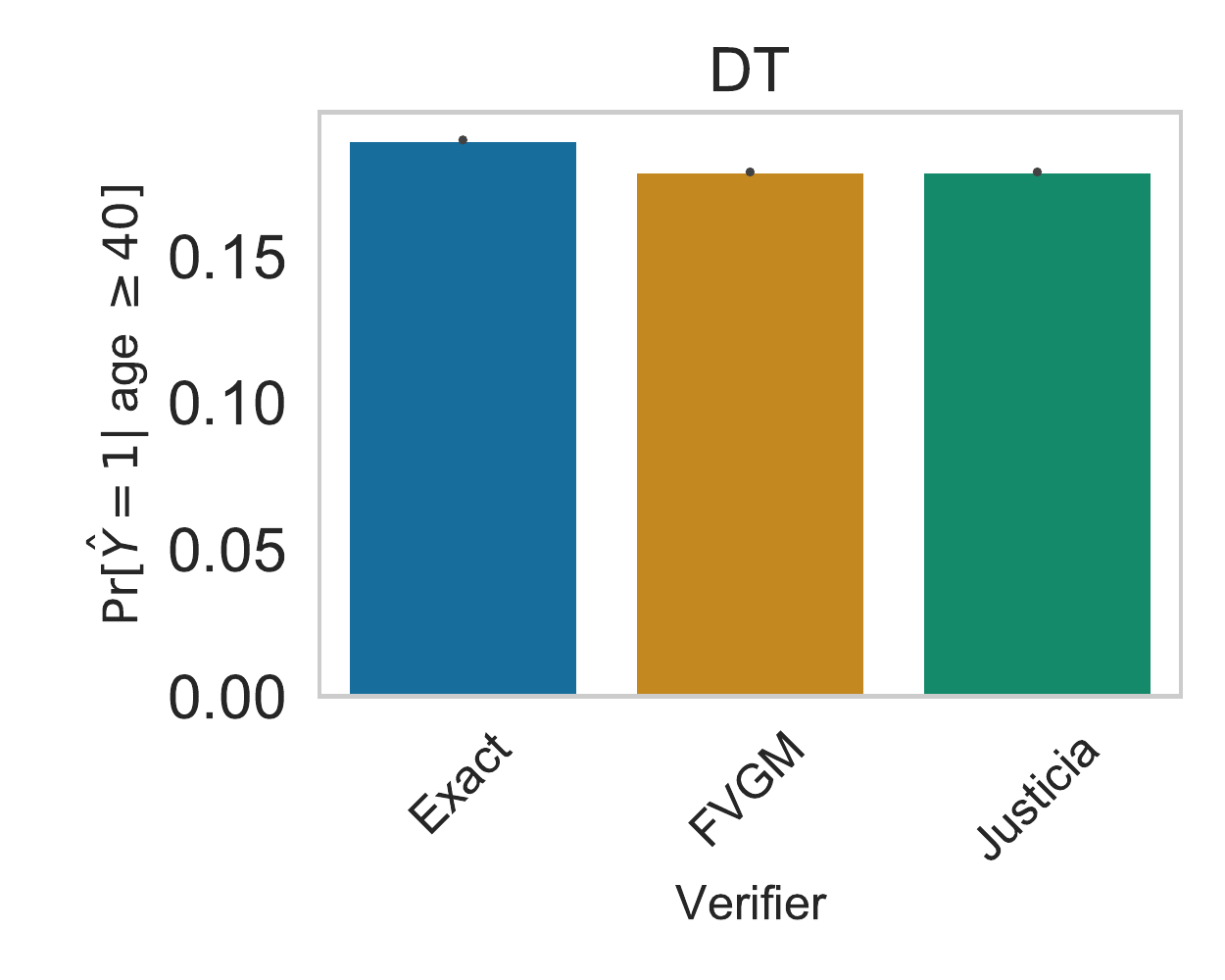}}
		\subfloat[]{\includegraphics[scale=0.4]{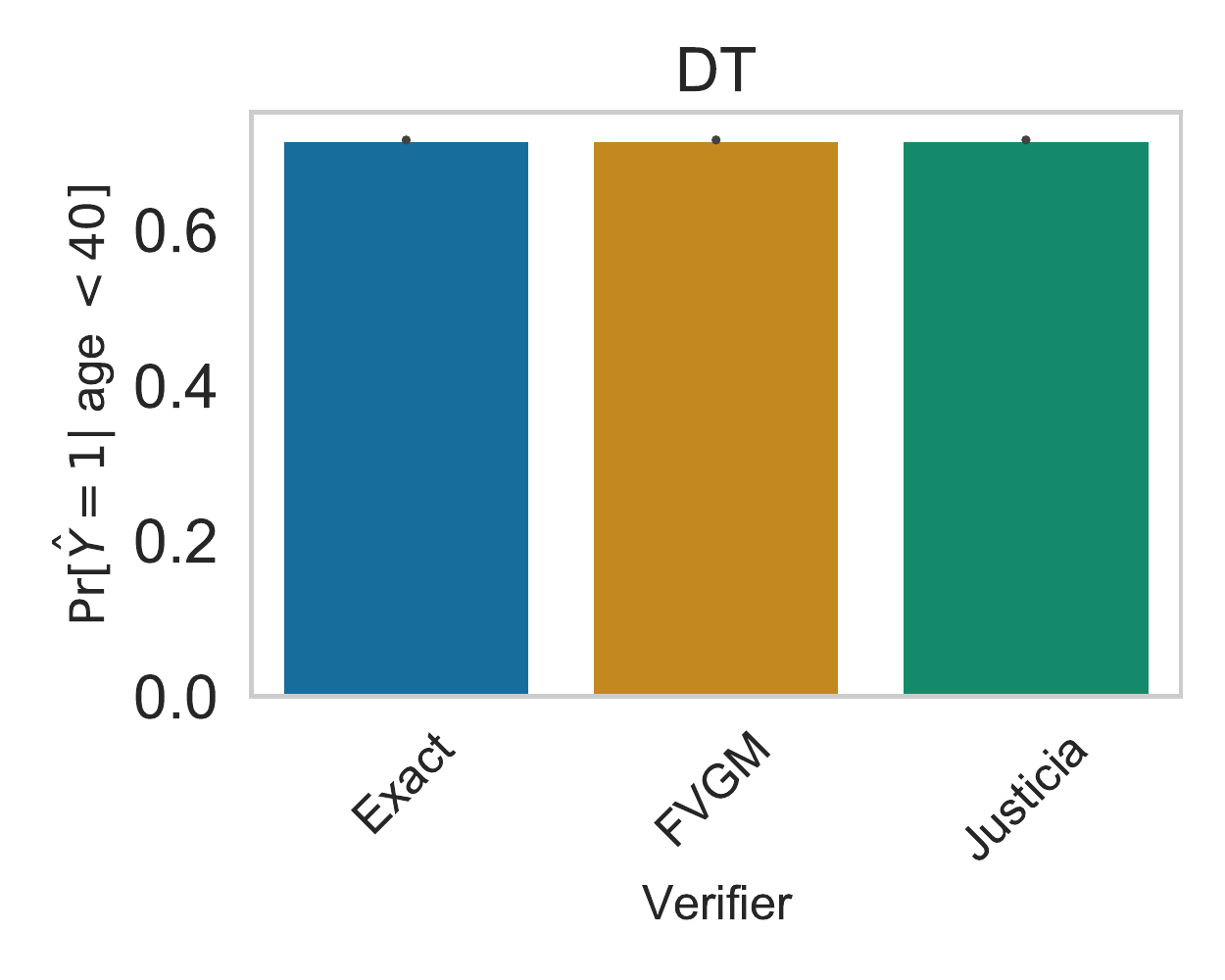}}\\
		\subfloat[]{\includegraphics[scale=0.33]{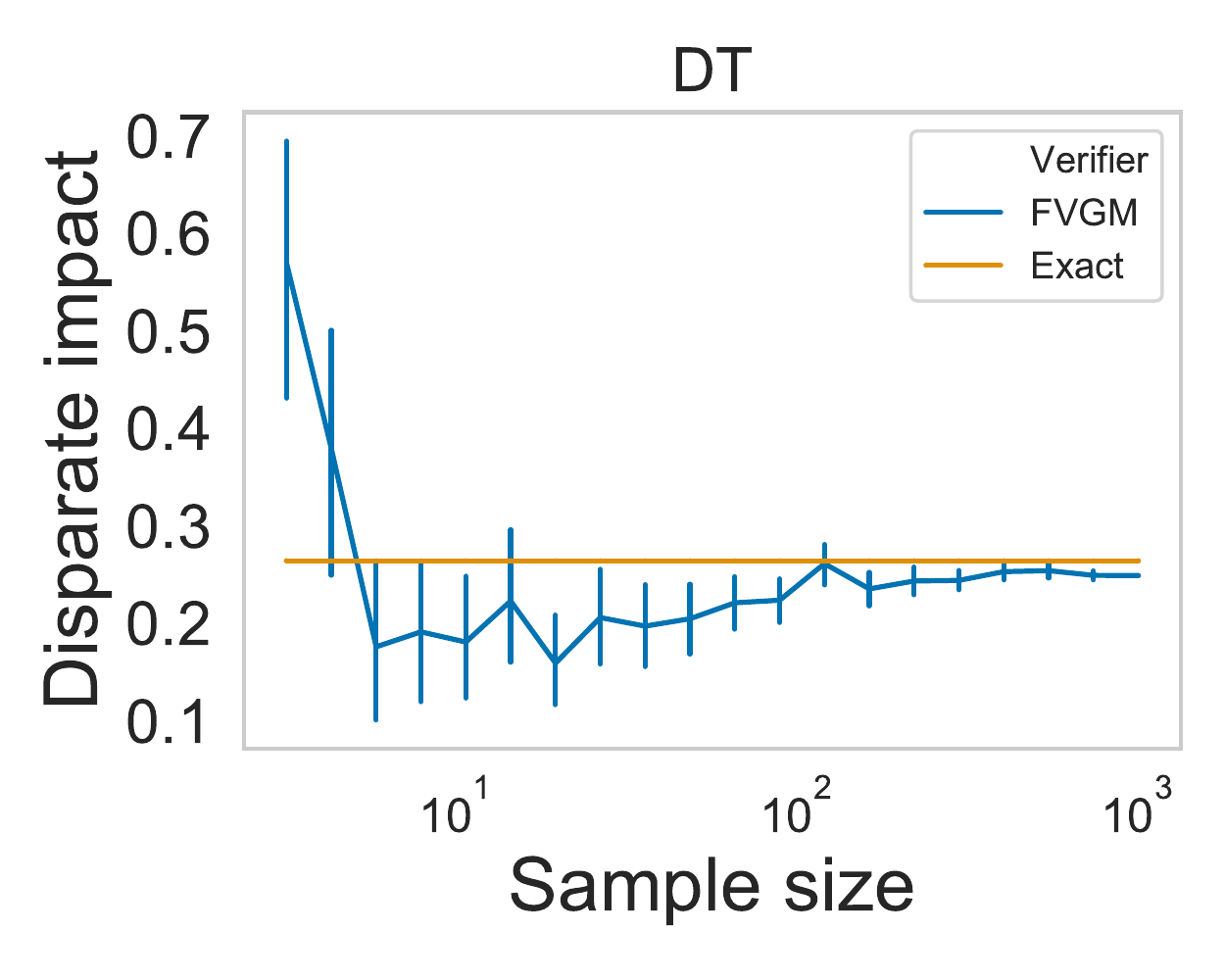}}
		\subfloat[]{\includegraphics[scale=0.33]{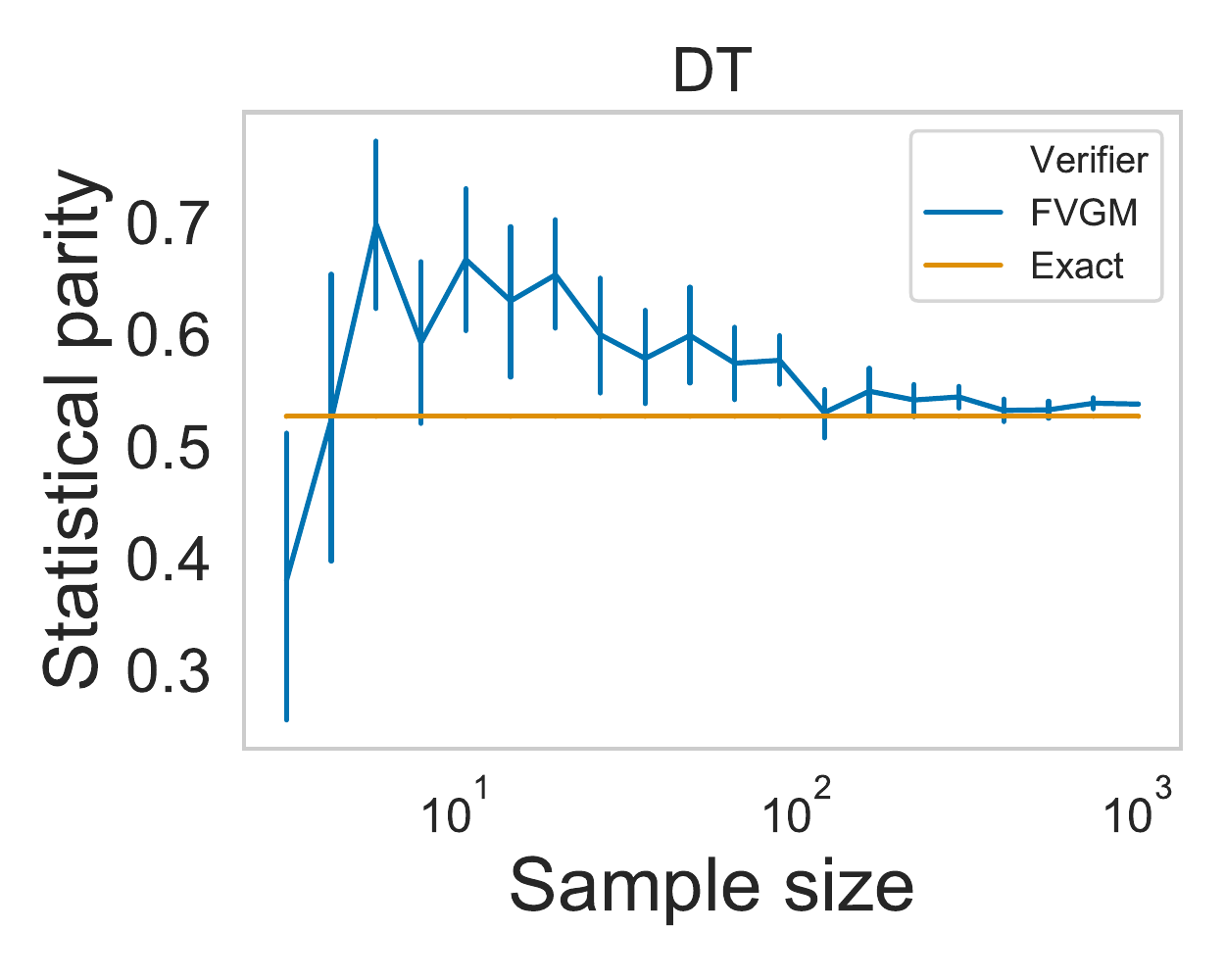}}

	\end{center}
	
	\caption{
		Computing the probability of positive prediction of Decision Tree (DT) classifiers using {\framework} and Justicia. {\framework} incorporates correlated features represented as a Bayesian network in Figure~\ref{fig:synthetic_bn}. We also present the effect of sample size on different fairness metrics such as DI and SP for DT classifier. }
	\label{fig:synthetic_results}
\end{figure}

\paragraph{Constructing Quantifiers $ \mathbf{Q} $.}
We now discuss the ordered set of quantifiers for a SSAT formula containing CNF $ \phi_{\hat{Y}} \wedge \phi_\BN $\textemdash the solution of which constitutes the maximum (minimum) probability of positive prediction of a CNF classifier. $ \phi_{\hat{Y}} \wedge \phi_\BN $ contains four different variables : (i) sensitive variables $ \sensitive $, (ii) non-sensitive variables $ \nonsensitive $, (iii) indicator variables $  \lambda_{V_i} $, and (iv) auxiliary variables $ \lambda_{v,\mathbf{u}} $. Among them, (iii) and (iv) are associated with $ \phi_\BN $ and the rest for $ \phi_{\hat{Y}} $. For computing the maximum probability of positive prediction of the classifier, we construct  an exists-random-exists (ERE) SSAT formula with quantifiers $\mathbf{Q}$ as follows: we set sensitive features $ \sensitive $ with existential quantifiers in the beginning of $ \mathbf{Q} $ followed by $  \lambda_{V_i}, \lambda_{v,\mathbf{u}} $ and $ X_j \in \nonsensitive \setminus \mathbf{V} $ with randomized quantifiers. Since sensitive variables are existentially quantified\textemdash similar to stochastic subset-sum problem\textemdash the solution of SSAT formula is maximized.
Finally, remaining variables appearing in the Bayesian network such as $ X_i \in \mathbf{V} $ are existentially quantified in $\mathbf{Q}$ as their assignment is fixed by indicator variables $ \lambda_{V_i} $. In contrast, for computing the minimum probability of positive prediction of the classifier, we consider an universal-random-exists (URE) SSAT formula  where we set sensitive features $ \sensitive $ as universal quantifiers with all other quantifiers remaining same.

\end{document}